\documentstyle[12pt]{article}
\newtheorem{definition}{Definition}[section]
\newtheorem{example}{Example}[section]
\newtheorem{remark}{Remark}[section]
\newtheorem{lemma}{Lemma}[section]
\newtheorem{theorem}[lemma]{Theorem}
\newtheorem{corollary}[lemma]{Corollary}

\newtheorem{optimization}{Optimization}
\newtheorem{fact}{Fact}

\begin{document}

\title{\Large \bf SLT-Resolution for the
Well-Founded Semantics}

\author{Yi-Dong Shen\thanks{Work completed during a visit at
Department of Computing Science, 
University of Alberta, Canada.}\\
{\small  Department of Computer Science,
Chongqing University, Chongqing 400044, P.R.China}\\
{\small Email: ydshen@cs.ualberta.ca}\\[.1in]
Li-Yan Yuan  and Jia-Huai You\\
{\small  Department of Computing Science, University of
Alberta, Edmonton, Alberta, Canada T6G 2H1}\\
{\small  Email: \{yuan, you\}@cs.ualberta.ca}}

\date{}

\maketitle
 
\begin{abstract} 
Global SLS-resolution and SLG-resolution are two representative
mechanisms for top-down evaluation of the well-founded
semantics of general logic programs. Global SLS-resolution
is linear for query evaluation but suffers from infinite loops and redundant
computations. In contrast, SLG-resolution resolves infinite 
loops and redundant computations by means of tabling, but it is not linear. 
The principal disadvantage of a non-linear approach is that it cannot 
be implemented using a simple, efficient stack-based memory structure 
nor can it be easily extended to handle some 
strictly sequential operators such as cuts in Prolog.

In this paper, we present a linear tabling method, called {\em SLT-resolution},
for top-down evaluation of the well-founded semantics.
SLT-resolution is a substantial extension of SLDNF-resolution with tabling.
Its main features include: (1) It resolves infinite loops and redundant computations 
while preserving the linearity. (2) It is terminating, and sound and complete w.r.t. 
the well-founded semantics for programs with the bounded-term-size property 
with non-floundering queries. Its time complexity is comparable with SLG-resolution and
polynomial for function-free logic programs. (3) Because of its linearity for query 
evaluation, SLT-resolution bridges the gap between the well-founded semantics
and standard Prolog implementation techniques. It can be implemented by an extension 
to any existing Prolog abstract machines such as WAM or ATOAM.\\[.1in]
\end{abstract} 
 
\section{Introduction}

The central component of existing logic programming systems is a refutation
procedure, which is based on the resolution rule created by Robinson \cite{Robinson65}.
The first such refutation procedure, called {\em SLD-resolution},
was introduced by Kowalski \cite{Kow74, EK76}, and further formalized
by Apt and Van Emden \cite{AV82}. SLD-resolution is only suitable for
positive logic programs, i.e. programs without negation.
Clark \cite{clark78} extended SLD-resolution to {\em SLDNF-resolution}
by introducing the {\em negation as finite failure rule}, which is used 
to infer negative information. SLDNF-resolution is suitable for 
general logic programs, by which a ground negative literal
$\neg A$ succeeds if $A$ finitely fails, and fails if $A$ succeeds.

As an operational/procedural semantics of logic programs, SLDNF-resolution has 
many advantages, among the most important of which is its {\em linearity}
of derivations. Let $G_0 \Rightarrow_{C_1, \theta_1} G_1 \Rightarrow$
$...  \Rightarrow_{C_i, \theta_i} G_i$ be a derivation 
with $G_0$ the top goal and $G_i$ the 
latest generated goal. A resolution is said to be 
{\em linear} for query evaluation if when
applying the most widely used {\em depth-first} search rule,
it makes the next derivation step either by expanding $G_i$ using
a program clause (or a tabled answer), which yields
$G_i \Rightarrow_{C_{i+1}, \theta_{i+1}} G_{i+1}$,
or by expanding $G_{i-1}$ via backtracking.\footnote{
The concept of ``linear'' here is different from the one
used for SL-resolution \cite{KK71}.} It is with such linearity
that SLDNF-resolution can be realized easily and 
efficiently using a simple stack-based memory structure
\cite{WAM83, ZHOU96}. This has been sufficiently demonstrated
by Prolog, the first and yet the most popular logic programming
language which implements SLDNF-resolution. 
 
However, SLDNF-resolution suffers from two
serious problems. One is that the declarative semantics it
relies on, i.e. the {\em completion of programs} \cite{clark78}, incurs some
anomalies (see \cite{Ld87, sheph88} 
for a detailed discussion); and
the other is that it may generate infinite loops and a large amount
of redundant sub-derivations \cite{BAK91, DD93, VL89}.

The first problem with SLDNF-resolution has been perfectly settled by the discovery
of the {\em well-founded semantics} \cite{VRS91}.\footnote{Some other
important semantics, such as the {\em stable model semantics} \cite{GL88},
are also proposed. However, for the purpose of query evaluation
the well-founded semantics seems to be the most natural and robust.} 
Two representative methods were then proposed for 
top-down evaluation of such a new semantics:
Global SLS-resolution \cite{Prz89,Ross92} 
and SLG-resolution \cite{CSW95,chen96}.

Global SLS-resolution is a direct extension
of SLDNF-resolution. It overcomes the semantic anomalies of SLDNF-resolution by
treating infinite derivations as {\em failed}
and infinite recursions through negation as {\em undefined}. 
Like SLDNF-resolution, it is linear for query evaluation. However, it
inherits from SLDNF-resolution the problem of infinite loops
and redundant computations. Therefore, as the authors themselves
pointed out, Global SLS-resolution can be considered as a theoretical
construct \cite{Prz89} and is not effective in general \cite{Ross92}. 

SLG-resolution (similarly, Tabulated SLS-resolution \cite{BD98})
is a tabling mechanism for top-down evaluation of the well-founded
semantics. The main idea of tabling is to store intermediate results
of relevant subgoals and then use them to solve variants of the subgoals
whenever needed. With tabling no variant subgoals will be recomputed by applying
the same set of program clauses, so infinite loops can be avoided and 
redundant computations be substantially reduced
\cite{BD98,chen96,TS86,VL89,war92}. Like all other existing
tabling mechanisms, SLG-resolution adopts the {\em solution-lookup mode}.
That is, all nodes in a search tree/forest are partitioned into two subsets, 
{\em solution} nodes and {\em lookup} nodes. Solution nodes produce child nodes 
only using program clauses, whereas lookup nodes produce child nodes only using answers
in the tables. As an illustration, consider the derivation
$p(X) \Rightarrow_{C_{p_1}, \theta_1} q(X) \Rightarrow_{C_{q_1}, \theta_2} p(Y)$. 
Assume that so far no answers of $p(X)$ have been derived (i.e., 
currently the table for $p(X)$ is empty).
Since $p(Y)$ is a variant of $p(X)$ and thus a lookup node, 
the next derivation step is to expand $p(X)$ against a program clause, 
instead of expanding the latest generated goal $p(Y)$.
Apparently, such kind of resolutions is not linear for query evaluation. 
As a result, SLG-resolution cannot be implemented using
a simple, efficient stack-based memory structure 
nor can it be easily extended to handle some 
strictly sequential operators such as cuts in Prolog because the
sequentiality of these operators fully depends 
on the linearity of derivations.\footnote{
It is well known that cuts are indispensable in
real world programming practices.} This has been evidenced
by the fact that XSB, the best known state-of-the-art tabling system that implements
SLG-resolution, disallows clauses like \\
\indent $\qquad p(.) \leftarrow ...,t(.),!,...$\\
because the tabled predicate $t$
occurs in the scope of a cut \cite{SSW94, SSW98, SSWFR98}.

One interesting question then arises: Can we have a {\em linear tabling} method
for top-down evaluation of the well-founded semantics of general logic programs, 
which resolves infinite loops and redundant computations (like SLG-resolution) 
without sacrificing the linearity of SLDNF-resolution (like Global SLS-resolution)? 
In this paper, we give a positive answer to this question by developing
a new tabling mechanism, called {\em SLT-resolution}. SLT-resolution is 
a substantial extension of SLDNF-resolution with tabling. Its main features 
are as follows. 

\begin{itemize}
\item
SLT-resolution is based on finite {\em SLT-trees}. The construction of 
SLT-trees can be viewed as that of SLDNF-trees with an enhancement of
some loop handling mechanisms. Consider again the derivation
$p(X) \Rightarrow_{C_{p_1}, \theta_1} q(X) \Rightarrow_{C_{q_1}, \theta_2} p(Y)$.
Note that the derivation has gone into a loop since the proof of $p(X)$ needs 
the proof of $p(Y)$, a variant of $p(X)$. By SLDNF- or Global SLS-resolution,
$P(Y)$ will be expanded using the same set of program clauses as $p(X)$.
Obviously, this will lead to an infinite loop of the form
$p(X) \Rightarrow_{C_{p_1}}...p(Y)\Rightarrow_{C_{p_1}}...p(Z)\Rightarrow_{C_{p_1}}...$
In contrast, SLT-resolution will break the loop by disallowing $p(Y)$ to use
the clause $C_{p_1}$ that has been used by $p(X)$. As a result, SLT-trees are 
guaranteed to be finite for programs with the bounded-term-size property.

\item
SLT-resolution makes use of tabling to reduce redundant computations,
but is linear for query evaluation. Unlike SLG-resolution and all other
existing top-down tabling methods, SLT-resolution does not distinguish
between solution and lookup nodes. All nodes will be expanded by 
applying existing answers in tables, followed by program clauses. 
For instance, in the above example derivation, since currently there is no 
tabled answer available to $p(Y)$, $p(Y)$ will be expanded using some program 
clauses. If no program clauses are available to $p(Y)$, SLT-resolution 
would move back to $q(X)$ (assume using a depth-first control strategy). 
This shows that SLT-resolution is linear for query evaluation.  
When SLT-resolution moves back to $p(X)$, all program clauses that have
been used by $p(Y)$ will no longer be used by $p(X)$. This avoids
redundant computations.

\item
SLT-resolution is terminating, and sound and complete w.r.t. the well-founded
semantics for any programs with the bounded-term-size property with non-floundering
queries. Moreover, its time complexity is comparable with SLG-resolution and
polynomial for function-free logic programs.

\item
Because of its linearity for query evaluation, SLT-resolution can be implemented
by an extension to any existing Prolog abstract machines such as WAM \cite{WAM83}
or ATOAM \cite{ZHOU96}. This differs significantly from non-linear resolutions
such as SLG-resolution since their derivations cannot be organized using
a stack-based memory structure, which is the key to the Prolog implementation.

\end{itemize}

\subsection{Notation and Terminology} 

We present our notation and review some standard terminology of
logic programs \cite{Ld87}.

Variables begin with a capital letter, and predicate, function 
and constant symbols with a lower case letter. Let $p$ be a predicate
symbol. By $p(\vec{X})$ we denote an atom with the list $\vec{X}$
of variables. Let $S=\{A_1,...,A_n\}$ be a set of atoms. 
By $\neg .S$ we denote the complement $\{\neg A_1,...,\neg A_n\}$ of $S$.

\begin{definition}
{\em
A {\em general logic program} (program for short) is a finite set
of (program) clauses of the form

$\qquad A\leftarrow L_1,..., L_n$

\noindent where $A$ is an atom and $L_i$s are literals. 
$A$ is called the {\em head} and $L_1,...,L_n$ is called the
{\em body} of the clause. If a program has no clause with negative
literals in its body, it is called a {\em positive} program.
}
\end{definition}

\begin{definition}[\cite{Ross92}]
{\em
Let $P$ be a program and $\bar{p}$, $\bar{f}$ and $\bar{c}$
be a predicate symbol, function symbol and constant symbol
respectively, none of which appears in $P$. The {\em augmented
program} $\bar{P}=P\cup \{\bar{p}(\bar{f}(\bar{c}))\}$.
}
\end{definition}

\begin{definition}
{\em
A {\em goal} is a headless clause
$\leftarrow L_1,..., L_n$ where each $L_i$ is called a {\em subgoal}.
When $n=0$, the ``$\leftarrow$'' symbol is omitted. A {\em computation
rule} (or {\em selection rule}) is a rule for selecting one subgoal from a goal.
}
\end{definition}

Let $G_j=\leftarrow L_1,..., L_i,..., L_n$ be a 
goal with $L_i$ a positive subgoal. Let $C_l=L\leftarrow F_1,...,F_m$ be 
a clause such that $L\theta = L_i\theta$ where $\theta$ is an mgu (i.e. 
most general unifier). The {\em resolvent} of $G_j$ and $C_l$ on $L_i$ is the goal
$G_k=\leftarrow (L_1,...,L_{i-1}, F_1,...,F_m, L_{i+1},..., L_n)\theta$.
In this case, we say that the proof of $G_j$ is reduced to the proof of $G_k$.

The initial goal, $G_0=\leftarrow L_1,..., L_n$, is called
a {\em top} goal. Without loss of generality, we shall assume throughout
the paper that a top goal consists only of one atom (i.e. $n=1$ and $L_1$
is a positive literal). Moreover, we assume that the same computation rule $R$ always 
selects subgoals at the same position in any goals. For instance,
if $L_i$ in the above goal $G_j$ is selected by $R$, then $F_1\theta$ in $G_k$
will be selected by $R$ since $L_i$ and $F_1\theta$ are at the same position
in their respective goals.  

\begin{definition}
{\em
Let $P$ be a program. The {\em Herbrand universe} of $P$ is the set
of ground terms that use the function symbols and constants in $P$.
(If there is no constant in $P$, then an arbitrary one is added.)
The {\em Herbrand base} of $P$ is the set of ground atoms formed
by predicates in $P$ whose arguments are in the Herbrand universe.
By $\exists (Q)$ and $\forall (Q)$ we denote respectively
the existential and universal closure of $Q$ over the Herbrand
universe.
}
\end{definition}
 
\begin{definition}
{\em
A {\em Herbrand instantiated clause} of a program $P$ is a ground
instance of some clause $C$ in $P$ that is obtained by replacing
all variables in $C$ with some terms in the Herbrand universe of $P$.
The {\em Herbrand instantiation} of $P$ is the set of all
Herbrand instantiated clauses of $P$. 
}
\end{definition}

\begin{definition}
{\em
Let $P$ be a program and $H_P$ its Herbrand base.
A {\em partial interpretation} $I$
of $P$ is a set $\{A_1,...,A_m,\neg B_1,...,\neg B_n\}$
such that $\{A_1,...,A_m,B_1,...,B_n\}\subseteq H_P$ and
$\{A_1,...,A_m\}\cap \{B_1,...,B_n\}=\emptyset$. We use
$I^+$ and $I^-$ to refer to $\{A_1,...,A_m\}$ and
$\{B_1,...,B_n\}$, respectively.
}
\end{definition}

\begin{definition}
{\em
By a {\em variant} of a literal $L$ we mean a literal $L'$ that is the same 
as $L$ up to variable renaming. (Note that $L$ is a variant of itself.)
}
\end{definition}

Finally, a substitution $\alpha$ is {\em more general than} a substitution
$\beta$ if there exists a substitution $\gamma$ such that
$\beta=\alpha\gamma$. Note that $\alpha$ is more general than
itself because $\alpha=\alpha\varepsilon$ where $\varepsilon$
is the identity substitution \cite{Ld87}.  
 
\section{The Well-Founded Semantics}

In this section we review the definition of the well-founded
semantics of logic programs. We also present a new constructive
definition of the greatest unfounded set of a program, which has
technical advantages for the proof of our results.

\begin{definition}[\cite{Ross92, VRS91}]
\label{unf-set}
{\em
Let $P$ be a program and $H_P$ its Herbrand base. Let $I$ be a
partial interpretation. $U\subseteq H_P$ is an {\em unfounded set of $P$
w.r.t. $I$} if each atom $A\in U$ satisfies the following condition: For
each Herbrand instantiated clause $C$ of $P$ whose head is $A$, at least 
one of the following holds:
\begin{enumerate}
\item
The complement of some literal in the body of $C$ is in $I$.

\item
Some positive literal in the body of $C$ is in $U$.
\end{enumerate}
The {\em greatest unfounded set} of $P$ w.r.t. $I$,
denoted $U_P(I)$, is the union of all sets that are unfounded
w.r.t. $I$.
}
\end{definition}

\begin{definition}[\cite{Ross92}]
\label{trans-ops1}
{\em
Define the following transformations:
\begin{itemize}
\item
$A\in T_P(I)$ if and only if there is a Herbrand instantiated clause of 
$P$, $A\leftarrow L_1,...,L_m$, such that all $L_i$ are in $I$. 

\item
$\bar T_P(I)=T_P(I)\cup I$.

\item
$M_P(I)=\bigcup_{k=1}^\infty {\bar T}_P^k(I)$, where 
$\bar T_P^1(I)=\bar T_P(I)$, and for any $i>1$
$\bar T_P^i(I)=\bar T_P(\bar T_P^{i-1}(I))$.

\item
$U_P(I)$ is the greatest unfounded set of $P$ w.r.t. $I$,
as in Definition \ref{unf-set}.

\item
$V_P(I)=M_P(I)\cup \neg . U_P(I)$.

\end{itemize}
}
\end{definition}

Since $T_P(I)$ derives only positive literals,
the following result is straightforward.

\begin{lemma}
\label{lem2-1}
$\neg A\in M_P(I)$ if and only if $\neg A\in I$.
\end{lemma}

\begin{definition}[\cite{Ross92, VRS91}]
\label{trans-ops2}
{\em
Let $\alpha$ and $\beta$ be countable ordinals. The 
partial interpretations $I_\alpha$ are
defined recursively by
\begin{enumerate}
\item
For limit ordinal $\alpha$, $I_\alpha=\bigcup_{\beta <\alpha} I_\beta$,
where $I_0=\emptyset$. 

\item
For successor ordinal $\alpha +1$, $I_{\alpha +1}=V_P(I_\alpha)$.
\end{enumerate}
}
\end{definition}

The transfinite sequence $I_\alpha$ is
monotonically increasing (i.e. $I_\beta \subseteq I_\alpha$ 
if $\beta\leq \alpha$), so there exists the first ordinal
$\delta$ such that $I_{\delta +1}=I_\delta$. This fixpoint
partial interpretation, denoted $WF(P)$, is called the 
{\em well-founded model} of $P$. Then for any $A\in H_P$,
$A$ is true if $A\in WF(P)$, false 
if $\neg A\in WF(P)$, and undefined otherwise. 

\begin{lemma}
\label{lem2-2}
For any $J\subseteq WF(P)$, $M_P(J)\subseteq WF(P)$ and
$\neg .U_P(J)\subseteq WF(P)$.
\end{lemma}

\noindent {\bf Proof:} Let $J\subseteq I_m$. Since $I_\alpha$ is
monotonically increasing, $M_P(J)\subseteq I_{m+1}\subseteq WF(P)$ and
$\neg .U_P(J)\subseteq I_{m+1}\subseteq WF(P)$. $\Box$

\vspace{4mm}

The following definition is adapted from \cite{Prz90}.
 
\begin{definition}
\label{p-i}
{\em
$P|I$ is obtained from the Herbrand instantiation $P_{H_P}$ of $P$ by
\begin{itemize}
\item 
first deleting all clauses with a literal in their bodies
whose complement is in $I$, 

\item
then deleting all negative literals in the remaining clauses.
\end{itemize}
}
\end{definition}

Clearly $P|I$ is a positive program. Note that for any partial
interpretation $I$, $M_P(I)$ is a partial interpretation 
that consists of $I$ and 
all ground atoms that are iteratively derivable from $P_{H_P}$
and $I$. We observe that the
greatest unfounded set $U_P(I)$ of $P$ w.r.t. $I$ can be
constructively defined based on $M_P(I)$ and $P|M_P(I)$.

\begin{definition}
\label{vp}
{\em
Define the following two transformations:
\begin{itemize}
\item
$ N_P(I)=H_P-\bigcup_{k=1}^\infty {\bar T}_{P|M_P(I)}^k (M_P(I))$.
\item 
$O_P(I)=\bigcup_{k=1}^\infty {\bar T}_{P|M_P(I)}^k (M_P(I))-M_P(I)$.
\end{itemize} 
}
\end{definition}

We will show that $N_P(I)=U_P(I)$ (see Theorem \ref{wfs-1}).
The following result is immediate.

\begin{lemma}
\label{lem2-3}
$M_P(I)^+$, $N_P(I)$ and $O_P(I)$ are mutually disjoint
and $H_P=M_P(I)^+\cup N_P(I)\cup O_P(I)$.
\end{lemma}

From Definitions \ref{p-i} and \ref{vp} it is easily seen that
$O_P(I)=\bigcup_{i=1}^\infty S_i$, which is generated iteratively 
as follows: First, for each $A\in S_1$ there must be
a Herbrand instantiated clause of $P$ of the form
\begin{equation}
A\leftarrow B_1,...,B_m,\neg D_1,...,\neg D_n \qquad\qquad
\qquad\qquad\label{equa1}
\end{equation}
where all $B_i$s and some $\neg D_j$s are in $M_P(I)$ and
for the remaining $\neg D_k$s (not empty; otherwise $A\in M_P(I)$) neither $D_k$ 
nor $\neg D_k$ is in $M_P(I)$. Note that the proof of $A$
can be reduced to the proof of $\neg D_k$s given $M_P(I)$.
Then for each $A\in S_2$ there must be a clause like (\ref{equa1})
above where no $D_j$ is in $M_P(I)$, some $B_i$s 
are in $M_P(I)$, and the remaining $B_k$s (not empty) are in 
$S_1$. Continuing such process of reduction, 
for each $A\in S_{l+1}$ with $l\geq 1$ there must be a clause like (\ref{equa1})
above where no $D_j$ is in $M_P(I)$, some $B_i$s 
are in $M_P(I)$, and the remaining $B_k$s (not empty) are in 
$\bigcup_{i=1}^l S_i$.

The following lemma shows a useful property of literals in $O_P(I)$.

\begin{lemma}
\label{lem2-4}
Given $M_P(I)$, the proof of any $A\in O_P(I)$ can be reduced to the proof
of a set of ground negative literals $\neg E_j$s where neither $E_j$ 
nor $\neg E_j$ is in $M_P(I)$.
\end{lemma}

\noindent {\bf Proof:} Let $O_P(I)=\bigcup_{i=1}^\infty S_i$. 
The lemma is proved by induction on $S_i$. Obviously,
it holds for each $A\in S_1$. As inductive hypothesis,
assume that the lemma holds for 
any $A\in S_i$ with $1\leq i\leq l$. We now prove that
it holds for each $A\in S_{l+1}$.

Let $A\in S_{l+1}$. For convenience of presentation, 
in clause (\ref{equa1}) above for $A$
let $\{B_1,...,B_f\}\subseteq M_P(I)$ $(f<m)$,
$\{B_{f+1},...,B_m\}\subseteq \bigcup_{i=1}^l S_i$, 
$\{\neg D_1,...,\neg D_e\}\subseteq M_P(I)$ $(e\leq n)$,
and for each $D_k\in \{D_{e+1},...,D_n\}$ neither $D_k$ 
nor $\neg D_k$ is in $M_P(I)$. By the inductive hypothesis
the proof of $B_{f+1},...,B_m$ can be reduced to the proof of
a set $NS=\{\neg N_1,...,\neg N_t\}$ of negative literals where 
neither $N_j$ nor $\neg N_j$ is in $M_P(I)$. So the proof of
$A$ can be reduced to the proof of $\{\neg N_1,...,\neg N_t,
\neg D_{e+1},...,\neg D_n\}$. $\Box$

\begin{theorem}
\label{wfs-1}
$N_P(I)=U_P(I)$. 
\end{theorem}

\noindent {\bf Proof:} Let $A\in N_P(I)$ and $A\leftarrow B_1,...,B_m,
\neg D_1,...,\neg D_n$ be a Herbrand instantiated clause of $P$ for $A$.
By Definition \ref{vp}, either some $\neg B_i$ or $D_j$ is in $M_P(I)$,
or (when $A\leftarrow B_1,...,B_m$ is in $P|M_P(I)$) there exists some $B_i$
such that neither $B_i\in M_P(I)^+$ nor $B_i\in O_P(I)$, i.e. $B_i\in N_P(I)$
(see Lemma \ref{lem2-3}). By Definition \ref{unf-set}, $N_P(I)$ is an 
unfounded set w.r.t. $I$, so $N_P(I)\subseteq U_P(I)$.

Assume, on the contrary, that there is an $A\in U_P(I)$ but $A\not\in N_P(I)$.  
Since $U_P(I)\cap M_P(I)^+=\emptyset$, $A\in O_P(I)$. So there exists
a Herbrand instantiated clause $C$ of $P$

$\quad A\leftarrow B_1,...,B_m,\neg D_1,...,\neg D_n$

\noindent such that $C$ does not satisfy point 1 of Definition \ref{unf-set}
(since $I\subseteq M_P(I)$) and 

$\quad A\leftarrow B_1,...,B_m$

\noindent is in $P|M_P(I)$ where each $B_i$ is either in $M_P(I)^+$
or in $O_P(I)$. Since $A\in U_P(I)$, by point 2 
of Definition \ref{unf-set} some $B_j\in U_P(I)$ and thus $B_j\in O_P(I)$.

Repeating the above process leads to an infinite chain: the proof
of $A$ needs the proof of $B_j^1$ that needs the proof of $B_j^2$, and so on,
where each $B_j^i\in O_P(I)$. Obviously, for no $B_j^i$ along the chain 
its proof can be reduced to a set of ground negative literals 
$\neg E_j$s where neither $E_j$ nor $\neg E_j$ is in $M_P(I)$. This
contradicts Lemma \ref{lem2-4}, so $U_P(I)\subseteq N_P(I)$.  $\Box$

\vspace{4mm}

Starting with $I=\emptyset$, we compute $M_P(I)$, followed by 
$O_P(I)$ and $N_P(I)$. By Lemma \ref{lem2-2} and Theorem \ref{wfs-1},
each $A\in M_P(I)^+$ (resp. $A\in N_P(I)$) is true (resp. false) 
under the well-founded semantics. $O_P(I)$ is a set of {\em temporarily 
undefined} ground literals whose truth values cannot be determined at 
this stage of transformations based on $I$. We then do iterative
computations by letting $I=M_P(I)\cup \neg .N_P(I)$ until we reach 
a fixpoint. This forms the basis on which our operational procedure is
designed for top-down computation of the well-founded semantics.

\section{SLT-Trees and SLT-Resolution}

In this section, we define SLT-trees and SLT-resolution.
Here ``SLT'' stands for ``{\em L}inear {\em T}abulated resolution using
a {\em S}election/computation rule.'' 

Recall the familiar notion of a {\em tree} for describing 
the search space of a top-down proof procedure. For convenience,
a node in such a tree is represented by $N_i:G_i$, where $N_i$ is the 
node name and $G_i$ is a goal labeling the node. Assume no two 
nodes have the same name. Therefore, we can refer to nodes by their names.
 
\begin{definition}[\cite{shen97} with slight modification] 
\label{alist} 
{\em 
An {\it ancestor list} $AL_A$ of pairs $(N_i, A_i)$,
where $N_i$ is a node name and $A_i$ is an atom, is 
associated with each subgoal $A$ in a tree, which 
is defined recursively as follows. 
\begin{enumerate} 
\item 
If $A$ is at the root, then $AL_A=\emptyset$ unless otherwise specified. 

\item
Let $A$ be at node $N_{i+1}$ and $N_i$ be its parent node. 
If $A$ is copied or instantiated from some subgoal $A'$ at $N_i$ then  
$AL_A=AL_{A'}$.

\item
Let $N_i:G_i$ be a node that contains a positive literal $B$. Let $A$ be 
at node $N_{i+1}$ that is obtained from $N_i$ by resolving $G_i$ against
a clause $B' \leftarrow L_1, ..., L_n$ on the literal $B$ with
an mgu $\theta$. If $A$ is $L_j \theta$ for some $1 \leq j \leq n$,
then $AL_A = \{(N_i,B)\} \cup AL_B$. 
\end{enumerate} 
} 
\end{definition} 

Apparently, for any subgoals $A$ and $B$ if $A$ is in the ancestor list
of $B$, i.e. $(\_,A) \in AL_B$, the proof of $A$ needs the proof of $B$.
Particularly, if $(\_,A) \in AL_B$ and $B$ is a variant of $A$, the derivation
goes into a loop. This leads to the following.

\begin{definition}
\label{loop-def}
{\em 
Let $R$ be a computation rule and 
$A_i$ and $A_k$ be two subgoals that are selected by $R$ at nodes $N_i$ and $N_k$, 
respectively. If $(N_i,A_i) \in AL_{A_k}$, $A_i$ (resp. $N_i$) is called an  
{\em ancestor subgoal} of $A_k$ (resp. {\em an ancestor node} of $N_k$).  
If $A_i$ is both an ancestor subgoal and a variant,
i.e. an {\em ancestor variant subgoal}, of $A_k$,
we say the derivation goes into a {\em loop}, where $N_k$ and
all its ancestor nodes involved in the loop are called {\em loop nodes}
and the clause used by $A_i$ to generate this loop
is called a {\em looping clause} of $A_k$ w.r.t. $A_i$. 
We say a node is {\em loop-dependent} if it is a loop node or an ancestor node
of some loop node. Nodes that are not loop-dependent are  
{\em loop-independent}.
} 
\end{definition}

In tabulated resolutions, intermediate positive and negative 
(or alternatively, undefined) answers of some subgoals will be stored 
in tables at some stages. Such answers are called {\em tabled answers}.
Let $TB_f$ be a table that stores some ground negative answers; i.e.  
for each $A\in TB_f$ $\neg A\in WF(P)$. In addition, we introduce a
special subgoal, $u^*$, which is assumed to occur in neither
programs nor top goals. $u^*$ will be used to substitute for
some ground negative subgoals whose truth values are 
temporarily undefined. We now define SLT-trees.

\begin{definition}[SLT-trees]
\label{slt-tree}
{\em
Let $P$ be a program, $G_0$ a top goal, and $R$ a computation rule.
Let $TB_f$ be a set of ground atoms such that for each $A\in TB_f$ $\neg A\in WF(P)$.
The {\em SLT-tree} $T_{G_0}$ for $(P \cup \{G_0\},TB_f)$ via $R$ is a tree 
rooted at node $N_0:G_0$ such that for any node $N_i:G_i$ in the tree
with $G_i=\leftarrow L_1,...,L_n$:

\begin{enumerate}
\item
If $n=0$ then $N_i$ is a {\em success} leaf, marked by $\Box_t$.

\item
If $L_1=u^*$ then $N_i$ is a {\em temporarily undefined} 
leaf, marked by $\Box_{u^*}$.

\item
Let $L_j$ be a positive literal selected by $R$. Let $C_{L_j}$ be the set
of clauses in $P$ whose heads unify with $L_j$ and $LC_{L_j}$ be the set of
looping clauses of $L_j$ w.r.t. its ancestor variant subgoals. If 
$C_{L_j}-LC_{L_j}=\emptyset$ then $N_i$ is a {\em failure} leaf, marked by $\Box_f$;
else the children of $N_i$ are obtained by resolving 
$G_i$ with each of the clauses in $C_{L_j}-LC_{L_j}$ over the literal $L_j$.

\item
Let $L_j=\neg A$ be a negative literal selected by $R$. If $A$ is not ground then
$N_i$ is a {\em flounder} leaf, marked by $\Box_{fl}$; else if 
$A$ is in $TB_f$ then $N_i$ has only one child that is labeled
by the goal $\leftarrow L_1,...,L_{j-1},L_{j+1},...,L_n$; else build an SLT-tree
$T_{\leftarrow A}$ for $(P \cup \{\leftarrow A\},TB_f)$ via $R$, where 
the subgoal $A$ at the root inherits the ancestor
list $AL_{L_j}$ of $L_j$. We consider the following cases:

\begin{enumerate}
\item
If $T_{\leftarrow A}$ has a success leaf then 
$N_i$ is a {\em failure} leaf, marked by $\Box_f$;

\item
If $T_{\leftarrow A}$ has no success leaf but a flounder leaf then
$N_i$ is a {\em flounder} leaf, marked by $\Box_{fl}$;

\item
Otherwise, $N_i$ has only one child that is labeled
by the goal $\leftarrow L_1,...,L_{j-1},L_{j+1},...,$ $L_n,u^*$ if $L_n\neq u^*$
or $\leftarrow L_1,...,L_{j-1},L_{j+1},...,L_n$ if $L_n= u^*$. 
\end{enumerate}
\end{enumerate} 
}
\end{definition}

In an SLT-tree, there may be four types of leaves: success leaves $\Box_t$,
failure leaves $\Box_f$, temporarily undefined 
leaves $\Box_{u^*}$, and flounder leaves $\Box_{fl}$. These leaves
respectively represent successful, failed, (temporarily) undefined,
and floundering derivations (see Definition \ref{branch}).
In this paper, we shall not discuss floundering $-$ a situation where
a non-ground negative literal is selected by a computation rule $R$
(see \cite{chan88, dra95, LAC99, Prz89-3} for discussion on such topic). Therefore,
in the sequel we assume that no SLT-trees contain flounder leaves.
 
The construction of SLT-trees can be viewed as that of SLDNF-trees 
\cite{clark78, Ld87}
enhanced with the following loop-handling mechanisms: (1) Loops are detected
using ancestor lists of subgoals. Positive loops occur within SLT-trees, whereas
negative loops (i.e. loops through negation) 
occur across SLT-trees (see point 4 of Definition \ref{slt-tree}, where the {\em child}
SLT-tree $T_{\leftarrow A}$ is connected to its parent SLT-tree by letting
$A$ at the root of $T_{\leftarrow A}$ inherit the ancestor
list $AL_{L_j}$ of $L_j$). (2) Loops are broken by disallowing
subgoals to use looping clauses for node expansion 
(see point 3 of Definition \ref{slt-tree}). 
This guarantees that SLT-trees are
finite (see Theorem \ref{tree-finite}). (3) Due to the exclusion of 
looping clauses, some answers may be missed in an SLT-tree. 
Therefore, for any ground negative
subgoal $\neg A$ its answer (true or false) can be definitely 
determined only when $A$ is given to be false (i.e. $A\in TB_f$) or
the proof of $A$ via the SLT-tree $T_{\leftarrow A}$ succeeds (i.e.
$T_{\leftarrow A}$ has a success leaf). Otherwise, $\neg A$ is assumed to be
temporarily undefined and is replaced by $u^*$ (see point 4 of Definition 
\ref{slt-tree}). Note that $u^*$ is only introduced to signify the existence
of subgoals whose truth values are temporarily undefined. Therefore, keeping
one $u^*$ in a goal is enough for such a purpose (see point 4 (c)). From 
point 2 of Definition \ref{slt-tree} we see that goals
with a subgoal $u^*$ cannot lead to a success leaf. However,
they may arrive at a failure leaf if one of the remaining subgoals fails.    
 
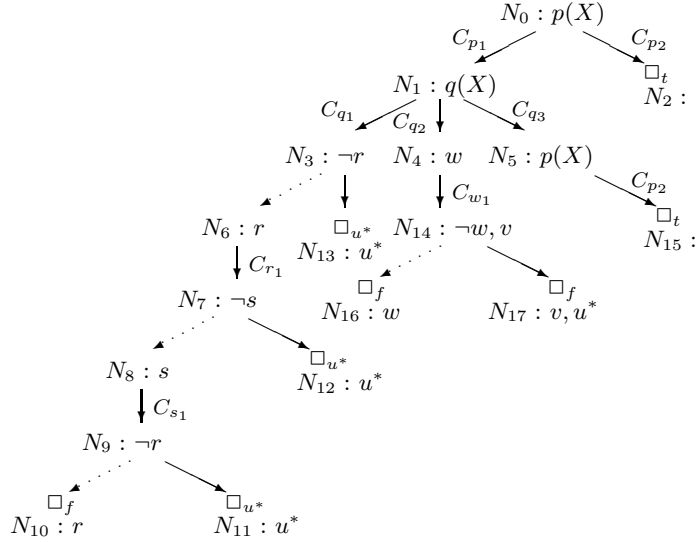
\begin{figure}[htb] 
\centering

\setlength{\unitlength}{3947sp}%
\begingroup\makeatletter\ifx\SetFigFont\undefined%
\gdef\SetFigFont#1#2#3#4#5{%
  \reset@font\fontsize{#1}{#2pt}%
  \fontfamily{#3}\fontseries{#4}\fontshape{#5}%
  \selectfont}%
\fi\endgroup%
\begin{picture}(4062,3321)(1051,-3286)
\thinlines
\put(3601,-561){\vector(-2,-1){380}}
\put(3901,-561){\vector( 2,-1){380}}
\put(3751,-561){\vector( 0,-1){200}}
\put(4351,-136){\vector(-2,-1){380}}
\put(4651,-136){\vector( 2,-1){380}}
\put(2621,-1226){\vector(-2,-1){0}}
\multiput(3001,-1036)(-63.33333,-31.66667){6}{\makebox(1.6667,11.6667){\SetFigFont{5}{6}{\rmdefault}{\mddefault}{\updefault}.}}
\put(3751,-1036){\vector( 0,-1){200}}
\put(2476,-1486){\vector( 0,-1){200}}
\put(1946,-2126){\vector(-2,-1){0}}
\multiput(2326,-1936)(-63.33333,-31.66667){6}{\makebox(1.6667,11.6667){\SetFigFont{5}{6}{\rmdefault}{\mddefault}{\updefault}.}}
\put(1876,-2386){\vector( 0,-1){200}}
\put(2026,-2836){\vector( 2,-1){380}}
\put(4726,-1036){\vector( 2,-1){380}}
\put(4051,-1486){\vector( 2,-1){380}}
\put(2551,-1936){\vector( 2,-1){380}}
\put(3151,-1036){\vector( 0,-1){200}}
\put(3371,-1676){\vector(-2,-1){0}}
\multiput(3751,-1486)(-63.33333,-31.66667){6}{\makebox(1.6667,11.6667){\SetFigFont{5}{6}{\rmdefault}{\mddefault}{\updefault}.}}
\put(1421,-3026){\vector(-2,-1){0}}
\multiput(1801,-2836)(-63.33333,-31.66667){6}{\makebox(1.6667,11.6667){\SetFigFont{5}{6}{\rmdefault}{\mddefault}{\updefault}.}}
\put(2776,-961){\makebox(0,0)[lb]{\smash{\SetFigFont{9}{10.8}{\rmdefault}{\mddefault}{\updefault}$N_3:$ $\neg r$ }}}
\put(3451,-961){\makebox(0,0)[lb]{\smash{\SetFigFont{9}{10.8}{\rmdefault}{\mddefault}{\updefault}$N_4:$ $w$ }}}
\put(4051,-961){\makebox(0,0)[lb]{\smash{\SetFigFont{9}{10.8}{\rmdefault}{\mddefault}{\updefault}$N_5:$ $p(X)$ }}}
\put(3451,-736){\makebox(0,0)[lb]{\smash{\SetFigFont{8}{9.6}{\rmdefault}{\mddefault}{\updefault}$C_{q_2}$}}}
\put(4201,-661){\makebox(0,0)[lb]{\smash{\SetFigFont{8}{9.6}{\rmdefault}{\mddefault}{\updefault}$C_{q_3}$}}}
\put(4951,-211){\makebox(0,0)[lb]{\smash{\SetFigFont{8}{9.6}{\rmdefault}{\mddefault}{\updefault}$C_{p_2}$}}}
\put(3826,-211){\makebox(0,0)[lb]{\smash{\SetFigFont{8}{9.6}{\rmdefault}{\mddefault}{\updefault}$C_{p_1}$}}}
\put(3001,-661){\makebox(0,0)[lb]{\smash{\SetFigFont{8}{9.6}{\rmdefault}{\mddefault}{\updefault}$C_{q_1}$}}}
\put(3451,-511){\makebox(0,0)[lb]{\smash{\SetFigFont{9}{10.8}{\rmdefault}{\mddefault}{\updefault}$N_1:$  $q(X)$}}}
\put(2251,-1411){\makebox(0,0)[lb]{\smash{\SetFigFont{9}{10.8}{\rmdefault}{\mddefault}{\updefault}$N_6:$ $ r$ }}}
\put(3826,-1186){\makebox(0,0)[lb]{\smash{\SetFigFont{8}{9.6}{\rmdefault}{\mddefault}{\updefault}$C_{w_1}$}}}
\put(3451,-1411){\makebox(0,0)[lb]{\smash{\SetFigFont{9}{10.8}{\rmdefault}{\mddefault}{\updefault}$N_{14}:$ $\neg w,v$ }}}
\put(2551,-1636){\makebox(0,0)[lb]{\smash{\SetFigFont{8}{9.6}{\rmdefault}{\mddefault}{\updefault}$C_{r_1}$}}}
\put(2101,-1861){\makebox(0,0)[lb]{\smash{\SetFigFont{9}{10.8}{\rmdefault}{\mddefault}{\updefault}$N_7:$ $\neg s$ }}}
\put(1651,-2311){\makebox(0,0)[lb]{\smash{\SetFigFont{9}{10.8}{\rmdefault}{\mddefault}{\updefault}$N_8:$ $s$ }}}
\put(1951,-2536){\makebox(0,0)[lb]{\smash{\SetFigFont{8}{9.6}{\rmdefault}{\mddefault}{\updefault}$C_{s_1}$}}}
\put(1501,-2761){\makebox(0,0)[lb]{\smash{\SetFigFont{9}{10.8}{\rmdefault}{\mddefault}{\updefault}$N_9:$ $\neg r$ }}}
\put(4126,-61){\makebox(0,0)[lb]{\smash{\SetFigFont{9}{10.8}{\rmdefault}{\mddefault}{\updefault}$N_0:$ $p(X)$ }}}
\put(4951,-1111){\makebox(0,0)[lb]{\smash{\SetFigFont{8}{9.6}{\rmdefault}{\mddefault}{\updefault}$C_{p_2}$}}}
\put(3226,-1786){\makebox(0,0)[lb]{\smash{\SetFigFont{9}{10.8}{\rmdefault}{\mddefault}{\updefault}$\Box_f$}}}
\put(3001,-1936){\makebox(0,0)[lb]{\smash{\SetFigFont{9}{10.8}{\rmdefault}{\mddefault}{\updefault}$N_{16}:$ $w$ }}}
\put(4051,-1936){\makebox(0,0)[lb]{\smash{\SetFigFont{9}{10.8}{\rmdefault}{\mddefault}{\updefault}$N_{17}:$ $v, u^*$ }}}
\put(1051,-3286){\makebox(0,0)[lb]{\smash{\SetFigFont{9}{10.8}{\rmdefault}{\mddefault}{\updefault}$N_{10}:$ $r$ }}}
\put(5026,-1486){\makebox(0,0)[lb]{\smash{\SetFigFont{9}{10.8}{\rmdefault}{\mddefault}{\updefault}$N_{15}:$}}}
\put(5101,-1336){\makebox(0,0)[lb]{\smash{\SetFigFont{9}{10.8}{\rmdefault}{\mddefault}{\updefault}$\Box_t$}}}
\put(4426,-1786){\makebox(0,0)[lb]{\smash{\SetFigFont{9}{10.8}{\rmdefault}{\mddefault}{\updefault}$\Box_f$}}}
\put(2926,-2236){\makebox(0,0)[lb]{\smash{\SetFigFont{9}{10.8}{\rmdefault}{\mddefault}{\updefault}$\Box_{u^*}$ }}}
\put(2851,-2386){\makebox(0,0)[lb]{\smash{\SetFigFont{9}{10.8}{\rmdefault}{\mddefault}{\updefault}$N_{12}:$ $u^*$ }}}
\put(1276,-3136){\makebox(0,0)[lb]{\smash{\SetFigFont{9}{10.8}{\rmdefault}{\mddefault}{\updefault}$\Box_f$}}}
\put(2401,-3136){\makebox(0,0)[lb]{\smash{\SetFigFont{9}{10.8}{\rmdefault}{\mddefault}{\updefault}$\Box_{u^*}$ }}}
\put(2326,-3286){\makebox(0,0)[lb]{\smash{\SetFigFont{9}{10.8}{\rmdefault}{\mddefault}{\updefault}$N_{11}:$ $u^*$ }}}
\put(5026,-436){\makebox(0,0)[lb]{\smash{\SetFigFont{9}{10.8}{\rmdefault}{\mddefault}{\updefault}$\Box_t$}}}
\put(5026,-586){\makebox(0,0)[lb]{\smash{\SetFigFont{9}{10.8}{\rmdefault}{\mddefault}{\updefault}$N_2:$}}}
\put(2851,-1561){\makebox(0,0)[lb]{\smash{\SetFigFont{9}{10.8}{\rmdefault}{\mddefault}{\updefault}$N_{13}:$ $u^*$ }}}
\put(3076,-1411){\makebox(0,0)[lb]{\smash{\SetFigFont{9}{10.8}{\rmdefault}{\mddefault}{\updefault}$\Box_{u^*}$ }}}
\end{picture}
 
\caption{The generalized SLT-tree $GT_{\leftarrow p(X)}$ for 
$(P_1\cup \{\leftarrow p(X)\},\emptyset)$.}\label{fig2}
\end{figure}

For convenience, we use dotted edges to connect parent and child
SLT-trees, so that negative loops can be clearly identified (see Figure \ref{fig2}).
Moreover, we refer to $T_{G_0}$, the {\em top} SLT-tree, 
along with all its descendant SLT-trees
as a {\em generalized} SLT-tree for $(P\cup \{G_0\},TB_f)$, denoted $GT_{P, G_0}$
(or simply $GT_{G_0}$ when no confusion would occur).
Therefore, a path of a generalized SLT-tree may come 
across several SLT-trees through dotted edges.

\begin{example}
\label{eg3-2}
{\em
Consider the following program and let $G_0=\leftarrow p(X)$ be the top goal.
\begin{tabbing} 
\hspace{.2in} $P_1$: \= $p(X) \leftarrow q(X).$ \`$C_{p_1}$ \\ 
\> $p(a).$ \`$C_{p_2}$ \\ 
\> $q(X) \leftarrow \neg r.$ \`$C_{q_1}$ \\ 
\> $q(X) \leftarrow w.$ \`$C_{q_2}$ \\ 
\> $q(X) \leftarrow p(X).$ \`$C_{q_3}$ \\ 
\> $r \leftarrow \neg s.$ \`$C_{r_1}$ \\ 
\> $s \leftarrow \neg r.$ \`$C_{s_1}$ \\ 
\> $w \leftarrow \neg w,v.$ \`$C_{w_1}$  
\end{tabbing} 
For convenience, let us choose the left-most computation rule and let $TB_f=\emptyset$. 
The generalized SLT-tree $GT_{\leftarrow p(X)}$
for $(P_1\cup \{\leftarrow p(X)\},\emptyset)$ is shown in Figure \ref{fig2},\footnote{
For simplicity, in depicting SLT-trees 
we omit the ``$\leftarrow$'' symbol in goals.} 
which consists of five SLT-trees that are rooted at $N_0$, $N_6$, $N_8$,
$N_{10}$ and $N_{16}$, respectively.
$N_2$ and $N_{15}$ are success leaves because they are labeled by an empty goal.
$N_{10}$, $N_{16}$ and $N_{17}$ are failure leaves because they have
no clauses to unify with except for the looping clauses $C_{r_1}$
(for $N_{10}$) and $C_{w_1}$ (for $N_{16}$). $N_{11}$, $N_{12}$ and
$N_{13}$ are temporarily undefined leaves because their goals consist only
of $u^*$. 
}
\end{example}

SLT-trees have some nice properties. Before proving those properties,
we reproduce the definition of bounded-term-size programs.
The following definition is adapted from \cite{VG89}.

\begin{definition}
\label{bounded-term-size}
{\em
A program has the {\em bounded-term-size} property if there is a function
$f(n)$ such that whenever a top goal $G_0$ has no argument whose term size exceeds
$n$, then no subgoals and tabled answers in any generalized SLT-tree
$GT_{G_0}$ have an argument whose term size exceeds $f(n)$.
}
\end{definition} 

The following result shows that the construction of SLT-trees
is always terminating for programs with the bounded-term-size property.

\begin{theorem}
\label{tree-finite}
Let $P$ be a program with the bounded-term-size property, 
$G_0$ a top goal and $R$ a computation rule. The generalized SLT-tree
$GT_{G_0}$ for $(P \cup \{G_0\},TB_f)$ via $R$ is finite.
\end{theorem}

\noindent {\bf Proof:}
The bounded-term-size property guarantees that no term occurring
on any path of $GT_{G_0}$ can have size greater than
$f(n)$, where $n$ is a bound on the size of terms in the top goal $G_0$.
Assume, on the contrary, that $GT_{G_0}$
is infinite. Then it must have an infinite path because its branching factor
(i.e. the average number of children of all nodes in the tree)
is bounded by the finite number of clauses in $P$. Since $P$ has only a finite
number of predicate, function and constant symbols, some positive subgoal $A_0$
selected by $R$ must have infinitely many variant descendants 
$A_1,A_2,..., A_i,...$ on the path 
such that the proof of $A_0$ needs the proof of $A_1$ that needs the proof of 
$A_2$, and so on. That is, $A_i$ is an ancestor variant subgoal of $A_j$ for any
$0\leq i <j$. Let $P$ have totally $m$ clauses that can unify with $A_0$.
Then by point 3 of Definition \ref{slt-tree}, $A_m$, when selected by $R$, 
will have no clause to unify with except for the $m$ looping clauses. That is, $A_m$
shoud be at a leaf, contradicting that it has variant decendants on the path. $\Box$

\begin{definition}
\label{branch}
{\em
Let $T_{G_0}$ be the SLT-tree for $(P \cup \{G_0\},TB_f)$.
A {\em successful} (resp. {\em failed or undefined}) 
branch of $T_{G_0}$ is a branch that ends at a
success (resp. failure or temporarily undefined) leaf.
A {\em correct answer substitution} for $G_0$ is given by 
$\theta=\theta_1...\theta_n$ where the $\theta_i$s are the most
general unifiers used at each step along a successful 
branch of $T_{G_0}$. 
An {\em SLT-derivation} of $(P \cup \{G_0\},TB_f)$ is a branch of $T_{G_0}$. 
}
\end{definition}

Another principal property of SLT-trees is that
correct answer substitutions for top goals
are sound w.r.t. the well-founded semantics.

\begin{theorem}
\label{ans-subs}
Let $P$ be a program with the bounded-term-size property, 
$G_0=\leftarrow Q_0$ a top goal, and 
$T_{G_0}$ the SLT-tree for $(P\cup\{G_0\},TB_f)$.
For any correct answer substitution $\theta$ for $G_0$ in $T_{G_0}$  
$WF(P) \models \forall (Q_0\theta)$.
\end{theorem}

\noindent {\bf Proof:}
Let $d$ be the depth of a successful branch. Without loss of generality,
assume the branch is of the form

$\quad$ $N_0:G_0\Rightarrow_{\theta_1,C_1}N_1:G_1\Rightarrow_{\theta_2,C_2} ...
\Rightarrow_{\theta_{d-1},C_{d-1}}N_{d-1}:G_{d-1}\Rightarrow_{\theta_d,C_d}\Box_t$

\noindent where $G_i=\leftarrow Q_i$ and $\theta=\theta_1...\theta_d$. 
We show, by induction on $0\leq k<d$, $WF(P) \models \forall 
(Q_k\theta_{k+1}...\theta_d)$.

Let $k=d-1$. Since $N_d$ is a success leaf, 
$G_{d-1}$ has only one literal, say $L$.
If $L$ is positive, $C_d$ must be a bodyless clause in $P$
such that $L\theta_d=C_d\theta_d$. In such a case, 
$WF(P) \models \forall (C_d)$, so that $WF(P) \models \forall (Q_k\theta_d)$.
Otherwise, $L=\neg A$ is a ground negative literal. By point 4 of Definition
\ref{slt-tree} $A\in TB_f$ and thus $WF(P) \models \neg A$. Therefore 
$WF(P) \models \forall (Q_k\theta_d)$ with $\theta_d=\emptyset$.

As induction hypothesis, assume that for $0<k<d$ $WF(P) \models \forall 
(Q_k\theta_{k+1}...\theta_d)$. We now prove $WF(P) \models \forall
(Q_{k-1}\theta_k\theta_{k+1}...\theta_d)$.

Let $G_{k-1}=\leftarrow L_1,...,L_n$
with $L_i$ being the selected literal. If $L_i=\neg A$ is negative,
$A$ must be ground and $A\in TB_f$ (otherwise either $N_{k-1}$ is a
flound leaf or a failure leaf, or $G_k$ contains a subgoal $u^*$
in which case $N_{k-1}$ will never lead to a success leaf). So
$WF(P) \models (L_i\theta_k)$ with $\theta_k=\emptyset$ and
$G_k=\leftarrow L_1,...,L_{i-1},L_{i+1},...,L_n$. 
By induction hypothesis we have

$\quad WF(P) \models \forall (Q_k
\theta_{k+1}...\theta_d) \Longrightarrow$\\
\indent $\quad WF(P) \models \forall ((L_1,...,L_{i-1},L_{i+1},...,L_n)
\theta_{k+1}...\theta_d) \Longrightarrow$\\
\indent $\quad WF(P) \models \forall ((L_1,...,L_{i-1},L_i,L_{i+1},...,L_n)
\theta_k\theta_{k+1}...\theta_d) \Longrightarrow$\\
\indent $\quad WF(P) \models \forall (Q_{k-1}\theta_k
\theta_{k+1}...\theta_d)$.

\noindent Otherwise, $L_i$ is positive. So there is a clause 
$L_i'\leftarrow B_1,...,B_m$ in $P$ with $L_i\theta_k=L_i'\theta_k$.
That is, $G_k=\leftarrow (L_1,...,L_{i-1},B_1,...,B_m,L_{i+1},...,L_n)\theta_k$.
Since $Q_k\theta_{k+1}...\theta_d$ is true in $WF(P)$, 
$(B_1,...,B_m)$ $\theta_k\theta_{k+1}...\theta_d$ is true in $WF(P)$.
So $L_i'\theta_k\theta_{k+1}...\theta_d$ is true in $WF(P)$. Therefore

$\quad WF(P) \models \forall (Q_k
\theta_{k+1}...\theta_d) \Longrightarrow$\\
\indent $\quad WF(P) \models \forall ((L_1,...,L_{i-1}, B_1,...,B_m, L_{i+1},...,L_n)\theta_k
\theta_{k+1}...\theta_d) \Longrightarrow$\\
\indent $\quad WF(P) \models \forall ((L_1,...,L_{i-1},L_i,L_{i+1},...,L_n)
\theta_k\theta_{k+1}...\theta_d) \Longrightarrow$\\
\indent $\quad WF(P) \models \forall (Q_{k-1}\theta_k
\theta_{k+1}...\theta_d)$. $\qquad\qquad\Box$

\vspace{4mm}

SLT-trees provide a basis for us to develop a sound 
and complete method for computing the well-founded semantics.

Observe that the concept of correct 
answer substitutions for a top goal $G_0$, defined in 
Definition \ref{branch}, can be extended to any goal 
$G_i$ at node $N_i$ in a generalized SLT-tree $GT_{G_0}$. 
This is done simply by adding a condition 
that the (sub-) branch starts at $N_i$. For instance, 
in Figure \ref{fig2} the branch that starts at $N_1$
and ends at $N_{15}$ yields a correct answer 
substitution $\theta_1\theta_2$ for the goal
$\leftarrow q(X)$ at $N_1$, where $\theta_1=\{X_1/X\}$ is the mgu of $q(X)$ unifying
with the head of $C_{q_3}$ and $\theta_2=\{X/a\}$ is the mgu of $p(X)$
at $N_5$ unifying with $C_{p_2}$. From the proof of Theorem \ref{ans-subs} 
it is easily seen that it
applies to correct answer substitutions for any goals in $GT_{G_0}$. 

Let $G_i$ be a goal in $GT_{G_0}$ and $L_j$ be the selected subgoal 
in $G_i$. Assume that $L_j$ is positive. The partial branches of
$GT_{G_0}$ that are used to prove $L_j$ constitute {\em sub-derivations}
for $L_j$. By Theorem \ref{ans-subs}, for any correct answer substitution $\theta$ 
built from a successful sub-derivation for $L_j$  
$WF(P) \models \forall (L_j\theta)$. We refer to such intermediate results like
$L_j\theta$ as {\em tabled positive answers}.

Let $TB_t^0$ consist of all tabled positive 
answers in $GT_{G_0}$. Then $P$ is equivalent
to $P^1=P\cup TB_t^0$ w.r.t. the well-founded 
semantics. Due to the addition of
tabled positive answers, a new generalized SLT-tree 
$GT_{G_0}^1$ for $(P^1\cup \{G_0\},TB_f)$ can be built
with possibly more tabled positive answers derived. 
Let $TB_t^1$ consist of all tabled positive 
answers in $GT_{G_0}^1$ but not in $TB_t^0$ 
and $P^2=P^1\cup TB_t^1$. Clearly $P^2$ is equivalent
to $P^1$ w.r.t. the well-founded semantics. 
Repeating this process we will generate a sequence of equivalent programs

$\quad$ $P^1,P^2,...,P^i,...$

\noindent where $P^i=P^{i-1}\cup TB_t^{i-1}$ and 
$TB_t^{i-1}$ consists of all tabled positive 
answers in $GT_{G_0}^{i-1}$ for $(P^{i-1}\cup \{G_0\},TB_f)$ 
but not in $\bigcup_{k=0}^{i-2} TB_t^k$, until we reach a fixpoint. 
This leads to the following useful function.

\begin{definition}
{\em
Let $P$ be a program, $G_0$ a top goal and $R$ a computation rule. Define
\begin{tabbing}
{\bf function} $SLTP(P,G_0,R,TB_t,TB_f)$ {\bf return} 
a generalized SLT-tree $GT_{G_0}$\\
$\ $ \= {\bf begin} \\
\> $\quad$ \= Build a generalized SLT-tree $GT_{G_0}$ 
for $(P \cup \{G_0\},TB_f)$ via $R$;\\
\>\> $NEW_t$ collects all tabled positive 
answers in $GT_{G_0}$ but not in $TB_t$;\\
\>\> {\bf if} \= $NEW_t = \emptyset$ {\bf then} return $GT_{G_0}$\\
\>\> $\qquad$ {\bf else} return $SLTP(P\cup NEW_t,G_0,R,TB_t\cup NEW_t,TB_f)$\\ 
\> {\bf end}
\end{tabbing}
}
\end{definition}  

The following two theorems show that for 
positive programs with the bounded-term-size property, 
the function call $SLTP(P,G_0,R,\emptyset,\emptyset)$
is terminating, and sound and complete w.r.t. the well-founded semantics.
So we call it {\em SLTP-resolution} (i.e. SLT-resolution for 
Positive programs).

\begin{theorem}
\label{termin-positive}
For positive programs with the bounded-term-size property 
SLTP-resolution terminates in finite time.
\end{theorem}

\noindent {\bf Proof:}
The function call $SLTP(P,G_0,$ $R,\emptyset,\emptyset)$ will
generate a sequence of generalized SLT-trees

$\quad$ $GT_{G_0}^0, GT_{G_0}^1,...,GT_{G_0}^i,...$

\noindent where $GT_{G_0}^0$ is the generalized 
SLT-tree for $(P \cup \{G_0\},\emptyset)$ via $R$, 
$GT_{G_0}^1$ is the generalized
SLT-tree for $(P\cup NEW_t^0 \cup \{G_0\},\emptyset)$ via $R$ where
$NEW_t^0$ consists of all tabled positive answers in $GT_{G_0}^0$, and
$GT_{G_0}^i$ is the generalized SLT-tree for 
$(P\cup NEW_t^0\cup NEW_t^1\cup ... \cup NEW_t^{i-1} 
\cup \{G_0\},\emptyset)$ via $R$ where
$NEW_t^{i-1}$ consists of all tabled positive answers in $GT_{G_0}^{i-1}$
but not in $\bigcup_{k=0}^{i-2} NEW_t^k$.
Since by Theorem \ref{tree-finite} the construction of each $GT_{G_0}^i$ is
terminating, it suffices to prove that there exists an $i\geq 0$
such that $NEW_t^i=\emptyset$.

Since $P$ has the bounded-term-size property and has only
a finite number of clauses, we have only a finite number of 
subgoals in all generalized SLT-trees $GT_{G_0}^i$s and
any subgoal has only a finite number of positive answers (up to variable 
renaming). Let $N$ be the number of all positive
answers of all subgoals in all $GT_{G_0}^i$s.
Since before the fixpoint is reached, from each $GT_{G_0}^i$ to $GT_{G_0}^{i+1}$
at least one new tabled positive answer to some subgoal will be derived, 
there must exist an $i\leq N+1$ such that $NEW_t^i=\emptyset$. $\Box$

\begin{theorem}
\label{sound-comp-positive}
Let $P$ be a positive program with the bounded-term-size property and 
$G_0\leftarrow Q_0$ a top goal. Let $GT_{G_0}$
be the generalized SLT-tree returned by
$SLTP(P,G_0,$ $R,\emptyset,\emptyset)$. For any (Herbrand) ground instance 
$Q_0\theta$ of $Q_0$ $WF(P)\models Q_0\theta$ 
if and only if there is a correct answer substitution
$\gamma$ for $G_0$ in $GT_{G_0}$ such that $\theta$ is an instance of 
$\gamma$.
\end{theorem}

The following lemma is required to prove this theorem.

\begin{lemma}
\label{lemma1-positive}
Let $GT_{G_0}^0,...,GT_{G_0}^i,...$ be 
a sequence of generalized SLT-trees generated by 
$SLTP($ $P,G_0,R,\emptyset,TB_f)$. For any $0\leq i <j$, if
$\theta$ is a correct answer substitution for $G_0$ in $GT_{G_0}^i$,
so is it in $GT_{G_0}^j$.
\end{lemma}

\noindent {\bf Proof:}
Assume that $GT_{G_0}^i$ and $GT_{G_0}^j$ are the 
generalized SLT-trees for $(P\cup NEW_t \cup \{G_0\},TB_f)$ and
$(P\cup NEW_t' \cup \{G_0\},TB_f)$, respectively. Then
$NEW_t\subseteq NEW_t'$. Let 

$\quad$ $N_0:G_0\Rightarrow_{\theta_1,C_1}N_1:G_1\Rightarrow_{\theta_2,C_2} ...
\Rightarrow_{\theta_{d-1},C_{d-1}}N_{d-1}:G_{d-1}\Rightarrow_{\theta_d,C_d}\Box_t$

\noindent be a successful branch in $GT_{G_0}^i$. At each derivation 
step $N_{k-1}:G_{k-1}\Rightarrow_{\theta_k,C_k}N_k:G_k$, let $L$ be 
the selected literal in $G_{k-1}$. If $L$ is a positive literal, $C_k$ 
is either a clause in $P$ or a tabled positive answer in $NEW_t$; i.e. 
$C_k\in P\cup NEW_t$ and thus $C_k\in P\cup NEW_t'$. So 
$N_{k-1}:G_{k-1}\Rightarrow_{\theta_k,C_k}N_k:G_k$ must be in $GT_{G_0}^j$.
Otherwise, $L=\neg A$ is a ground negative literal. In this case $A\in TB_f$ 
(otherwise either $N_{k-1}$ is a failure leaf or $G_k$ contains a subgoal $u^*$
in which case $N_{k-1}$ will never lead to a success leaf)
and thus $N_{k-1}:G_{k-1}\Rightarrow_{\theta_k,
C_k}N_k:G_k$ must be in $GT_{G_0}^j$ as well, where $\theta_k=\emptyset$
and $C_k=\neg A$. Therefore, the above 
successful branch will appear in $GT_{G_0}^j$. $\Box$

\vspace{4mm}

\noindent {\bf Proof of Theorem \ref{sound-comp-positive}:} 
$(\Longleftarrow)$ 
The function call $SLTP(P,G_0,$ $R,\emptyset,\emptyset)$ will
generate a sequence of generalized SLT-trees

$\quad$ $GT_{G_0}^0, GT_{G_0}^1,...,GT_{G_0}^k=GT_{G_0}$

\noindent where $GT_{G_0}^0$ is the generalized 
SLT-tree for $(P \cup \{G_0\},\emptyset)$, $GT_{G_0}^1$ is the generalized
SLT-tree for $(P^1 \cup \{G_0\},\emptyset)$ with $P^1=P\cup NEW_t^0$, and
$GT_{G_0}$ is the generalized SLT-tree for
$(P^k\cup \{G_0\},\emptyset)$ with $P^k=P^{k-1}\cup NEW_t^{k-1}$ where 
$NEW_t^{k-1}$ is all tabled positive answers in $GT_{G_0}^{k-1}$
but not in $\bigcup_{i=0}^{k-2} NEW_t^i$. Since $NEW_t^i$s are
sets of tabled positive answers, $P$ is equivalent to $P^1$ that
is equivalent to $P^2$ that ... that is equivalent to $P^k$
under the well-founded semantics. By Theorem \ref{ans-subs},
for any correct answer substitution $\gamma$ for $G_0$ in $GT_{G_0}$
$WF(P^k)\models \forall (Q_0\gamma)$ and thus
$WF(P)\models \forall (Q_0\gamma)$.

$(\Longrightarrow)$ Assume $WF(P)\models Q_0\theta$.
By the definition of the well-founded semantics, there
must be a $\gamma$ more general than $\theta$ such that $Q_0\gamma$ 
can be derived by iteratively applying some clauses in $P$. That is, 
we have a backward chain of the form
\begin{equation}
\leftarrow Q_0\Rightarrow_{\theta_1,C_1}\leftarrow Q_1\Rightarrow_{\theta_2,C_2} ...
\Rightarrow_{\theta_{d-1},C_{d-1}}\leftarrow Q_{d-1}\Rightarrow_{\theta_d,C_d}\Box
\qquad\qquad\qquad \label{chain1}
\end{equation}
where $\gamma=\theta_1...\theta_d$ and the $C_i$s are in $P$. 
We consider two cases.

Case 1: There is no loop or there are loops in (\ref{chain1})
but no looping clauses are used. By Definition \ref{slt-tree}
$GT_{G_0}^0$ must have a successful branch corresponding to (\ref{chain1}).
By Lemma \ref{lemma1-positive} $GT_{G_0}$ contains such a branch, too.

\begin{figure}[htb]

\setlength{\unitlength}{3947sp}%
\begingroup\makeatletter\ifx\SetFigFont\undefined%
\gdef\SetFigFont#1#2#3#4#5{%
  \reset@font\fontsize{#1}{#2pt}%
  \fontfamily{#3}\fontseries{#4}\fontshape{#5}%
  \selectfont}%
\fi\endgroup%
\begin{picture}(3000,4095)(301,-4036)
\thinlines
\put(3226,-1186){\vector( 0,-1){225}}
\put(3226,-586){\vector( 0,-1){225}}
\put(3226,-1861){\vector( 0,-1){225}}
\thicklines
\multiput(3226,-166)(0.00000,-60.00000){3}{\makebox(6.6667,10.0000){\SetFigFont{10}{12}{\rmdefault}{\mddefault}{\updefault}.}}
\multiput(3226,-1441)(0.00000,-60.00000){3}{\makebox(6.6667,10.0000){\SetFigFont{10}{12}{\rmdefault}{\mddefault}{\updefault}.}}
\multiput(3226,-841)(0.00000,-60.00000){3}{\makebox(6.6667,10.0000){\SetFigFont{10}{12}{\rmdefault}{\mddefault}{\updefault}.}}
\multiput(3226,-2116)(0.00000,-60.00000){3}{\makebox(6.6667,10.0000){\SetFigFont{10}{12}{\rmdefault}{\mddefault}{\updefault}.}}
\multiput(3226,-2641)(0.00000,-60.00000){3}{\makebox(6.6667,10.0000){\SetFigFont{10}{12}{\rmdefault}{\mddefault}{\updefault}.}}
\multiput(3226,-3166)(0.00000,-60.00000){3}{\makebox(6.6667,10.0000){\SetFigFont{10}{12}{\rmdefault}{\mddefault}{\updefault}.}}
\multiput(3226,-3691)(0.00000,-60.00000){3}{\makebox(6.6667,10.0000){\SetFigFont{10}{12}{\rmdefault}{\mddefault}{\updefault}.}}
\put(2926,-61){\makebox(0,0)[lb]{\smash{\SetFigFont{9}{10.8}{\rmdefault}{\mddefault}{\updefault}$N_0:$ $\leftarrow Q_0$}}}
\put(3301,-1336){\makebox(0,0)[lb]{\smash{\SetFigFont{7}{8.4}{\rmdefault}{\mddefault}{\updefault}$C_{p_j}$}}}
\put(1576,-511){\makebox(0,0)[lb]{\smash{\SetFigFont{9}{10.8}{\rmdefault}{\mddefault}{\updefault}$\qquad\qquad\qquad\quad$ $N_{l_0}:$ $\leftarrow p(\vec{X_0}),B_0$}}}
\put(3301,-736){\makebox(0,0)[lb]{\smash{\SetFigFont{7}{8.4}{\rmdefault}{\mddefault}{\updefault}$C_{p_j}$}}}
\put(3301,-2011){\makebox(0,0)[lb]{\smash{\SetFigFont{7}{8.4}{\rmdefault}{\mddefault}{\updefault}$C_{p_j}$}}}
\put(301,-1786){\makebox(0,0)[lb]{\smash{\SetFigFont{9}{10.8}{\rmdefault}{\mddefault}{\updefault}$\qquad\qquad\qquad\quad$ $N_{l_m}:$ $\leftarrow p(\vec{X}_m),B_m,B_{m-1}\theta_{m-1},...,B_1\theta_{m-1}...\theta_1,B_0\theta_{m-1}...\theta_0$}}}
\put(301,-2461){\makebox(0,0)[lb]{\smash{\SetFigFont{9}{10.8}{\rmdefault}{\mddefault}{\updefault}$\qquad\qquad\qquad\quad$ $N_{x_m}:$ $\leftarrow B_m\gamma_m,B_{m-1}\gamma_m\theta_{m-1},...,B_1\gamma_m\theta_{m-1}...\theta_1,B_0\gamma_m\theta_{m-1}...\theta_0$}}}
\put(526,-2986){\makebox(0,0)[lb]{\smash{\SetFigFont{9}{10.8}{\rmdefault}{\mddefault}{\updefault}$\qquad\qquad\qquad\quad$ $N_{x_1}:$ $\leftarrow B_1\gamma_m...\gamma_1\theta_{m-1}...\theta_1,B_0\gamma_m...\gamma_1\theta_{m-1}...\theta_0$}}}
\put(1351,-1111){\makebox(0,0)[lb]{\smash{\SetFigFont{9}{10.8}{\rmdefault}{\mddefault}{\updefault}$\qquad\qquad\qquad\quad$ $N_{l_1}:$ $\leftarrow p(\vec{X_1}),B_1,B_0\theta_0$}}}
\put(1201,-3511){\makebox(0,0)[lb]{\smash{\SetFigFont{9}{10.8}{\rmdefault}{\mddefault}{\updefault}$\qquad\qquad\qquad\quad$ $N_{x_0}:$ $\leftarrow B_0\gamma_m...\gamma_0\theta_{m-1}...\theta_0$}}}
\put(2851,-4036){\makebox(0,0)[lb]{\smash{\SetFigFont{9}{10.8}{\rmdefault}{\mddefault}{\updefault}$N_t:$ $\Box$}}}
\end{picture}

\caption{An SLD-derivation with loops. $\qquad\qquad$}\label{fig-SC-positive}
\end{figure}
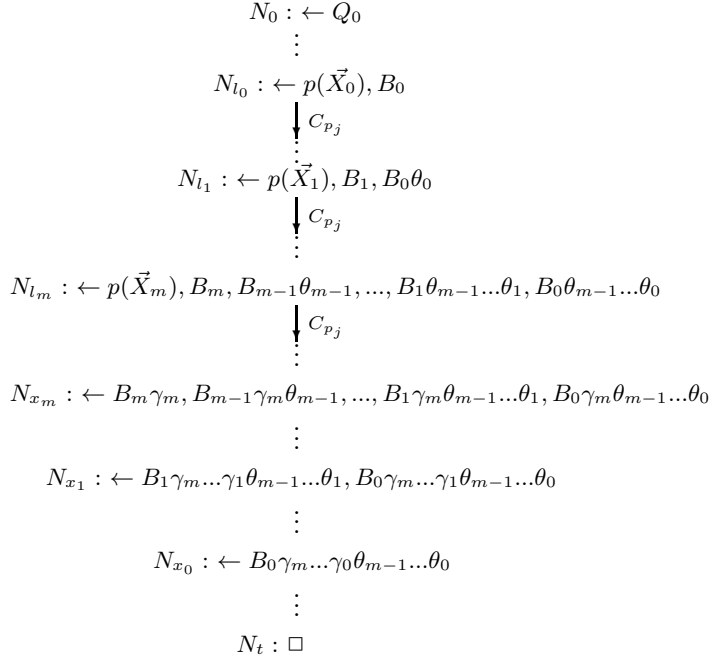

Case 2: There are loops in (\ref{chain1}) with looping clauses applied.
With no loss in generality, assume the backward chain (\ref{chain1}) corresponds to 
the SLD-derivation shown in Figure \ref{fig-SC-positive}, where

\begin{enumerate}
\item[(1)]
The segments between $N_0$ and $N_{l_0}$ and between $N_{x_0}$ and $N_t$
contain no loops. For any $0\leq i <m$ $p(\vec{X_i})$ is an ancestor variant 
subgoal of $p(\vec{X}_{i+1})$. Obviously $C_{p_j}$ is a looping
clause of $p(\vec{X}_{i+1})$ w.r.t. $p(\vec{X_i})$. 

\item[(2)]
For $0\leq i <m$ from $N_{l_i}$ to $N_{l_{i+1}}$ 
the proof of $p(\vec{X_i})$ reduces to 
the proof of $(p(\vec{X}_{i+1}),B_{i+1})$ with
a substitution $\theta_i$ for $p(\vec{X_i})$, where
each $B_k$ $(0\leq k \leq m)$ is a set of subgoals. 

\item[(3)]
The sub-derivation between $N_{l_m}$ and $N_{x_m}$ 
contains no loops and yields an answer
$p(\vec{X}_m)\gamma_m$ to $p(\vec{X}_m)$. The correct answer substitution
$\gamma_m$ for $p(\vec{X}_m)$ is then applied to the remaining
subgoals of $N_{l_m}$ (see node $N_{x_m}$), which leads to an answer
$p(\vec{X}_{m-1})\gamma_m\gamma_{m-1}\theta_{m-1}$ to $p(\vec{X}_{m-1})$. 
Such process continues recursively until an answer
$p(\vec{X_0})\gamma_m...\gamma_0\theta_{m-1}...\theta_0$ to $p(\vec{X_0})$
is produced at $N_{x_0}$.
\end{enumerate}

Since $C_{p_j}$ is a looping clause, the branch below $N_{l_1}$ via $C_{p_j}$
will not occur in any SLT-trees. We first prove that a variant of the answer
$p(\vec{X_0})\gamma_m...\gamma_0\theta_{m-1}...\theta_0$ 
to $p(\vec{X_0})$ will be derived and used as a tabled positive
answer by SLTP-resolution.

Since $p(\vec{X}_0)$ and $p(\vec{X}_m)$ are variants, the sub-derivation between
$N_{l_m}$ and $N_{x_m}$ will appear directly below $N_{l_0}$
via $C_{p_j}$ in $GT_{G_0}^0$, without going through $N_{l_1}$. 
Thus a variant of the answer $p(\vec{X}_m)\gamma_m$ to $p(\vec{X}_m)$ 
will be derived and added to $NEW_t^0$. 

Since $p(\vec{X}_0)$ and $p(\vec{X}_{m-1})$ are variants, 
the sub-derivation between $N_{l_{m-1}}$ and $N_{x_{m-1}}$, where
the sub-derivation between $N_{l_m}$ and $N_{x_m}$ is replaced by directly 
using the tabled positive answer $p(\vec{X}_m)\gamma_m$ in $NEW_t^0$, 
will appear directly below $N_{l_0}$
via $C_{p_j}$ in $GT_{G_0}^1$, without going through $N_{l_1}$.
Thus  a variant of the answer
$p(\vec{X}_{m-1})\gamma_m\gamma_{m-1}\theta_{m-1}$ to $p(\vec{X}_{m-1})$
will be derived and added to $NEW_t^1$.

Continue the above process iteratively. 
After $n$ $(n\leq m)$ iterations, a variant of the answer
$p(\vec{X_0})\gamma_m...\gamma_0\theta_{m-1}...\theta_0$ to $p(\vec{X_0})$
will be derived in $GT_{G_0}^n$ and added to $NEW_t^n$.

Since by assumption there is no loop between $N_0$ 
and $N_{l_0}$ and between $N_{x_0}$ and $N_t$, $GT_{G_0}^{n+1}$ must
contain a successful branch corresponding to Figure \ref{fig-SC-positive}
except that the sub-derivation between $N_{l_1}$ and $N_{x_0}$ is
replaced by directly applying the tabled positive answer
$p(\vec{X_0})\gamma_m...\gamma_0\theta_{m-1}...\theta_0$.
This branch has the same correct answer substitution for $G_0$ as Figure
\ref{fig-SC-positive} (up to variable renaming). By Lemma \ref{lemma1-positive} 
$GT_{G_0}$ contains such a branch, too,
so we conclude the proof. $\Box$

\vspace{4mm}

From the above proof it is easily seen that SLTP-resolution exhausts all tabled
positive answers for all selected positive subgoals in $GT_{G_0}$. 
The following result is immediate.

\begin{corollary}
\label{corol1-positive}
Let $P$ be a positive program with the bounded-term-size property,
$G_0$ a top goal, and $GT_{G_0}$ the generalized SLT-tree 
returned by $SLTP(P,G_0,R,\emptyset,\emptyset)$. Let $TB_t$ consist of
all tabled positive answers in $GT_{G_0}$. Then
\begin{enumerate}
\item
Let $A$ be a selected literal at some node in $GT_{G_0}$.
For any (Herbrand) ground instance $A\theta$ of $A$
$WF(P)\models A\theta$ if and only if there is a tabled answer $A'$ in $TB_t$
such that $A\theta$ is an instance of $A'$.

\item
Let $G_i=\leftarrow Q_i$ be a goal in $GT_{G_0}$. For any 
(Herbrand) ground instance $Q_i\theta$ of $Q_i$
$WF(P)\models Q_i\theta$ if and only if there is a correct answer substitution
$\gamma$ for $G_i$ such that $\theta$ is an instance of $\gamma$.
\end{enumerate}
\end{corollary}

For a positive program, the well-founded semantics has a unique
two-valued (minimal) model and the generalized SLT-tree $GT_{G_0}$ 
returned by $SLTP(P,G_0,$ $R,\emptyset,\emptyset)$ contains only
success and failure leaves. So the following result is immediate
to Corollary \ref{corol1-positive}.

\begin{corollary}
\label{corol2-positive}
Let $P$ be a positive program with the bounded-term-size property,
$G_0$ a top goal, and 
$GT_{G_0}$ the generalized SLT-tree
returned by $SLTP(P,G_0,R,\emptyset,\emptyset)$.
For any goal $G_i=\leftarrow Q_i$ at some node $N_i$ in $GT_{G_0}$, if
all branches starting at $N_i$ end with a failure leaf
then $WF(P) \models \neg\exists (Q_i)$.
\end{corollary}
 
Apparently Corollary \ref{corol2-positive} 
does not hold with general logic programs because
their generalized SLT-trees may contain temporarily undefined leaves.
For instance, although $N_{10}$ labeled by $\leftarrow r$ in Figure \ref{fig2}
ends only with a failure leaf, $r$ is not false in $WF(P_1)$ 
because it has another sub-derivation in $GT_{\leftarrow p(X)}$,
$N_6\rightarrow N_7\rightarrow N_{12}$, that ends with a
temporarily undefined leaf. However, it turns out that the ground atom $w$ in
Figure \ref{fig2} is false in $WF(P_1)$ because all its sub-derivations
(i.e., $N_{16}$ and $N_4\rightarrow N_{14}\rightarrow N_{17}$)
end with a failure leaf. This observation is supported by 
the following theorem.

\begin{theorem}
\label{main-th1}
Let $P$ be a program with the bounded-term-size property and 
$GT_{G_0}$ the generalized SLT-tree returned by 
$SLTP(P,G_0,R,\emptyset,\emptyset)$. Let $TB_t$ consist of all 
tabled positive answers in $GT_{G_0}$

\begin{enumerate}

\item
For any selected positive literal $A$ in $GT_{G_0}$, 
$A\theta\in M_P(\emptyset)$ if and only if there 
is a correct answer substitution for
$A$ in $GT_{G_0}$ that is more general than $\theta$
if and only if there is an $A'\in TB_t$
with $A\theta$ as an instance. In particular, when $A$ is ground,
$A\in M_P(\emptyset)$ if and only if $A\in TB_t$.

\item
Let $A$ be a selected ground positive 
literal in $GT_{G_0}$. Let $S$ be the set of
selected subgoals at the leaf nodes of all
sub-derivations for $A$. $A\in N_P(\emptyset)$
if and only if all sub-derivations for $A$ and $S$ 
end with a failure leaf. 
\end{enumerate}
\end{theorem}

\noindent {\bf Proof:} 1. Note that clauses with negative literals in
their bodies do not contribute to deriving positive answers in $M_P(\emptyset)$
(see Definition \ref{trans-ops1}). This is true in
$SLTP(P,G_0,R,\emptyset,\emptyset)$ as well because a selected  
subgoal $\neg B$ either fails (when $B$ succeeds) or  
is temporarily undefined (otherwise). Let $P^+$ be a positive program
obtained from $P$ by removing all clauses with negative literals in their bodies.
Then $M_P(\emptyset)=M_{P^+}(\emptyset)$ and all tabled positive
answers in $GT_{G_0}$ are derived from $P^+\cup \{G_0\}$. Since
$M_{P^+}(\emptyset)$ is the positive part of $WF(P^+)$, we have
\begin{tabbing}
$\qquad A\theta\in M_P(\emptyset)$\= $\Longleftrightarrow 
A\theta\in M_{P^+}(\emptyset)$ \\
\> $\Longleftrightarrow WF(P^+)\models A\theta$ \\
\> $\Longleftrightarrow$ \= (By Corollary \ref{corol1-positive})
there is an answer substitution for $A$ in $GT_{G_0}$ \\
\>\> that is more general than $\theta$\\
\>  $\Longleftrightarrow$ there is an $A'\in TB_t$
with $A\theta$ as an instance.
\end{tabbing}
When $A$ is ground,
\begin{tabbing}
$\qquad A\in M_P(\emptyset)$\= $\Longleftrightarrow$ 
there is an answer substitution for $A$ in $GT_{G_0}$ \\
\> $\Longleftrightarrow$ (By Definition \ref{branch}) 
there is a successful sub-derivation for $A$ in $GT_{G_0}$ \\
\> $\Longleftrightarrow$ $A\in TB_t$.
\end{tabbing}

2. $(\Longleftarrow)$ By point 1 above $A\not\in M_P(\emptyset)$.
Suppose, on the contrary, that $A\in O_P(\emptyset)$.
Then by Definition \ref{vp} there exists a clause $C$ in $P$ of the form

$\qquad A'\leftarrow B_1,...,B_m, \neg D_1,...,\neg D_n$

\noindent such that one of its Herbrand instantiated clauses is of the form

$\qquad A\leftarrow (B_1,...,B_m, \neg D_1,...,\neg D_n)\theta$

\noindent where no $D_i\theta$ is in $M_P(\emptyset)$ and each $B_i\theta$ is
either in $M_P(\emptyset)$ or in $O_P(\emptyset)$. That is, $A$ can be derived
through a backward chain of the form

$\qquad A\Rightarrow_{S_1} B_1\theta,...,B_m\theta,\neg D_1\theta,
...,\neg D_n\theta\Rightarrow_{S_2}
E_1,...,\neg F_k\Rightarrow_{S_3}...\Rightarrow_{S_t}\Box$

\noindent where each step is performed by either 
resolving a ground positive literal
like $B_i\theta$ with an answer in $M_P(\emptyset)$ (if $B_i\theta
\in M_P(\emptyset)$) or
with a Herbrand instantiated clause of $P$ (otherwise),
or removing a negative literal like
$\neg D_i\theta$ where $D_i\theta\not\in M_P(\emptyset)$.

Based on point 1 above, it is easy to construct a sub-derivation for $A$,
using clauses in $P$ and tabled answers in $TB_t$,
that corresponds to the above backward chain. First we have

$\qquad\leftarrow A\Rightarrow_{C,\theta_0}
\leftarrow B_1\theta_0,...,B_m\theta_0,\neg D_1\theta_0,
...,\neg D_n\theta_0$

\noindent where $\theta_0$ is the most general unifier of $A$ and $A'$.
For each $B_i\theta_0$, if
$B_i\theta$ is resolved with a Herbrand instantiated clause of $P$
(resp. with an answer in $M_P(\emptyset)$) then there is a clause in $P$
(resp. a tabled positive answer in $TB_t$) to resolve with 
$B_i\theta_0$. For each $\neg D_i\theta_0$, if $\theta_0=\theta$
then $\neg D_i\theta_0$ is treated as $u^*$. 
As a result, we will generate a sub-derivation for $A$
of the form

$\qquad \leftarrow A\Rightarrow_{C,\theta_0}...\Rightarrow_{C_{i-1},
\theta_{i-1}}\leftarrow L_1,L_2,...,L_k \Rightarrow_{C_i,\theta_i}...
\Rightarrow_{C_l,\theta_l} \Box_{u^*}$

If no looping clause is used along the above sub-derivation for $A$,
this sub-derivation must be in $GT_{G_0}$. Otherwise, without 
loss of generality assume the above sub-derivation is of the form

\noindent $\quad \leftarrow A\Rightarrow_{C,\theta_0}...
\leftarrow L_1,L_2,...,L_k \Rightarrow_{C_i,\theta_i}...
\leftarrow L_1',F_1,...,F_j,(L_2,...,L_k)\gamma\Rightarrow_{C_i,\theta_i'}
...\Rightarrow_{C_l,\theta_l} \Box_{u^*}$ 

\noindent where $L_1$ is an ancestor variant subgoal of $L_1'$
and $L_1'$ is selected to resolve with the looping clause $C_i$.
It is easily seen that this sub-derivation can be shortened by removing
the sub-derivation between $L_1$ and $L_1'$ because if 
$L_1',F_1,...,F_j,(L_2,...,L_k)\gamma$ can be reduced to $\Box_{u^*}$,
so can $L_1,L_2,...,L_k$. Obviously, the shortened sub-derivation
(or its variant form) will appear in $GT_{G_0}$. 
This contradicts that all sub-derivations of $A$ and $S$ in 
$GT_{G_0}$ end with a failure leaf.  

$(\Longrightarrow)$ Assume $A\in N_P(\emptyset)$ but, on the contrary, that
there is a sub-derivation for $A$ in $GT_{G_0}$ that ends with
a temporarily undefined leaf. Let the sub-derivation be of the form

$\qquad \leftarrow A\Rightarrow_{C,\theta_0}...\Rightarrow_{C_{i-1},
\theta_{i-1}}\leftarrow L_1,L_2,...,L_k \Rightarrow_{C_i,\theta_i}...
\Rightarrow_{C_l,\theta_l} \Box_{u^*}$

\noindent where each derivation step is done by either resolving 
a selected positive literal with a clause in $P$ or with a tabled positive 
answer in $TB_t$, or treating a selected negative ground literal 
$\neg F$ as $u^*$ where $F\not\in TB_t$. Since by point 1 of this theorem
$M_P(\emptyset)$ consists of all (Herbrand) ground instances of 
tabled positive answer in $TB_t$, the above
sub-derivation must have a Herbrand instantiated ground instance of the form

$\qquad A\Rightarrow_{S_1} ...\Rightarrow_{S_j} E_1,...,E_m, 
\neg F_1,...,\neg F_n\Rightarrow_{S_{j+1}}...\Rightarrow_{S_t}\Box$

\noindent where each step is performed by either resolving a positive ground
literal with a Herbrand instantiated clause of $P$ or with
an answer in $M_P(\emptyset)$, or removing a negative ground literal 
$\neg F$ where $F\not\in M_P(\emptyset)$. However, by Definition \ref{vp}
the above backward chain implies that $A$ is in $O_P(\emptyset)$, 
contradicting $A\in N_P(\emptyset)$. 

Now assume that $A\in N_P(\emptyset)$ and 
all sub-derivations for $A$ end with a failure leaf,
but, on the contrary, that there is a sub-derivation 
for $B\in S$ in $GT_{G_0}$ that ends with a temporarily 
undefined leaf. Then $B$ must be an ancestor subgoal
of $B$. That is, there must be two sub-derivations
for $B$ in $GT_{G_0}$ of the form

$\qquad \leftarrow B\Rightarrow ...
\leftarrow A, ... \Rightarrow...
\Rightarrow \Box_f \leftarrow B, ...$

$\qquad \leftarrow B\Rightarrow ... 
\Rightarrow \Box_{u^*}$

\noindent The first sub-derivation suggests that
the answers of $A$ depend on $B$. By the first part of
the argument for $(\Longrightarrow)$, the second 
sub-derivation implies $B \not\in N_P(\emptyset)$.
Combining the two leads to $A \not\in N_P(\emptyset)$, 
which contradicts the assumption $A \in N_P(\emptyset)$.
$\Box$

\vspace{4mm}

Theorem \ref{main-th1} is useful, by which the truth value of all 
selected ground negative literals can be determined in an iterative way.
For any selected ground negative literal $\neg A$, if 
all sub-derivations of $A$  and $S$ (defined in Theorem \ref{main-th1})
in $GT_{G_0}$ end with a failure leaf, 
$A$ is called a {\em tabled negative answer}. All tabled negative
answers will be collected in $TB_f$. 

We are now in a position to define SLT-resolution for general 
logic programs.

\begin{definition}[SLT-resolution]
\label{slt}
{\em
Let $P$ be a program, $G_0$ a top goal and $R$ a computation rule. 
{\em SLT-resolution} proves $G_0$ by calling the function 
$SLT(P,G_0,R,\emptyset,\emptyset)$, which is defined as follows:
\begin{tabbing}
{\bf function} $SLT(P,G_0,R,TB_t, TB_f)$ {\bf return} a 
generalized SLT-tree $GT_{G_0}$\\
$\ $ \= {\bf begin} \\
\> $\quad$ \= $GT_{G_0}=SLTP(P,G_0,R,TB_t,TB_f)$; \\
\>\>  $NEW_t$ collects all tabled positive answers in $GT_{G_0}$ 
but not in $TB_t$;\\
\>\>  $NEW_f$ collects all tabled negative answers in $GT_{G_0}$ 
but not in $TB_f$;\\
\>\> {\bf if} \= $NEW_f = \emptyset$ {\bf then} return $GT_{G_0}$\\
\>\> $\qquad$ {\bf else} return $SLT(P\cup NEW_t,G_0,R,TB_t\cup 
NEW_t,TB_f\cup NEW_f)$ \\
\> {\bf end} 
\end{tabbing}
}
\end{definition} 

\begin{definition}
\label{true-false}
{\em
Let $G_0=\leftarrow Q_0$ be a top goal and
$T_{G_0}$ be the top SLT-tree in $GT_{G_0}$ which is returned by 
$SLT(P,G_0,R,\emptyset,\emptyset)$. $G_0$ is {\em true} in $P$ with an
answer $Q_0\theta$ if there is a correct answer substitution for
$G_0$ in $T_{G_0}$ that is more general than $\theta$;
{\em false} in $P$ if all branches of $T_{G_0}$ end with a failure
leaf; {\em undefined} in $P$ if neither $G_0$ is false nor $T_{G_0}$
has successful branches.
}
\end{definition} 

\begin{example}
\label{eg3-3}
{\em
(Cont. of Example \ref{eg3-2}) To evaluate $G_0=\leftarrow p(X)$,
we call $SLT(P_1,G_0,R,\emptyset,\emptyset)$. This
immediately invokes $SLTP(P_1,G_0,R,\emptyset,\emptyset)$, which generates
the generalized SLT-tree $GT_{\leftarrow p(X)}$ for 
$(P_1\cup \{\leftarrow p(X)\},\emptyset)$
as shown in Figure \ref{fig2}. The tabled positive answers in
$GT_{\leftarrow p(X)}$ are then  collected in $NEW_t^0$,
i.e. $NEW_t^0=\{p(a),q(a)\}$. So $P_1^1=P_1\cup NEW_t^0$. (Note that 
the bodyless program clause $C_{p_2}$ can be ignored in $P_1^1$ since it has
become a tabled answer. See Section \ref{sec-dup} for such kind of optimizations). 
The generalized SLT-tree $GT_{\leftarrow p(X)}^1$ for 
$(P_1^1\cup \{\leftarrow p(X)\},\emptyset)$ is then generated, which is like
$GT_{\leftarrow p(X)}$ except that $N_2$ gets a new child node $N_{2'}$ $-$ a
success leaf, by unifying $q(X)$ with the tabled positive answer $q(a)$ 
in $P_1^1$ (see Figure \ref{fig3}).
Clearly, the addition of this success leaf does not yield any new 
tabled positive answers, i.e. $NEW_t^1=\emptyset$. Therefore 
$SLTP(P_1,G_0,R,\emptyset,\emptyset)$ returns $GT_{\leftarrow p(X)}^1$.

\begin{figure}[htb]
\centering

\setlength{\unitlength}{3947sp}%
\begingroup\makeatletter\ifx\SetFigFont\undefined%
\gdef\SetFigFont#1#2#3#4#5{%
  \reset@font\fontsize{#1}{#2pt}%
  \fontfamily{#3}\fontseries{#4}\fontshape{#5}%
  \selectfont}%
\fi\endgroup%
\begin{picture}(4062,3321)(1051,-3286)
\thinlines
\put(3315,-560){\vector(-4,-1){529.412}}
\put(4951,-1111){\makebox(0,0)[lb]{\smash{\SetFigFont{8}{9.6}{\rmdefault}{\mddefault}{\updefault}$p(a)$}}}
\put(2701,-586){\makebox(0,0)[lb]{\smash{\SetFigFont{8}{9.6}{\rmdefault}{\mddefault}{\updefault}$q(a)$}}}
\put(2251,-211){\makebox(0,0)[lb]{\smash{\SetFigFont{8}{9.6}{\rmdefault}{\mddefault}{\updefault}$p(a)$}}}
\put(3601,-561){\vector(-2,-1){380}}
\put(3901,-561){\vector( 2,-1){380}}
\put(3751,-561){\vector( 0,-1){200}}
\put(2621,-1226){\vector(-2,-1){0}}
\multiput(3001,-1036)(-63.33333,-31.66667){6}{\makebox(1.6667,11.6667){\SetFigFont{5}{6}{\rmdefault}{\mddefault}{\updefault}.}}
\put(3751,-1036){\vector( 0,-1){200}}
\put(2476,-1486){\vector( 0,-1){200}}
\put(1946,-2126){\vector(-2,-1){0}}
\multiput(2326,-1936)(-63.33333,-31.66667){6}{\makebox(1.6667,11.6667){\SetFigFont{5}{6}{\rmdefault}{\mddefault}{\updefault}.}}
\put(1876,-2386){\vector( 0,-1){200}}
\put(2026,-2836){\vector( 2,-1){380}}
\put(4726,-1036){\vector( 2,-1){380}}
\put(4051,-1486){\vector( 2,-1){380}}
\put(2551,-1936){\vector( 2,-1){380}}
\put(3151,-1036){\vector( 0,-1){200}}
\put(3371,-1676){\vector(-2,-1){0}}
\multiput(3751,-1486)(-63.33333,-31.66667){6}{\makebox(1.6667,11.6667){\SetFigFont{5}{6}{\rmdefault}{\mddefault}{\updefault}.}}
\put(1421,-3026){\vector(-2,-1){0}}
\multiput(1801,-2836)(-63.33333,-31.66667){6}{\makebox(1.6667,11.6667){\SetFigFont{5}{6}{\rmdefault}{\mddefault}{\updefault}.}}
\put(3301,-136){\vector( 2,-1){380}}
\put(2776,-136){\vector(-2,-1){380}}
\put(3451,-961){\makebox(0,0)[lb]{\smash{\SetFigFont{9}{10.8}{\rmdefault}{\mddefault}{\updefault}$N_4:$ $w$ }}}
\put(4051,-961){\makebox(0,0)[lb]{\smash{\SetFigFont{9}{10.8}{\rmdefault}{\mddefault}{\updefault}$N_5:$ $p(X)$ }}}
\put(2251,-1411){\makebox(0,0)[lb]{\smash{\SetFigFont{9}{10.8}{\rmdefault}{\mddefault}{\updefault}$N_6:$ $ r$ }}}
\put(3826,-1186){\makebox(0,0)[lb]{\smash{\SetFigFont{8}{9.6}{\rmdefault}{\mddefault}{\updefault}$C_{w_1}$}}}
\put(3451,-1411){\makebox(0,0)[lb]{\smash{\SetFigFont{9}{10.8}{\rmdefault}{\mddefault}{\updefault}$N_{14}:$ $\neg w,v$ }}}
\put(2551,-1636){\makebox(0,0)[lb]{\smash{\SetFigFont{8}{9.6}{\rmdefault}{\mddefault}{\updefault}$C_{r_1}$}}}
\put(2101,-1861){\makebox(0,0)[lb]{\smash{\SetFigFont{9}{10.8}{\rmdefault}{\mddefault}{\updefault}$N_7:$ $\neg s$ }}}
\put(1651,-2311){\makebox(0,0)[lb]{\smash{\SetFigFont{9}{10.8}{\rmdefault}{\mddefault}{\updefault}$N_8:$ $s$ }}}
\put(1951,-2536){\makebox(0,0)[lb]{\smash{\SetFigFont{8}{9.6}{\rmdefault}{\mddefault}{\updefault}$C_{s_1}$}}}
\put(1501,-2761){\makebox(0,0)[lb]{\smash{\SetFigFont{9}{10.8}{\rmdefault}{\mddefault}{\updefault}$N_9:$ $\neg r$ }}}
\put(3226,-1786){\makebox(0,0)[lb]{\smash{\SetFigFont{9}{10.8}{\rmdefault}{\mddefault}{\updefault}$\Box_f$}}}
\put(3001,-1936){\makebox(0,0)[lb]{\smash{\SetFigFont{9}{10.8}{\rmdefault}{\mddefault}{\updefault}$N_{16}:$ $w$ }}}
\put(4051,-1936){\makebox(0,0)[lb]{\smash{\SetFigFont{9}{10.8}{\rmdefault}{\mddefault}{\updefault}$N_{17}:$ $v, u^*$ }}}
\put(1051,-3286){\makebox(0,0)[lb]{\smash{\SetFigFont{9}{10.8}{\rmdefault}{\mddefault}{\updefault}$N_{10}:$ $r$ }}}
\put(5026,-1486){\makebox(0,0)[lb]{\smash{\SetFigFont{9}{10.8}{\rmdefault}{\mddefault}{\updefault}$N_{15}:$}}}
\put(5101,-1336){\makebox(0,0)[lb]{\smash{\SetFigFont{9}{10.8}{\rmdefault}{\mddefault}{\updefault}$\Box_t$}}}
\put(4426,-1786){\makebox(0,0)[lb]{\smash{\SetFigFont{9}{10.8}{\rmdefault}{\mddefault}{\updefault}$\Box_f$}}}
\put(2926,-2236){\makebox(0,0)[lb]{\smash{\SetFigFont{9}{10.8}{\rmdefault}{\mddefault}{\updefault}$\Box_{u^*}$ }}}
\put(2851,-2386){\makebox(0,0)[lb]{\smash{\SetFigFont{9}{10.8}{\rmdefault}{\mddefault}{\updefault}$N_{12}:$ $u^*$ }}}
\put(1276,-3136){\makebox(0,0)[lb]{\smash{\SetFigFont{9}{10.8}{\rmdefault}{\mddefault}{\updefault}$\Box_f$}}}
\put(2401,-3136){\makebox(0,0)[lb]{\smash{\SetFigFont{9}{10.8}{\rmdefault}{\mddefault}{\updefault}$\Box_{u^*}$ }}}
\put(2326,-3286){\makebox(0,0)[lb]{\smash{\SetFigFont{9}{10.8}{\rmdefault}{\mddefault}{\updefault}$N_{11}:$ $u^*$ }}}
\put(2851,-1561){\makebox(0,0)[lb]{\smash{\SetFigFont{9}{10.8}{\rmdefault}{\mddefault}{\updefault}$N_{13}:$ $u^*$ }}}
\put(3076,-1411){\makebox(0,0)[lb]{\smash{\SetFigFont{9}{10.8}{\rmdefault}{\mddefault}{\updefault}$\Box_{u^*}$ }}}
\put(3301,-511){\makebox(0,0)[lb]{\smash{\SetFigFont{9}{10.8}{\rmdefault}{\mddefault}{\updefault}$N_2:$  $q(X)$}}}
\put(2776,-961){\makebox(0,0)[lb]{\smash{\SetFigFont{9}{10.8}{\rmdefault}{\mddefault}{\updefault}$N_3:$ $\neg r$ }}}
\put(2551,-811){\makebox(0,0)[lb]{\smash{\SetFigFont{9}{10.8}{\rmdefault}{\mddefault}{\updefault}$\Box_t$}}}
\put(2326,-961){\makebox(0,0)[lb]{\smash{\SetFigFont{9}{10.8}{\rmdefault}{\mddefault}{\updefault}$N_{2'}:$}}}
\put(2701,-61){\makebox(0,0)[lb]{\smash{\SetFigFont{9}{10.8}{\rmdefault}{\mddefault}{\updefault}$N_0:$ $p(X)$ }}}
\put(3526,-211){\makebox(0,0)[lb]{\smash{\SetFigFont{8}{9.6}{\rmdefault}{\mddefault}{\updefault}$C_{p_1}$}}}
\put(2251,-436){\makebox(0,0)[lb]{\smash{\SetFigFont{9}{10.8}{\rmdefault}{\mddefault}{\updefault}$\Box_t$}}}
\put(2026,-586){\makebox(0,0)[lb]{\smash{\SetFigFont{9}{10.8}{\rmdefault}{\mddefault}{\updefault}$N_1:$}}}
\put(3451,-736){\makebox(0,0)[lb]{\smash{\SetFigFont{7}{8.4}{\rmdefault}{\mddefault}{\updefault}$C_{q_2}$}}}
\put(3001,-736){\makebox(0,0)[lb]{\smash{\SetFigFont{7}{8.4}{\rmdefault}{\mddefault}{\updefault}$C_{q_1}$}}}
\put(4201,-661){\makebox(0,0)[lb]{\smash{\SetFigFont{8}{9.6}{\rmdefault}{\mddefault}{\updefault}$C_{q_3}$}}}
\end{picture}

\caption{The generalized SLT-tree $GT_{\leftarrow p(X)}^1$ for 
$(P_1^1\cup \{\leftarrow p(X)\},\emptyset)$. $\qquad\qquad$}\label{fig3}
\end{figure} 

It is easily seen that $GT_{\leftarrow p(X)}^1$ contains one new
tabled negative answer $w$; i.e. $NEW_f^1=\{w\}$ (note 
that $\neg w$ is a selected literal at $N_{14}$ and all sub-derivations 
for $w$ in $GT_{\leftarrow p(X)}^1$ end with a failure leaf).
Let $TB_t^1=NEW_t^0\cup NEW_t^1$ and $TB_f^1=NEW_f^1$.
Since $NEW_f^1\neq\emptyset$, $SLT(P_1\cup TB_t^1,G_0,R, 
TB_t^1,TB_f^1)$ is recursively called, which invokes 
$SLTP(P_1\cup TB_t^1,G_0,R,$ $TB_t^1,TB_f^1)$. This builds
a generalized SLT-tree $GT_{\leftarrow p(X)}^2$ for 
$(P_1^2\cup \{\leftarrow p(X)\},TB_f^1)$
where $P_1^2=P_1\cup TB_t^1$ (see Figure \ref{fig4}). 
Obviously, $GT_{\leftarrow p(X)}^2$ contains neither new tabled
positive answers nor new tabled negative answers. Therefore,
SLT-resolution stops with $GT_{\leftarrow p(X)}^2$ returned.
By Definition \ref{true-false}, $G_0$ is true with an
answer $p(a)$.
}
\end{example}

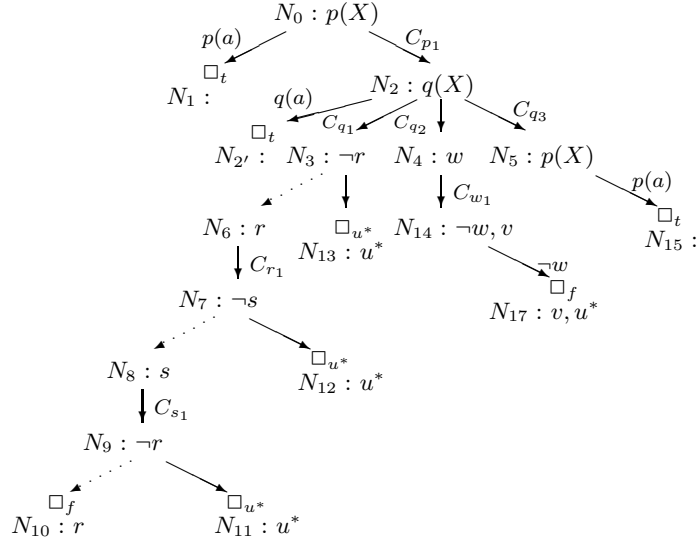
\begin{figure}[htb]
\centering

\setlength{\unitlength}{3947sp}%
\begingroup\makeatletter\ifx\SetFigFont\undefined%
\gdef\SetFigFont#1#2#3#4#5{%
  \reset@font\fontsize{#1}{#2pt}%
  \fontfamily{#3}\fontseries{#4}\fontshape{#5}%
  \selectfont}%
\fi\endgroup%
\begin{picture}(4062,3321)(1051,-3286)
\thinlines
\put(3315,-560){\vector(-4,-1){529.412}}
\put(4951,-1111){\makebox(0,0)[lb]{\smash{\SetFigFont{8}{9.6}{\rmdefault}{\mddefault}{\updefault}$p(a)$}}}
\put(2701,-586){\makebox(0,0)[lb]{\smash{\SetFigFont{8}{9.6}{\rmdefault}{\mddefault}{\updefault}$q(a)$}}}
\put(2251,-211){\makebox(0,0)[lb]{\smash{\SetFigFont{8}{9.6}{\rmdefault}{\mddefault}{\updefault}$p(a)$}}}
\put(4351,-1636){\makebox(0,0)[lb]{\smash{\SetFigFont{8}{9.6}{\rmdefault}{\mddefault}{\updefault}$\neg w$}}}
\put(3601,-561){\vector(-2,-1){380}}
\put(3901,-561){\vector( 2,-1){380}}
\put(3751,-561){\vector( 0,-1){200}}
\put(2621,-1226){\vector(-2,-1){0}}
\multiput(3001,-1036)(-63.33333,-31.66667){6}{\makebox(1.6667,11.6667){\SetFigFont{5}{6}{\rmdefault}{\mddefault}{\updefault}.}}
\put(3751,-1036){\vector( 0,-1){200}}
\put(2476,-1486){\vector( 0,-1){200}}
\put(1946,-2126){\vector(-2,-1){0}}
\multiput(2326,-1936)(-63.33333,-31.66667){6}{\makebox(1.6667,11.6667){\SetFigFont{5}{6}{\rmdefault}{\mddefault}{\updefault}.}}
\put(1876,-2386){\vector( 0,-1){200}}
\put(2026,-2836){\vector( 2,-1){380}}
\put(4726,-1036){\vector( 2,-1){380}}
\put(4051,-1486){\vector( 2,-1){380}}
\put(2551,-1936){\vector( 2,-1){380}}
\put(3151,-1036){\vector( 0,-1){200}}
\put(1421,-3026){\vector(-2,-1){0}}
\multiput(1801,-2836)(-63.33333,-31.66667){6}{\makebox(1.6667,11.6667){\SetFigFont{5}{6}{\rmdefault}{\mddefault}{\updefault}.}}
\put(3301,-136){\vector( 2,-1){380}}
\put(2776,-136){\vector(-2,-1){380}}
\put(3451,-961){\makebox(0,0)[lb]{\smash{\SetFigFont{9}{10.8}{\rmdefault}{\mddefault}{\updefault}$N_4:$ $w$ }}}
\put(4051,-961){\makebox(0,0)[lb]{\smash{\SetFigFont{9}{10.8}{\rmdefault}{\mddefault}{\updefault}$N_5:$ $p(X)$ }}}
\put(2251,-1411){\makebox(0,0)[lb]{\smash{\SetFigFont{9}{10.8}{\rmdefault}{\mddefault}{\updefault}$N_6:$ $ r$ }}}
\put(3826,-1186){\makebox(0,0)[lb]{\smash{\SetFigFont{8}{9.6}{\rmdefault}{\mddefault}{\updefault}$C_{w_1}$}}}
\put(3451,-1411){\makebox(0,0)[lb]{\smash{\SetFigFont{9}{10.8}{\rmdefault}{\mddefault}{\updefault}$N_{14}:$ $\neg w,v$ }}}
\put(2551,-1636){\makebox(0,0)[lb]{\smash{\SetFigFont{8}{9.6}{\rmdefault}{\mddefault}{\updefault}$C_{r_1}$}}}
\put(2101,-1861){\makebox(0,0)[lb]{\smash{\SetFigFont{9}{10.8}{\rmdefault}{\mddefault}{\updefault}$N_7:$ $\neg s$ }}}
\put(1651,-2311){\makebox(0,0)[lb]{\smash{\SetFigFont{9}{10.8}{\rmdefault}{\mddefault}{\updefault}$N_8:$ $s$ }}}
\put(1951,-2536){\makebox(0,0)[lb]{\smash{\SetFigFont{8}{9.6}{\rmdefault}{\mddefault}{\updefault}$C_{s_1}$}}}
\put(1501,-2761){\makebox(0,0)[lb]{\smash{\SetFigFont{9}{10.8}{\rmdefault}{\mddefault}{\updefault}$N_9:$ $\neg r$ }}}
\put(4051,-1936){\makebox(0,0)[lb]{\smash{\SetFigFont{9}{10.8}{\rmdefault}{\mddefault}{\updefault}$N_{17}:$ $v, u^*$ }}}
\put(1051,-3286){\makebox(0,0)[lb]{\smash{\SetFigFont{9}{10.8}{\rmdefault}{\mddefault}{\updefault}$N_{10}:$ $r$ }}}
\put(5026,-1486){\makebox(0,0)[lb]{\smash{\SetFigFont{9}{10.8}{\rmdefault}{\mddefault}{\updefault}$N_{15}:$}}}
\put(5101,-1336){\makebox(0,0)[lb]{\smash{\SetFigFont{9}{10.8}{\rmdefault}{\mddefault}{\updefault}$\Box_t$}}}
\put(4426,-1786){\makebox(0,0)[lb]{\smash{\SetFigFont{9}{10.8}{\rmdefault}{\mddefault}{\updefault}$\Box_f$}}}
\put(2926,-2236){\makebox(0,0)[lb]{\smash{\SetFigFont{9}{10.8}{\rmdefault}{\mddefault}{\updefault}$\Box_{u^*}$ }}}
\put(2851,-2386){\makebox(0,0)[lb]{\smash{\SetFigFont{9}{10.8}{\rmdefault}{\mddefault}{\updefault}$N_{12}:$ $u^*$ }}}
\put(1276,-3136){\makebox(0,0)[lb]{\smash{\SetFigFont{9}{10.8}{\rmdefault}{\mddefault}{\updefault}$\Box_f$}}}
\put(2401,-3136){\makebox(0,0)[lb]{\smash{\SetFigFont{9}{10.8}{\rmdefault}{\mddefault}{\updefault}$\Box_{u^*}$ }}}
\put(2326,-3286){\makebox(0,0)[lb]{\smash{\SetFigFont{9}{10.8}{\rmdefault}{\mddefault}{\updefault}$N_{11}:$ $u^*$ }}}
\put(2851,-1561){\makebox(0,0)[lb]{\smash{\SetFigFont{9}{10.8}{\rmdefault}{\mddefault}{\updefault}$N_{13}:$ $u^*$ }}}
\put(3076,-1411){\makebox(0,0)[lb]{\smash{\SetFigFont{9}{10.8}{\rmdefault}{\mddefault}{\updefault}$\Box_{u^*}$ }}}
\put(3301,-511){\makebox(0,0)[lb]{\smash{\SetFigFont{9}{10.8}{\rmdefault}{\mddefault}{\updefault}$N_2:$  $q(X)$}}}
\put(2776,-961){\makebox(0,0)[lb]{\smash{\SetFigFont{9}{10.8}{\rmdefault}{\mddefault}{\updefault}$N_3:$ $\neg r$ }}}
\put(2551,-811){\makebox(0,0)[lb]{\smash{\SetFigFont{9}{10.8}{\rmdefault}{\mddefault}{\updefault}$\Box_t$}}}
\put(2326,-961){\makebox(0,0)[lb]{\smash{\SetFigFont{9}{10.8}{\rmdefault}{\mddefault}{\updefault}$N_{2'}:$}}}
\put(2701,-61){\makebox(0,0)[lb]{\smash{\SetFigFont{9}{10.8}{\rmdefault}{\mddefault}{\updefault}$N_0:$ $p(X)$ }}}
\put(3526,-211){\makebox(0,0)[lb]{\smash{\SetFigFont{8}{9.6}{\rmdefault}{\mddefault}{\updefault}$C_{p_1}$}}}
\put(2251,-436){\makebox(0,0)[lb]{\smash{\SetFigFont{9}{10.8}{\rmdefault}{\mddefault}{\updefault}$\Box_t$}}}
\put(2026,-586){\makebox(0,0)[lb]{\smash{\SetFigFont{9}{10.8}{\rmdefault}{\mddefault}{\updefault}$N_1:$}}}
\put(3451,-736){\makebox(0,0)[lb]{\smash{\SetFigFont{7}{8.4}{\rmdefault}{\mddefault}{\updefault}$C_{q_2}$}}}
\put(3001,-736){\makebox(0,0)[lb]{\smash{\SetFigFont{7}{8.4}{\rmdefault}{\mddefault}{\updefault}$C_{q_1}$}}}
\put(4201,-661){\makebox(0,0)[lb]{\smash{\SetFigFont{8}{9.6}{\rmdefault}{\mddefault}{\updefault}$C_{q_3}$}}}
\end{picture}

\caption{The generalized SLT-tree $GT_{\leftarrow p(X)}^2$ for 
$(P_1^2\cup \{\leftarrow p(X)\},TB_f^1)$. $\qquad\qquad$}\label{fig4}
\end{figure} 

\section{Soundness and Completeness of SLT-resolution}
In this section we establish the termination, 
soundness and completeness of SLT-resolution.

\begin{theorem}
\label{termin-normal}
For programs with the bounded-term-size property
SLT-resolution terminates in finite time.
\end{theorem}

\noindent {\bf Proof:}
Let $P$ be a program with the bounded-term-size property. Since $P$ 
has only a finite number of clauses, we have only a finite number,
say $N$, of ground subgoals in all generalized SLT-trees $GT_{G_0}^i$s. 
Before SLT-resolution stops, in each new recursion via $SLT()$
at least one new tabled negative answer will be derived.
Therefore, there are at most $N$ recursions in SLT-resolution.
By Theorem \ref{termin-positive}, each recursion (i.e. the execution
of $SLTP()$) will terminate in finite time, so we conclude the proof. $\Box$

\vspace{4mm}

By Theorem \ref{termin-normal},
for programs with the bounded-term-size property, by calling
$SLT(P,$ $G_0,R,\emptyset,\emptyset)$ SLT-resolution
generates a finite sequence of generalized SLT-trees:
\begin{eqnarray}
GT_{G_0}^1 & = & SLTP(P,G_0,R,\emptyset,\emptyset), \nonumber\\
GT_{G_0}^2 & = & SLTP(P^1,G_0,R,TB_t^1,TB_f^1), \qquad\qquad\qquad\nonumber \\
           & \vdots \label{seq} & \label{sequence}\\
GT_{G_0}^{k+1} & = & SLTP(P^k,G_0,R,TB_t^k,TB_f^k) \nonumber, 
\end{eqnarray}
where for each $1\leq i\leq k$, 
$P^i=P\cup TB_t^i$, and $TB_t^i$ and $TB_f^i$ respectively consist of 
all tabled positive and negative answers 
in all $GT_{G_0}^j$s $(j\leq i)$. $GT_{G_0}^{k+1}$
will be returned since it contains no new tabled answers
(see Definition \ref{slt}).

To simplify our presentation, in the following lemmas/corollaries/theorems, 
we assume that $P$ is a program with the bounded-term-size property, $G_0$
is a top goal, $GT_{G_0}=GT_{G_0}^{k+1}$ is as defined in (\ref{sequence}), 
and $T_{G_0}$ is the top SLT-tree in $GT_{G_0}$.

\begin{lemma}
\label{sc-lem}
Let $GT_{G_0}^i$ $(i\geq 1)$ be as defined in (\ref{sequence}).
For any selected ground subgoal $A$ in $GT_{G_0}^{i+1}$,
if $A$ is in $TB_f^i$ then all sub-derivations for $A$ in 
$GT_{G_0}^{i+1}$ will end with a failure leaf.
\end{lemma}

\noindent {\bf Proof:} $A\in TB_f^i$ indicates that if $A$ is 
a selected ground subgoal in $GT_{G_0}^i$, all its sub-derivations
end with a failure leaf. This implies that the truth value of $A$ 
does not depend on any selected negative subgoals whose truth values are
temporarily undefined in $GT_{G_0}^i$. Since $GT_{G_0}^{i+1}$ is derived
from $GT_{G_0}^i$ simply by treating some selected negative subgoals $\neg B$
whose truth values are temporarily undefined in $GT_{G_0}^i$ as true by
assuming $B$ is false, such process obviously will not affect the truth value
of $A$. Therefore, all sub-derivations for $A$ in $GT_{G_0}^{i+1}$ will
end with a failure leaf. $\Box$

\begin{lemma}
\label{sc-lem1}
$\qquad$
\begin{enumerate}
\item
For any selected positive literal $A$ in $GT_{G_0}$,
there is a correct answer substitution $\gamma$ for $A$ in $GT_{G_0}$
if and only if $A\gamma\in TB_t^k$ (up to variable renaming).
\item
For any selected positive literal $A$ at any node $N_i$ in $T_{G_0}$,
there is a correct answer substitution $\gamma$ for $A$ in $GT_{G_0}$
if and only if there is a correct answer substitution $\gamma$ for $A$ 
at node $N_i$ in $T_{G_0}$ (up to variable renaming).
\end{enumerate}
\end{lemma}

\noindent {\bf Proof:}
Point 1 is straightforward by the fact that
$GT_{G_0}$ contains no new tabled positive answers. By point 1,
all correct answer substitutions for $A$ in $GT_{G_0}$ are in $TB_t^k$. 
Hence point 2 follows immediately from the fact that the selected literal 
$A$ at node $N_i$ in $T_{G_0}$ will
use all tabled answers in $TB_t^k$ that unify with $A$. $\Box$

\begin{lemma}
\label{sc-lem2}
For any selected positive literal $A$ in $GT_{G_0}$, 
$A\theta\in M_{P^k}(\neg .TB_f^k)$ if and only if there 
is a correct answer substitution for
$A$ in $GT_{G_0}$ that is more general than $\theta$, and 
for any selected ground positive 
literal $A$ in $GT_{G_0}$, $A\in N_{P^k}(\neg .TB_f^k)$
if and only if all sub-derivations for $A$ 
and $S$ (defined in Theorem \ref{main-th1}) 
end with a failure leaf.
\end{lemma}

\noindent {\bf Proof:}
Let $GT_{G_0}^1=SLTP(P,G_0,R,\emptyset,\emptyset)$
and $TB_t^1$ and $TB_f^1$ consist of all tabled positive and
negative answers in $GT_{G_0}^1$, respectively. By Theorem \ref{main-th1},
for any selected positive literal $A$ in $GT_{G_0}^1$, 
$A\theta\in M_P(\emptyset)$ if and only if there is 
a correct answer substitution for
$A$ in $GT_{G_0}^1$ that is more general than $\theta$,
 and that for any selected ground negative literal 
$\neg A$ in $GT_{G_0}^1$, $A\in N_P(\emptyset)$
if and only if all sub-derivations for $A$ in 
$GT_{G_0}^1$ end with a failure leaf.
Let $P^1=P\cup TB_t^1$. Then $P^1$ is equivalent to $P$ under the
well-founded semantics.   

Let $GT_{G_0}^2=SLTP(P^1,G_0,R,TB_t^1,TB_f^1)$. Observe that 
$SLTP(P^1,G_0,R,TB_t^1,$ $TB_f^1)$ works in the same way as 
$SLTP(P^1,G_0,R,\emptyset,\emptyset)$ except whenever a negative
subgoal $\neg A$ with $A\in TB_f^1$ is selected, it will directly be 
treated as true instead of trying to prove $A$ by building a
child SLT-tree $T_{\leftarrow A}$ for $\leftarrow A$. 
When a positive subgoal $A\in TB_f^1$ is selected, 
all sub-derivations for $A$ will still be generated.
However, By Lemma \ref{sc-lem} all these sub-derivations
will end with a failure leaf, which implies that $A$ is
false. Therefore $SLTP(P^1,G_0,R,TB_t^1,TB_f^1)$
can be viewed as $SLTP(P^1,G_0,R,\emptyset,\emptyset)$ with the 
exception that all selected ground subgoals in $TB_f^1$ are 
treated as false instead of being temporarily undefined. This means that
$SLTP(P^1,G_0,R,TB_t^1,TB_f^1)$
has the same relationship to $M_{P^1}(\neg .TB_f^1)$ and
$N_{P^1}(\neg .TB_f^1)$ as $SLTP(P^1,G_0,R,\emptyset,\emptyset)$
to $M_{P^1}(\emptyset)$ and $N_{P^1}(\emptyset)$. That is, by 
Theorem \ref{main-th1} for any selected positive literal $A$ in $GT_{G_0}^2$, 
$A\theta\in M_{P^1}(\neg .TB_f^1)$ if and only 
if there is a correct answer substitution for
$A$ in $GT_{G_0}^2$ that is more general than $\theta$, and 
for any selected ground literal $A$ in $GT_{G_0}^2$, $A\in N_{P^1}(\neg .TB_f^1)$
if and only if all sub-derivations for $A$ and
$S$ in $GT_{G_0}^2$ end with a failure leaf. 

Continuing the above arguments, we will reach the same conclusion
for any $GT_{G_0}^{i+1}=SLTP(P^i,G_0,R,TB_t^i,TB_f^i)$ 
with $i\geq 1$. $\Box$

\vspace{4mm}

In the above proof we have $TB_f^1\subseteq N_P(\emptyset)$, so that
$\neg .TB_f^1\subseteq WF(P)$. Meanwhile, for each $A\in TB_t^1$ we have
$M_P(\emptyset)\models \forall (A)$, so that $WF(P) \models \forall (A)$.
Therefore, $P^1=P\cup TB_t^1$ is equivalent to $P$ under the
well-founded semantics, and by Lemma \ref{lem2-2} $M_{P^1}(\neg .TB_f^1)
\subseteq WF(P)$ and $\neg .N_{P^1}(\neg .TB_f^1)\subseteq WF(P)$.
For the same reason we have $TB_f^2\subseteq N_{P^1}(\neg .TB_f^1)$, so that
$\neg .TB_f^2\subseteq WF(P)$; and for each $A\in TB_t^2$ we have
$M_{P^1}(\neg .TB_f^1)\models \forall (A)$, so that $WF(P) \models \forall (A)$.
This leads to $P^2=P\cup TB_t^2$ being equivalent to $P$ under the
well-founded semantics, $M_{P^2}(\neg .TB_f^2)
\subseteq WF(P)$ and $\neg .N_{P^2}(\neg .TB_f^2)\subseteq WF(P)$.
Repeating this process leads to the following result.

\begin{corollary}
\label{sc-cor2}
For any $i\geq 1$, if $A\in TB_f^i$ then $WF(P)\models \neg A$,
and if $A\in TB_t^i$ then $WF(P)\models \forall (A)$.
\end{corollary}

\begin{lemma}
\label{sc-lem3}
$\qquad$
\begin{enumerate}
\item
Let $A$ be a selected positive literal in $GT_{G_0}$.  
For any (Herbrand) ground instance $A\theta$ of $A$, 
$WF(P)\models A\theta$ if and only if $A\theta\in M_{P^k}(\neg .TB_f^k)$.
\item
For any selected ground negative literal $\neg A$ in $GT_{G_0}$,
$WF(P)\models \neg A$ if and only if $A\in TB_f^k$.
\end{enumerate}
\end{lemma}

\noindent {\bf Proof:} 1. $(\Longleftarrow)$ Assume
$A\theta\in M_{P^k}(\neg .TB_f^k)$. By Lemma \ref{sc-lem2} there 
is a correct answer substitution for $A$ in $GT_{G_0}$ that is 
more general than $\theta$. Since $TB_t^k$ consists of all tabled
positive answers in all $GT_{G_0}^i$s $(i\geq 1)$, there is an
$A\gamma\in TB_t^k$ with $\gamma$ more general than $\theta$.
By Corollary \ref{sc-cor2} $WF(P)\models \forall (A\gamma)$, so that
$WF(P)\models A\theta$.

$(\Longrightarrow)$ Assume $WF(P)\models A\theta$. Since $P^k$
is equivalent to $P$ under the well-founded semantics, 
$WF(P^k)\models A\theta$. Assume, on the contrary,
$A\theta\not\in M_{P^k}(\neg .TB_f^k)$. Since $\neg .N_{P^k}(\neg 
.TB_f^k)$ $\subseteq WF(P^k)$, $A\theta$ is in $O_{P^k}(\neg .TB_f^k)$.
So there exists a ground backward chain of the form
\begin{equation}
\qquad A\theta\Rightarrow_{S_1}... \Rightarrow_{S_i}
B_1,...,B_m,\neg D_1,...,\neg D_n\Rightarrow_{S_{i+1}}
...\Rightarrow_{S_t}\Box
\qquad\qquad\qquad\qquad \label{chain2}
\end{equation}
where each step is performed by either resolving a positive literal
like $B_j$ with an answer in $M_{P^k}(\neg .TB_f^k)$
(when $B_j\in M_{P^k}(\neg .TB_f^k)$) or with 
a Herbrand instantiated clause of $P$ (otherwise), 
or removing a negative literal like
$\neg D_j$ where $D_j\not\in M_{P^k}(\neg .TB_f^k)$. 
Observe that for each negative literal
$\neg D$ occurring in the chain, either $D\in TB_f^k$
or $D\in O_{P^k}(\neg .TB_f^k)$ or $D\in N_{P^k}(\neg .TB_f^k)$.
However, since $A\theta$ is true in $WF(P^k)$, $D$ must be false in $WF(P^k)$. 
If $D$ is in $TB_f^k$, it has already been treated to be false; otherwise,  
by Definition \ref{trans-ops2} $\neg D$ cannot be derived unless we 
assume some atoms in $N_{P^k}(\neg .TB_f^k)-TB_f^k$ to be false. 
This implies that for each negative literal $\neg D$ occurring in the 
above chain with $D\not\in TB_f^k$, the proof of $D$ will be recursively
reduced to the proof of some literals in $N_{P^k}(\neg .TB_f^k)-TB_f^k$. 

By using similar arguments of Theorem \ref{main-th1}, we can have
a sub-derivation $SD_A$ for $A$ in $GT_{G_0}$, which corresponds to 
the backward chain (\ref{chain2}), that ends with a temporarily undefined
leaf. In $SD_A$, each selected ground negative 
literal $\neg D$ is true if $D\in TB_f^k$; temporarily undefined, 
otherwise (note $D\not\in M_{P^k}(\neg .TB_f^k)$). Since the sub-derivation
ends with a temporarily undefined leaf, it has at least one 
selected ground negative literal $\neg D$ with $D\not\in TB_f^k$.
Let 
\begin{center}
$S=\{D|D\not\in TB_f^k$ and $\neg D$ is a selected ground negative
literal in $SD_A$$\}$. 
\end{center}
Then by Lemma \ref{lem2-3} each $D\in S$ is either in 
$N_{P^k}(\neg .TB_f^k)-TB_f^k$ 
or in $O_{P^k}(\neg .TB_f^k)$. We consider two cases. 

Case 1. There exists a $D\in S$ with 
$D\in N_{P^k}(\neg .TB_f^k)-TB_f^k$. 
By Lemma \ref{sc-lem2} all sub-derivations for $D$ in $GT_{G_0}$ 
end with a failure leaf. Since $D$ is not in $TB_f^k$, it is a 
new tabled negative answer in $GT_{G_0}$, which contradicts that
$GT_{G_0}$ has no new tabled negative answers.

Case 2. Every $D\in S$ is in $O_{P^k}(\neg .TB_f^k)$. 
Since the backward chain (\ref{chain2}) is an instance of the
sub-derivation $SD_A$, all $D$s in $S$ must be false in $WF(P^k)$.
However, as discussed above no $\neg D$ can be derived unless we 
assume some atoms in $N_{P^k}(\neg .TB_f^k)-TB_f^k$ to be false. That is,
the proof of each $D\in S$ can be recursively
reduced to the proof of some literals in $N_{P^k}(\neg .TB_f^k)-TB_f^k$.
So $GT_{G_0}$ must have a subpath of the form 
\begin{tabbing}
$\qquad$ \= $...\neg D$\= $\ \cdots\triangleright$ $D\Rightarrow ...\Rightarrow$\\
\>         $...\neg E_1$\> $\ \cdots\triangleright$ $E_1\Rightarrow ...\Rightarrow$\\
\>\> $\quad \vdots$\\
\>         $...\neg E_t$\> $\ \cdots\triangleright$ $E_t$
\end{tabbing}
\noindent where $E_t\in N_{P^k}(\neg .TB_f^k)-
TB_f^k$. For the same reason as in the first case, $E_t$ should be
a new tabled negative answer in $GT_{G_0}$, which leads to a contradiction. 

2. $(\Longleftarrow)$ Immediate from Corollary \ref{sc-cor2}.

$(\Longrightarrow)$ Assume $WF(P)\models \neg A$ but on the
contrary $A\not\in TB_f^k$. By point 1 of this lemma,
$A\not\in M_{P^k}(\neg .TB_f^k)$. If $A\in N_{P^k}(\neg .TB_f^k)$ then
by Lemma \ref{sc-lem2} all sub-derivations for $A$ in $GT_{G_0}$ will
end with a failure leaf. Since $A$ is not in $TB_f^k$, it is a 
new tabled negative answer, contradicting that
$GT_{G_0}$ has no new tabled negative answers. 
So $A\in O_{P^k}(\neg .TB_f^k)$.

Similar to the arguments for point 1 of this lemma, 
the proof of $A$ can be recursively
reduced to the proof of some literals in $N_{P^k}(\neg .TB_f^k)-
TB_f^k$, which will lead to
new tabled negative answers in $GT_{G_0}$, a contradiction. $\Box$

\begin{lemma}
\label{sc-lem4}
Let $G_0\leftarrow A$ be a top goal (with $A$ an atom). 
$WF(P)\models \neg\exists (A)$ if and only if all branches of $T_{G_0}$
end with a failure leaf.
\end{lemma}

\noindent {\bf Proof:} 
$(\Longleftarrow)$ Assume all branches of $T_{G_0}$ end with a failure
leaf. Let $A\theta$ be a ground instance of $A$. By Lemmas \ref{sc-lem1}
(point 2) and \ref{sc-lem2}, $A\theta\not\in M_{P^k}(\neg .TB_f^k)$,
so by Lemma \ref{sc-lem3} $WF(P)\not\models A\theta$. Assume, on the 
contrary, $WF(P)\not\models\neg A\theta$. By Corollary \ref{sc-cor2},
$A\theta\not\in N_{P^k}(\neg .TB_f^k)$ and thus 
$A\theta\in O_{P^k}(\neg .TB_f^k)$. Then there exists a ground 
backward chain of the form
\begin{equation}
\qquad A\theta\Rightarrow_{S_1}... \Rightarrow_{S_i}
B_1,...,B_m,\neg D_1,...,\neg D_n\Rightarrow_{S_{i+1}}
...\Rightarrow_{S_t}\Box
\qquad\qquad\qquad\qquad \label{chain3}
\end{equation}
where each step is performed by either resolving a positive literal
like $B_j$ with an answer in $M_{P^k}(\neg .TB_f^k)$
(when $B_j\in M_{P^k}(\neg .TB_f^k)$) or with 
a Herbrand instantiated clause of $P$ (otherwise), 
or removing a negative literal like
$\neg D_j$ where $D_j\not\in M_{P^k}(\neg .TB_f^k)$. 
Observe that for each negative literal
$\neg D$ occurring in the chain, either $D\in TB_f^k$
or $D\in O_{P^k}(\neg .TB_f^k)$ or $D\in N_{P^k}(\neg .TB_f^k)$.
However, since $A\theta$ is neither true nor false in $WF(P)$, 
there exists at least one $D\in O_{P^k}(\neg .TB_f^k)$.

By using similar arguments of Theorem \ref{main-th1}, $T_{G_0}$ must have
a branch, which corresponds to the backward chain (\ref{chain3}), 
that ends with a temporarily undefined leaf. This contradicts the 
assumption that all branches of $T_{G_0}$ end with a failure leaf.
Therefore, for any ground instance $A\theta$ of $A$
$WF(P)\models\neg A\theta$. That is, $WF(P)\models \neg\exists (A)$. 

$(\Longrightarrow)$ Assume $WF(P)\models \neg\exists (A)$. 
By Lemmas \ref{sc-lem3} and \ref{sc-lem2}, there is no sub-derivation
for $A$ that ends with a success leaf in $GT_{G_0}$.

Now assume, on the contrary, that $T_{G_0}$ has a branch $BR$ that
ends with a temporarily undefined leaf. Then $BR$ has at least one
ground instance corresponding to the ground backward chain like 
(\ref{chain3}). Since $A\theta$ is false in $WF(P)$, there
exists at least one ground negative literal $\neg D$ in the chain 
such that $D$ is true in $WF(P)$. 
This means that there is a selected ground negative literal $\neg D$
in $BR$ such that $D$ is true in $WF(P)$. By Corollary \ref{sc-cor2}
$D\not\in TB_f^k$, so by Definition \ref{slt-tree}
a child SLT-tree $T_{\leftarrow D}$ must be
built where $D$ is a selected positive literal. Since $BR$ is a
temporarily undefined branch, $\neg D$ cannot fail, so
$T_{\leftarrow D}$ has no successful branch (i.e. $\neg D$ is
treated as $u^*$; see point 4 of Definition \ref{slt-tree}). 
By Lemma \ref{sc-lem2} $D\not\in M_{P^k}(\neg .TB_f^k)$
and by Lemma \ref{sc-lem3} $WF(P)\not\models D$, which contradicts 
that $D$ is true in $WF(P)$. Therefore, all branches of $T_{G_0}$
must end with a failure leaf. $\Box$

\vspace{4mm}

Now we are ready to show the
soundness and completeness of SLT-resolution.

\begin{theorem}
\label{sound-comp-normal}
Let $\bar{P}$ be the augmented version of $P$.
Let $G_0\leftarrow A$ be a top goal (with $A$ an atom) and
$\theta$ a substitution for the variables of $A$. Assume 
neither $A$ nor $\theta$ contains the 
symbols $\bar{p}$ or $\bar{f}$ or $\bar{c}$.
\begin{enumerate}
\item
$WF(P) \models \exists (A)$ if and only if $G_0$ is true 
in $P$ with an instance of $A$;

\item
$WF(P) \models \neg \exists (A)$ if and only if $G_0$ is false in $P$;

\item
$WF(P) \not\models \exists (A)$ and  $WF(P) \not\models \neg\exists (A)$
if and only if $G_0$ is undefined in $P$;

\item
If $G_0$ is true in $P$ with an answer $A\theta$ then
$WF(P) \models \forall (A\theta)$;

\item
If $WF(\bar{P}) \models \forall (A\theta)$ then $G_0$ is true in $P$ with 
an answer $A\theta$.
\end{enumerate}
\end{theorem}

\noindent {\bf Proof:} 
\begin{enumerate}
\item
Immediate from Lemmas \ref{sc-lem2} and 
\ref{sc-lem3}.

\item
Immediate from Lemma \ref{sc-lem4}.

\item
Immediate from points 1 and 2 of this theorem.

\item
Assume $G_0$ is true in $P$ with an answer $A\theta$.
Then there is a correct answer substitution $\gamma$ in $T_{G_0}$
that is more general than $\theta$. By Theorem \ref{ans-subs}
$WF(P^k) \models \forall (A\gamma)$ and thus
$WF(P) \models \forall (A\gamma)$ since $P^k$ is equivalent to
$P$ w.r.t. the well-founded semantics. Therefore
$WF(P) \models \forall (A\theta)$.

\item
Note that $\bar{P}=P\cup \{\bar{p}(\bar{f}(\bar{c}))\}$.
Let $T_{G_0}'$ be the top SLT-tree in $GT_{G_0}'$ that 
is returned by $SLT(\bar{P},G_0,R,\emptyset,\emptyset)$.
Since none of the symbols $\bar{p}$ or $\bar{f}$ or $\bar{c}$
appears in $P\cup \{G_0\}$, $T_{G_0}'=T_{G_0}$ and
$GT_{G_0}'=GT_{G_0}$.

Let $\{X_0,...,X_n\}$ be the set of variables appearing
in $A\theta$ and $\alpha$ be the ground substitution
$\{X_0/\bar{c}, X_1/\bar{f}(\bar{c}),...,X_n/\bar{f}^n(\bar{c})\}$.
Then $WF(\bar{P})\models A\theta\alpha$ and by Lemmas
\ref{sc-lem3}, \ref{sc-lem2} and \ref{sc-lem1} 
there is a correct answer substitution $\gamma$ for $G_0$ in 
$T_{G_0}'$ that is more general than $\theta\alpha$. That is, there exists
a substitution $\beta$ such that 
$\gamma\beta=\theta\alpha$. Since $T_{G_0}'=T_{G_0}$,
$\gamma$ contains neither $\bar{f}$ nor $\bar{c}$. So the only 
occurrences of $\bar{f}$ and $\bar{c}$ in $\gamma\beta$
are in $\beta$. Let $\beta'$ be obtained from $\beta$ by
replacing every occurrence of $\bar{f}^i(\bar{c})$ by
the variable $X_i$. Then $\gamma\beta'=\theta$ and thus
$\gamma$ is more general than $\theta$. 

Since $T_{G_0}'=T_{G_0}$, there is a correct answer 
substitution $\gamma$ for $G_0$ in 
$T_{G_0}$ that is more general than $\theta$. Therefore,
by Definition \ref{true-false} $G_0$ is true in $P$ with 
an answer $A\theta$. $\Box$ 
\end{enumerate}

Observe that in point 5 of Theorem \ref{sound-comp-normal}
we used the augmented program $\bar{P}$ to characterize
part of the completeness of SLT-resolution. The concept of
augmented programs was introduced by Van Gelder, Ross
and Schlipf \cite{VRS91}, which is used to deal with the so
called {\em universal query problem} \cite{Prz89-2}.
As indicated by Ross \cite{Ross92}, we cannot substitute
$P$ for $\bar{P}$ in point 5 of Theorem \ref{sound-comp-normal}.
A very simple illustrating example is that let $P=\{p(a)\}$ and
$G_0=\leftarrow p(X)$, we have $WF(P)\models \forall (p(X)\{X/X\})$
under Herbrand interpretations, but we have no correct answer
substitution for $G_0$ in $T_{G_0}$ that is more general than
$\{X/X\}$.

\section{Optimizations of SLT-resolution}
\label{sec5}
The objective of this paper
is to develop an evaluation procedure for the well-founded
semantics that is linear, free of infinite loops and with less redundant
computations. Clearly, SLT-resolution is linear and with no infinite loops.
However, like SLDNF-trees, SLT-trees defined in Definition 
\ref{slt-tree} may contain a lot of duplicated
sub-branches. SLT-resolution can be considerably optimized by eliminating
those redundant computations. In this section we present three 
effective methods for the optimization of SLT-resolution.
 
\subsection{Negation as the Finite Failure of Loop-Independent Nodes}
From Definition \ref{slt} we see that SLT-resolution exhausts the
answers of the top goal $G_0$ by recursively calling the function $SLTP()$.
Obviously, the less the number of recursions is, the more efficient
SLT-resolution would be. In this subsection we identify a large
class of recursions that can easily be avoided. We start with an
example.

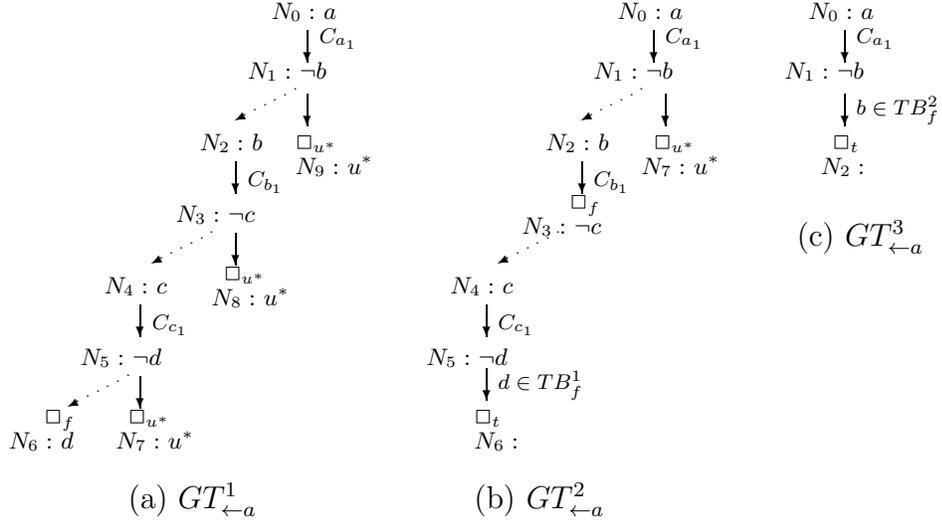
\begin{figure}[htb]
\centering

\setlength{\unitlength}{3947sp}%
\begingroup\makeatletter\ifx\SetFigFont\undefined%
\gdef\SetFigFont#1#2#3#4#5{%
  \reset@font\fontsize{#1}{#2pt}%
  \fontfamily{#3}\fontseries{#4}\fontshape{#5}%
  \selectfont}%
\fi\endgroup%
\begin{picture}(5325,3180)(226,-2611)
\put(3151,-2611){\makebox(0,0)[lb]{\smash{\SetFigFont{12}{14.4}{\rmdefault}{\mddefault}{\updefault}(b) $GT_{\leftarrow a}^2$}}}
\put(976,-2611){\makebox(0,0)[lb]{\smash{\SetFigFont{12}{14.4}{\rmdefault}{\mddefault}{\updefault}(a) $GT_{\leftarrow a}^1$}}}
\put(5176,-961){\makebox(0,0)[lb]{\smash{\SetFigFont{12}{14.4}{\rmdefault}{\mddefault}{\updefault}(c) $GT_{\leftarrow a}^3$}}}
\thinlines
\put(1651,-436){\vector( 0,-1){200}}
\put(1121,-1076){\vector(-2,-1){0}}
\multiput(1501,-886)(-63.33333,-31.66667){6}{\makebox(1.6667,11.6667){\SetFigFont{5}{6}{\rmdefault}{\mddefault}{\updefault}.}}
\put(1051,-1336){\vector( 0,-1){200}}
\put(596,-1976){\vector(-2,-1){0}}
\multiput(976,-1786)(-63.33333,-31.66667){6}{\makebox(1.6667,11.6667){\SetFigFont{5}{6}{\rmdefault}{\mddefault}{\updefault}.}}
\put(2101,389){\vector( 0,-1){200}}
\put(1646,-176){\vector(-2,-1){0}}
\multiput(2026, 14)(-63.33333,-31.66667){6}{\makebox(1.6667,11.6667){\SetFigFont{5}{6}{\rmdefault}{\mddefault}{\updefault}.}}
\put(1051,-1786){\vector( 0,-1){200}}
\put(1651,-886){\vector( 0,-1){200}}
\put(2101,-11){\vector( 0,-1){200}}
\put(3826,-436){\vector( 0,-1){200}}
\put(3296,-1076){\vector(-2,-1){0}}
\multiput(3676,-886)(-63.33333,-31.66667){6}{\makebox(1.6667,11.6667){\SetFigFont{5}{6}{\rmdefault}{\mddefault}{\updefault}.}}
\put(3226,-1336){\vector( 0,-1){200}}
\put(4276,389){\vector( 0,-1){200}}
\put(3821,-176){\vector(-2,-1){0}}
\multiput(4201, 14)(-63.33333,-31.66667){6}{\makebox(1.6667,11.6667){\SetFigFont{5}{6}{\rmdefault}{\mddefault}{\updefault}.}}
\put(3226,-1736){\vector( 0,-1){200}}
\put(4351,-11){\vector( 0,-1){200}}
\put(5476,389){\vector( 0,-1){200}}
\put(5476,-11){\vector( 0,-1){200}}
\put(1426,-361){\makebox(0,0)[lb]{\smash{\SetFigFont{9}{10.8}{\rmdefault}{\mddefault}{\updefault}$N_2:$ $ b$ }}}
\put(1726,-586){\makebox(0,0)[lb]{\smash{\SetFigFont{8}{9.6}{\rmdefault}{\mddefault}{\updefault}$C_{b_1}$}}}
\put(1276,-811){\makebox(0,0)[lb]{\smash{\SetFigFont{9}{10.8}{\rmdefault}{\mddefault}{\updefault}$N_3:$ $\neg c$ }}}
\put(826,-1261){\makebox(0,0)[lb]{\smash{\SetFigFont{9}{10.8}{\rmdefault}{\mddefault}{\updefault}$N_4:$ $c$ }}}
\put(1126,-1486){\makebox(0,0)[lb]{\smash{\SetFigFont{8}{9.6}{\rmdefault}{\mddefault}{\updefault}$C_{c_1}$}}}
\put(676,-1711){\makebox(0,0)[lb]{\smash{\SetFigFont{9}{10.8}{\rmdefault}{\mddefault}{\updefault}$N_5:$ $\neg d$ }}}
\put(226,-2236){\makebox(0,0)[lb]{\smash{\SetFigFont{9}{10.8}{\rmdefault}{\mddefault}{\updefault}$N_6:$ $d$ }}}
\put(451,-2086){\makebox(0,0)[lb]{\smash{\SetFigFont{9}{10.8}{\rmdefault}{\mddefault}{\updefault}$\Box_f$}}}
\put(1876,464){\makebox(0,0)[lb]{\smash{\SetFigFont{9}{10.8}{\rmdefault}{\mddefault}{\updefault}$N_0:$ $a$ }}}
\put(2176,314){\makebox(0,0)[lb]{\smash{\SetFigFont{8}{9.6}{\rmdefault}{\mddefault}{\updefault}$C_{a_1}$}}}
\put(1726, 89){\makebox(0,0)[lb]{\smash{\SetFigFont{9}{10.8}{\rmdefault}{\mddefault}{\updefault}$N_1:$ $\neg b$ }}}
\put(976,-2086){\makebox(0,0)[lb]{\smash{\SetFigFont{9}{10.8}{\rmdefault}{\mddefault}{\updefault}$\Box_{u^*}$ }}}
\put(901,-2236){\makebox(0,0)[lb]{\smash{\SetFigFont{9}{10.8}{\rmdefault}{\mddefault}{\updefault}$N_7:$ $u^*$ }}}
\put(2026,-361){\makebox(0,0)[lb]{\smash{\SetFigFont{9}{10.8}{\rmdefault}{\mddefault}{\updefault}$\Box_{u^*}$ }}}
\put(2026,-511){\makebox(0,0)[lb]{\smash{\SetFigFont{9}{10.8}{\rmdefault}{\mddefault}{\updefault}$N_9:$ $u^*$ }}}
\put(1576,-1186){\makebox(0,0)[lb]{\smash{\SetFigFont{9}{10.8}{\rmdefault}{\mddefault}{\updefault}$\Box_{u^*}$ }}}
\put(1501,-1336){\makebox(0,0)[lb]{\smash{\SetFigFont{9}{10.8}{\rmdefault}{\mddefault}{\updefault}$N_8:$ $u^*$ }}}
\put(3601,-361){\makebox(0,0)[lb]{\smash{\SetFigFont{9}{10.8}{\rmdefault}{\mddefault}{\updefault}$N_2:$ $ b$ }}}
\put(3901,-586){\makebox(0,0)[lb]{\smash{\SetFigFont{8}{9.6}{\rmdefault}{\mddefault}{\updefault}$C_{b_1}$}}}
\put(3001,-1261){\makebox(0,0)[lb]{\smash{\SetFigFont{9}{10.8}{\rmdefault}{\mddefault}{\updefault}$N_4:$ $c$ }}}
\put(3301,-1486){\makebox(0,0)[lb]{\smash{\SetFigFont{8}{9.6}{\rmdefault}{\mddefault}{\updefault}$C_{c_1}$}}}
\put(2851,-1711){\makebox(0,0)[lb]{\smash{\SetFigFont{9}{10.8}{\rmdefault}{\mddefault}{\updefault}$N_5:$ $\neg d$ }}}
\put(4051,464){\makebox(0,0)[lb]{\smash{\SetFigFont{9}{10.8}{\rmdefault}{\mddefault}{\updefault}$N_0:$ $a$ }}}
\put(4351,314){\makebox(0,0)[lb]{\smash{\SetFigFont{8}{9.6}{\rmdefault}{\mddefault}{\updefault}$C_{a_1}$}}}
\put(3901, 89){\makebox(0,0)[lb]{\smash{\SetFigFont{9}{10.8}{\rmdefault}{\mddefault}{\updefault}$N_1:$ $\neg b$ }}}
\put(3151,-2236){\makebox(0,0)[lb]{\smash{\SetFigFont{9}{10.8}{\rmdefault}{\mddefault}{\updefault}$N_6:$  }}}
\put(3151,-2086){\makebox(0,0)[lb]{\smash{\SetFigFont{9}{10.8}{\rmdefault}{\mddefault}{\updefault}$\Box_t$}}}
\put(3451,-886){\makebox(0,0)[lb]{\smash{\SetFigFont{9}{10.8}{\rmdefault}{\mddefault}{\updefault}$N_3:$ $\neg c$ }}}
\put(3751,-736){\makebox(0,0)[lb]{\smash{\SetFigFont{9}{10.8}{\rmdefault}{\mddefault}{\updefault}$\Box_f$}}}
\put(4201,-511){\makebox(0,0)[lb]{\smash{\SetFigFont{9}{10.8}{\rmdefault}{\mddefault}{\updefault}$N_7:$ $u^*$ }}}
\put(4276,-361){\makebox(0,0)[lb]{\smash{\SetFigFont{9}{10.8}{\rmdefault}{\mddefault}{\updefault}$\Box_{u^*}$ }}}
\put(3301,-1861){\makebox(0,0)[lb]{\smash{\SetFigFont{8}{9.6}{\rmdefault}{\mddefault}{\updefault}$d\in TB_f^1$}}}
\put(5251,464){\makebox(0,0)[lb]{\smash{\SetFigFont{9}{10.8}{\rmdefault}{\mddefault}{\updefault}$N_0:$ $a$ }}}
\put(5551,314){\makebox(0,0)[lb]{\smash{\SetFigFont{8}{9.6}{\rmdefault}{\mddefault}{\updefault}$C_{a_1}$}}}
\put(5101, 89){\makebox(0,0)[lb]{\smash{\SetFigFont{9}{10.8}{\rmdefault}{\mddefault}{\updefault}$N_1:$ $\neg b$ }}}
\put(5401,-361){\makebox(0,0)[lb]{\smash{\SetFigFont{9}{10.8}{\rmdefault}{\mddefault}{\updefault}$\Box_t$ }}}
\put(5326,-511){\makebox(0,0)[lb]{\smash{\SetFigFont{9}{10.8}{\rmdefault}{\mddefault}{\updefault}$N_2:$  }}}
\put(5551,-136){\makebox(0,0)[lb]{\smash{\SetFigFont{8}{9.6}{\rmdefault}{\mddefault}{\updefault}$b\in TB_f^2$}}}
\end{picture}

\caption{The generalized SLT-trees $GT_{\leftarrow a}^1$, 
$GT_{\leftarrow a}^2$ and $GT_{\leftarrow a}^3$.}\label{fig5-1}
\end{figure}

\begin{example}
\label{eg5-1}
{\em
Consider the following program:
\begin{tabbing} 
\hspace{.2in} $P_2$: \= $a \leftarrow \neg b.$ \`$C_{a_1}$ \\ 
\> $b \leftarrow \neg c.$ \`$C_{b_1}$\\
\> $c \leftarrow \neg d.$ \`$C_{c_1}$  
\end{tabbing} 
Let $G_0=\leftarrow a$ be the top goal. 
Calling $SLT(P_2,G_0,R,\emptyset,\emptyset)$
immediately invokes $SLTP(P_2,G_0,$ $R,\emptyset,\emptyset)$, which builds
the first generalized SLT-tree $GT_{\leftarrow a}^1$
as shown in Figure \ref{fig5-1} (a). Since there is no tabled positive 
answer in $GT_{\leftarrow a}^1$ ($TB_t^1=\emptyset$), 
the first tabled negative answer $d$ is 
derived, which yields $TB_f^1=\{d\}$. Then 
$SLT(P_2,G_0,R,\emptyset,TB_f^1)$ is called, which
invokes $SLTP(P_2,G_0,R,\emptyset,TB_f^1)$ that builds the
second generalized SLT-tree $GT_{\leftarrow a}^2$
as shown in Figure \ref{fig5-1} (b). $GT_{\leftarrow a}^2$ has
a new tabled positive answer $c$, so
$SLTP(P_2\cup \{c\},G_0,R,\{c\},TB_f^1)$ is executed, which produces no
new tabled positive answers. The second tabled negative
answer $b$ is then obtained from $GT_{\leftarrow a}^2$. So far,
$TB_t^2=\{c\}$ and $TB_f^2=\{b,d\}$. Next,
$SLT(P_2\cup TB_t^2,G_0,R,TB_t^2,TB_f^2)$ is called, which
invokes $SLTP(P_2\cup TB_t^2,G_0,R,TB_t^2,TB_f^2)$ that builds the
third generalized SLT-tree $GT_{\leftarrow a}^3$
as shown in Figure \ref{fig5-1} (c). We see $a$ is true in
$GT_{\leftarrow a}^3$. As a result, to derive the first answer of $a$ 
$SLT()$ is called three times and $SLTP()$ four times.

}
\end{example}

Carefully examining the generalized SLT-tree 
$GT_{\leftarrow a}^1$ in Figure \ref{fig5-1},
we notice that it contains no loops. That is, all nodes in it 
are loop-independent. Consider the selected positive literal $d$
at $N_6$. Since there is no sub-derivation for $d$ starting at $N_6$
that ends with a temporarily undefined leaf and the proof of $d$
is independent of all its ancestor subgoals, the set of sub-derivations
for $d$ will remain unchanged throughout the recursions of $SLT()$;
i.e. it will not change in all $GT_{\leftarrow a}^i$s $(i>1)$
in which $d$ is a selected positive literal. This means that all answers
of $d$ can be determined only based on its sub-derivations starting at 
$N_6$ in $GT_{\leftarrow a}^1$, which leads to the following result. 

\begin{theorem}
\label{th5-1}
Let $GT_{G_0}=GT_{G_0}^{k+1}=SLTP(P^k,G_0,R,TB_t^k,TB_f^k)$, which
is returned by $SLT(P,G_0,R,\emptyset,\emptyset)$.
Let $A$ be a selected positive literal at a loop-independent node 
$N_i$ in $GT_{G_0}^{j+1}=SLTP(P^j,G_0,R,TB_t^j,TB_f^j)$ $(j\leq k)$ 
in which all sub-derivations $SD_A$
for $A$ starting at $N_i$ end with a non-temporarily undefined leaf.
Then $\theta$ is a correct answer substitution for $A$ in $SD_A$
if and only if $A\theta$ is a tabled positive answer for $A$ in $TB_t^k$;
and $A$ is false in $P$ if and only if all branches of $SD_A$ end with a failure
leaf.
\end{theorem}

\noindent {\bf Proof:}
Let $T_{\leftarrow A}$ be the SLT-tree for $(P\cup \{\leftarrow A\},TB_f^j)$.
Since $N_i$ is loop-independent, $SD_A=T_{\leftarrow A}$. Furthermore,
since no branches in $T_{\leftarrow A}$ end with a temporarily
undefined leaf, no new sub-derivations for $A$ will be generated
via further recursions of $SLT()$. Therefore, in view of
the fact that $TB_t^k$ consists of all
tabled positive answers in all $GT_{G_0}^l$s,
$\theta$ is a correct answer substitution for $A$ in $SD_A$
if and only if $A\theta$ is a tabled positive answer for $A$ in $TB_t^k$. 
And by Lemma \ref{sc-lem4}, $A$ is false in $P$ if and only if 
all branches of $SD_A$ end with a failure leaf. $\Box$

\vspace{4mm}

Theorem \ref{th5-1} allows us to make the following enhancement
of SLT-trees: 

\begin{optimization}
\label{opt1}
{\em
In Definition \ref{slt-tree}
change (c) of point 4 to (d) and add before it
\begin{enumerate}
\item[(c)]
If the root of $T_{\leftarrow A}$ is loop-independent
and all branches of $T_{\leftarrow A}$ end with a failure leaf
then $N_i$ has only one child that is labeled
by the goal $\leftarrow L_1,...,L_{j-1},L_{j+1},...,L_n$;
\end{enumerate}
}
\end{optimization}

\begin{figure}[htb]
\centering

\setlength{\unitlength}{3947sp}%
\begingroup\makeatletter\ifx\SetFigFont\undefined%
\gdef\SetFigFont#1#2#3#4#5{%
  \reset@font\fontsize{#1}{#2pt}%
  \fontfamily{#3}\fontseries{#4}\fontshape{#5}%
  \selectfont}%
\fi\endgroup%
\begin{picture}(1950,2805)(226,-2236)
\thinlines
\put(1651,-436){\vector( 0,-1){200}}
\put(1051,-1336){\vector( 0,-1){200}}
\put(596,-1976){\vector(-2,-1){0}}
\multiput(976,-1786)(-63.33333,-31.66667){6}{\makebox(1.6667,11.6667){\SetFigFont{5}{6}{\rmdefault}{\mddefault}{\updefault}.}}
\put(2101,389){\vector( 0,-1){200}}
\put(1646,-176){\vector(-2,-1){0}}
\multiput(2026, 14)(-63.33333,-31.66667){6}{\makebox(1.6667,11.6667){\SetFigFont{5}{6}{\rmdefault}{\mddefault}{\updefault}.}}
\put(1051,-1786){\vector( 0,-1){200}}
\put(2101,-11){\vector( 0,-1){200}}
\put(1046,-1101){\vector(-2,-1){0}}
\multiput(1426,-911)(-63.33333,-31.66667){6}{\makebox(1.6667,11.6667){\SetFigFont{5}{6}{\rmdefault}{\mddefault}{\updefault}.}}
\put(1426,-361){\makebox(0,0)[lb]{\smash{\SetFigFont{9}{10.8}{\rmdefault}{\mddefault}{\updefault}$N_2:$ $ b$ }}}
\put(1726,-586){\makebox(0,0)[lb]{\smash{\SetFigFont{8}{9.6}{\rmdefault}{\mddefault}{\updefault}$C_{b_1}$}}}
\put(826,-1261){\makebox(0,0)[lb]{\smash{\SetFigFont{9}{10.8}{\rmdefault}{\mddefault}{\updefault}$N_4:$ $c$ }}}
\put(1126,-1486){\makebox(0,0)[lb]{\smash{\SetFigFont{8}{9.6}{\rmdefault}{\mddefault}{\updefault}$C_{c_1}$}}}
\put(676,-1711){\makebox(0,0)[lb]{\smash{\SetFigFont{9}{10.8}{\rmdefault}{\mddefault}{\updefault}$N_5:$ $\neg d$ }}}
\put(226,-2236){\makebox(0,0)[lb]{\smash{\SetFigFont{9}{10.8}{\rmdefault}{\mddefault}{\updefault}$N_6:$ $d$ }}}
\put(451,-2086){\makebox(0,0)[lb]{\smash{\SetFigFont{9}{10.8}{\rmdefault}{\mddefault}{\updefault}$\Box_f$}}}
\put(1876,464){\makebox(0,0)[lb]{\smash{\SetFigFont{9}{10.8}{\rmdefault}{\mddefault}{\updefault}$N_0:$ $a$ }}}
\put(2176,314){\makebox(0,0)[lb]{\smash{\SetFigFont{8}{9.6}{\rmdefault}{\mddefault}{\updefault}$C_{a_1}$}}}
\put(1726, 89){\makebox(0,0)[lb]{\smash{\SetFigFont{9}{10.8}{\rmdefault}{\mddefault}{\updefault}$N_1:$ $\neg b$ }}}
\put(976,-2086){\makebox(0,0)[lb]{\smash{\SetFigFont{9}{10.8}{\rmdefault}{\mddefault}{\updefault}$\Box_t$ }}}
\put(901,-2236){\makebox(0,0)[lb]{\smash{\SetFigFont{9}{10.8}{\rmdefault}{\mddefault}{\updefault}$N_7:$ }}}
\put(2026,-361){\makebox(0,0)[lb]{\smash{\SetFigFont{9}{10.8}{\rmdefault}{\mddefault}{\updefault}$\Box_t$ }}}
\put(2026,-511){\makebox(0,0)[lb]{\smash{\SetFigFont{9}{10.8}{\rmdefault}{\mddefault}{\updefault}$N_8:$ }}}
\put(1201,-886){\makebox(0,0)[lb]{\smash{\SetFigFont{9}{10.8}{\rmdefault}{\mddefault}{\updefault}$N_3:$ $\neg c$ }}}
\put(1576,-736){\makebox(0,0)[lb]{\smash{\SetFigFont{9}{10.8}{\rmdefault}{\mddefault}{\updefault}$\Box_f$}}}
\end{picture}

\caption{The generalized SLT-tree $GT_{\leftarrow a}^1$ for 
$(P_2\cup \{\leftarrow a\},\emptyset)$. $\qquad\qquad$}\label{fig5-2}
\end{figure}
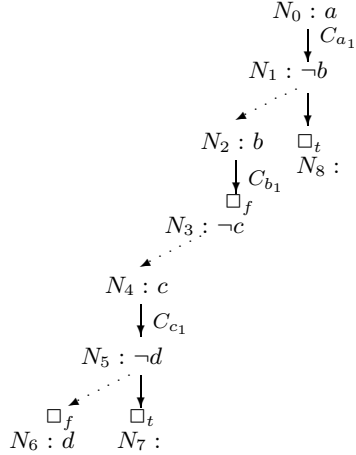

\begin{example}
\label{eg5-2}
{\em (Cont. of Example \ref{eg5-1})
By applying the optimized algorithm for constructing 
SLT-trees, SLT-resolution will build 
the generalized SLT-tree $GT_{\leftarrow a}^1$
as shown in Figure \ref{fig5-2}. Since $N_6$ is loop-independent,
by Theorem \ref{th5-1} $d$ is false and thus $\neg d$ is true, 
which leads to $c$ true and $\neg c$ false. Likewise, since
$N_2$ is loop-independent, $b$ is false, which leads to
$a$ true. As a result, to derive the first answer of $a$ 
$SLT()$ is called ones and $SLTP()$ ones, which shows 
a great improvement in efficiency over the former version.
}
\end{example}

It is easy to see that when the root of $T_{G_0}$ is loop-independent,
$T_{G_0}$ is an SLDNF-tree and thus
SLT-resolution coincides with SLDNF-resolution.
Due to this reason, we call Optimization \ref{opt1}, which reduces
recursions of $SLT()$, {\em negation as the finite 
failure of loop-independent nodes.}

\subsection{Answer Completion}
In this subsection we further optimize SLT-resolution 
by implementing the intuition that
if all answers of a positive literal $A$ have been derived and stored
in the table $TB_t^i$ or $TB_f^i$ after the generation of
$GT_{G_0}^i$, then all sub-derivations for $A$ in $GT_{G_0}^{i+1}$, which 
are generated by applying program clauses
(not tabled answers) to $A$, can be pruned because
they produce no new answers for $A$. Again we begin with an example.

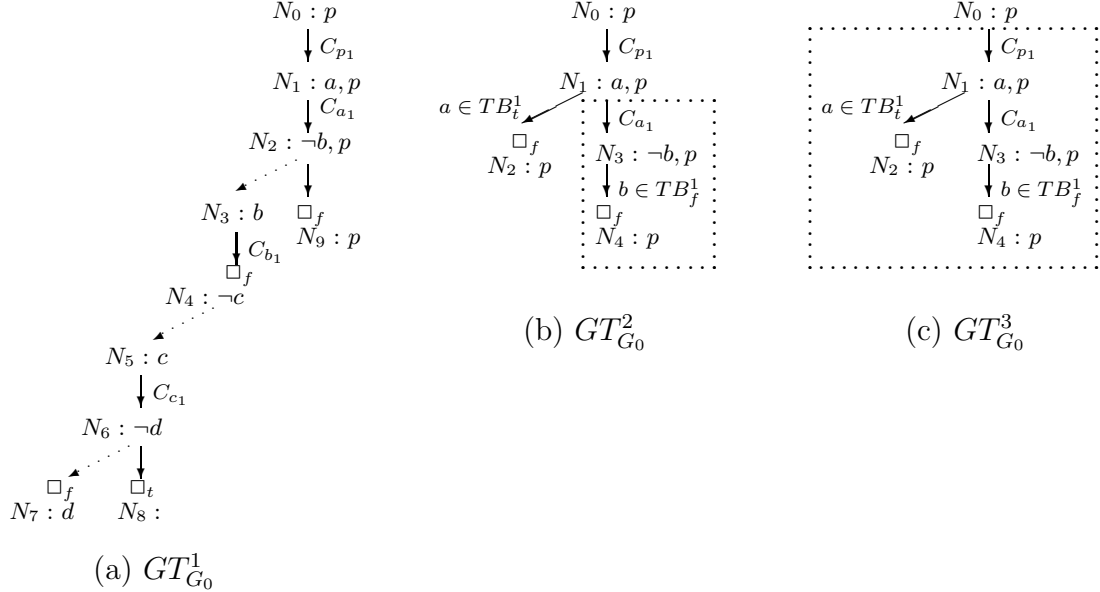
\begin{figure}[htb]
\centering

\setlength{\unitlength}{3947sp}%
\begingroup\makeatletter\ifx\SetFigFont\undefined%
\gdef\SetFigFont#1#2#3#4#5{%
  \reset@font\fontsize{#1}{#2pt}%
  \fontfamily{#3}\fontseries{#4}\fontshape{#5}%
  \selectfont}%
\fi\endgroup%
\begin{picture}(6847,3621)(526,-3436)
\thicklines
\multiput(4126,-1486)(75.00000,0.00000){12}{\makebox(6.6667,10.0000){\SetFigFont{10}{12}{\rmdefault}{\mddefault}{\updefault}.}}
\multiput(4126,-436)(75.00000,0.00000){12}{\makebox(6.6667,10.0000){\SetFigFont{10}{12}{\rmdefault}{\mddefault}{\updefault}.}}
\multiput(4126,-1486)(0.00000,70.00000){16}{\makebox(6.6667,10.0000){\SetFigFont{10}{12}{\rmdefault}{\mddefault}{\updefault}.}}
\multiput(4951,-1486)(0.00000,70.00000){16}{\makebox(6.6667,10.0000){\SetFigFont{10}{12}{\rmdefault}{\mddefault}{\updefault}.}}
\multiput(5551,-1486)(72.00000,0.00000){26}{\makebox(6.6667,10.0000){\SetFigFont{10}{12}{\rmdefault}{\mddefault}{\updefault}.}}
\multiput(5551, 14)(72.00000,0.00000){26}{\makebox(6.6667,10.0000){\SetFigFont{10}{12}{\rmdefault}{\mddefault}{\updefault}.}}
\multiput(5551,-1486)(0.00000,71.42857){22}{\makebox(6.6667,10.0000){\SetFigFont{10}{12}{\rmdefault}{\mddefault}{\updefault}.}}
\multiput(7351,-1486)(0.00000,71.42857){22}{\makebox(6.6667,10.0000){\SetFigFont{10}{12}{\rmdefault}{\mddefault}{\updefault}.}}
\put(1051,-3436){\makebox(0,0)[lb]{\smash{\SetFigFont{12}{14.4}{\rmdefault}{\mddefault}{\updefault}(a) $GT_{G_0}^1$}}}
\put(3751,-1936){\makebox(0,0)[lb]{\smash{\SetFigFont{12}{14.4}{\rmdefault}{\mddefault}{\updefault}(b) $GT_{G_0}^2$}}}
\put(6151,-1936){\makebox(0,0)[lb]{\smash{\SetFigFont{12}{14.4}{\rmdefault}{\mddefault}{\updefault}(c) $GT_{G_0}^3$}}}
\thinlines
\put(1351,-2161){\vector( 0,-1){200}}
\put(1351,-2611){\vector( 0,-1){200}}
\put(1951,-1261){\vector( 0,-1){200}}
\put(2401,-436){\vector( 0,-1){200}}
\put(1946,-1001){\vector(-2,-1){0}}
\multiput(2326,-811)(-63.33333,-31.66667){6}{\makebox(1.6667,11.6667){\SetFigFont{5}{6}{\rmdefault}{\mddefault}{\updefault}.}}
\put(2401,-836){\vector( 0,-1){200}}
\put(2401, 14){\vector( 0,-1){200}}
\put(896,-2801){\vector(-2,-1){0}}
\multiput(1276,-2611)(-63.33333,-31.66667){6}{\makebox(1.6667,11.6667){\SetFigFont{5}{6}{\rmdefault}{\mddefault}{\updefault}.}}
\put(4276,-436){\vector( 0,-1){200}}
\put(4276,-836){\vector( 0,-1){200}}
\put(4276, 14){\vector( 0,-1){200}}
\put(4126,-386){\vector(-2,-1){380}}
\put(6676,-436){\vector( 0,-1){200}}
\put(6676,-836){\vector( 0,-1){200}}
\put(6676, 14){\vector( 0,-1){200}}
\put(6526,-386){\vector(-2,-1){380}}
\put(1425,-1934){\vector(-2,-1){0}}
\multiput(1805,-1744)(-63.33333,-31.66667){6}{\makebox(1.6667,11.6667){\SetFigFont{5}{6}{\rmdefault}{\mddefault}{\updefault}.}}
\put(1126,-2086){\makebox(0,0)[lb]{\smash{\SetFigFont{9}{10.8}{\rmdefault}{\mddefault}{\updefault}$N_5:$ $c$ }}}
\put(1426,-2311){\makebox(0,0)[lb]{\smash{\SetFigFont{8}{9.6}{\rmdefault}{\mddefault}{\updefault}$C_{c_1}$}}}
\put(976,-2536){\makebox(0,0)[lb]{\smash{\SetFigFont{9}{10.8}{\rmdefault}{\mddefault}{\updefault}$N_6:$ $\neg d$ }}}
\put(526,-3061){\makebox(0,0)[lb]{\smash{\SetFigFont{9}{10.8}{\rmdefault}{\mddefault}{\updefault}$N_7:$ $d$ }}}
\put(751,-2911){\makebox(0,0)[lb]{\smash{\SetFigFont{9}{10.8}{\rmdefault}{\mddefault}{\updefault}$\Box_f$}}}
\put(1276,-2911){\makebox(0,0)[lb]{\smash{\SetFigFont{9}{10.8}{\rmdefault}{\mddefault}{\updefault}$\Box_t$ }}}
\put(1201,-3061){\makebox(0,0)[lb]{\smash{\SetFigFont{9}{10.8}{\rmdefault}{\mddefault}{\updefault}$N_8:$ }}}
\put(1501,-1711){\makebox(0,0)[lb]{\smash{\SetFigFont{9}{10.8}{\rmdefault}{\mddefault}{\updefault}$N_4:$ $\neg c$ }}}
\put(1876,-1561){\makebox(0,0)[lb]{\smash{\SetFigFont{9}{10.8}{\rmdefault}{\mddefault}{\updefault}$\Box_f$}}}
\put(1726,-1186){\makebox(0,0)[lb]{\smash{\SetFigFont{9}{10.8}{\rmdefault}{\mddefault}{\updefault}$N_3:$ $ b$ }}}
\put(2026,-1411){\makebox(0,0)[lb]{\smash{\SetFigFont{8}{9.6}{\rmdefault}{\mddefault}{\updefault}$C_{b_1}$}}}
\put(2176,-361){\makebox(0,0)[lb]{\smash{\SetFigFont{9}{10.8}{\rmdefault}{\mddefault}{\updefault}$N_1:$ $a,p$ }}}
\put(2476,-511){\makebox(0,0)[lb]{\smash{\SetFigFont{8}{9.6}{\rmdefault}{\mddefault}{\updefault}$C_{a_1}$}}}
\put(2026,-736){\makebox(0,0)[lb]{\smash{\SetFigFont{9}{10.8}{\rmdefault}{\mddefault}{\updefault}$N_2:$ $\neg b,p$ }}}
\put(2326,-1186){\makebox(0,0)[lb]{\smash{\SetFigFont{9}{10.8}{\rmdefault}{\mddefault}{\updefault}$\Box_f$ }}}
\put(2326,-1336){\makebox(0,0)[lb]{\smash{\SetFigFont{9}{10.8}{\rmdefault}{\mddefault}{\updefault}$N_9:$ $p$ }}}
\put(2176, 89){\makebox(0,0)[lb]{\smash{\SetFigFont{9}{10.8}{\rmdefault}{\mddefault}{\updefault}$N_0:$ $p$ }}}
\put(2476,-136){\makebox(0,0)[lb]{\smash{\SetFigFont{8}{9.6}{\rmdefault}{\mddefault}{\updefault}$C_{p_1}$}}}
\put(4201,-1186){\makebox(0,0)[lb]{\smash{\SetFigFont{9}{10.8}{\rmdefault}{\mddefault}{\updefault}$\Box_f$ }}}
\put(4051, 89){\makebox(0,0)[lb]{\smash{\SetFigFont{9}{10.8}{\rmdefault}{\mddefault}{\updefault}$N_0:$ $p$ }}}
\put(4351,-586){\makebox(0,0)[lb]{\smash{\SetFigFont{8}{9.6}{\rmdefault}{\mddefault}{\updefault}$C_{a_1}$}}}
\put(4201,-1336){\makebox(0,0)[lb]{\smash{\SetFigFont{9}{10.8}{\rmdefault}{\mddefault}{\updefault}$N_4:$ $p$ }}}
\put(4351,-136){\makebox(0,0)[lb]{\smash{\SetFigFont{8}{9.6}{\rmdefault}{\mddefault}{\updefault}$C_{p_1}$}}}
\put(3226,-511){\makebox(0,0)[lb]{\smash{\SetFigFont{8}{9.6}{\rmdefault}{\mddefault}{\updefault}$a\in TB_t^1$}}}
\put(4201,-811){\makebox(0,0)[lb]{\smash{\SetFigFont{9}{10.8}{\rmdefault}{\mddefault}{\updefault}$N_3:$ $\neg b,p$ }}}
\put(3976,-361){\makebox(0,0)[lb]{\smash{\SetFigFont{9}{10.8}{\rmdefault}{\mddefault}{\updefault}$N_1:$ $a,p$ }}}
\put(3676,-736){\makebox(0,0)[lb]{\smash{\SetFigFont{9}{10.8}{\rmdefault}{\mddefault}{\updefault}$\Box_f$ }}}
\put(3526,-886){\makebox(0,0)[lb]{\smash{\SetFigFont{9}{10.8}{\rmdefault}{\mddefault}{\updefault}$N_2:$ $p$ }}}
\put(6601,-1186){\makebox(0,0)[lb]{\smash{\SetFigFont{9}{10.8}{\rmdefault}{\mddefault}{\updefault}$\Box_f$ }}}
\put(6451, 89){\makebox(0,0)[lb]{\smash{\SetFigFont{9}{10.8}{\rmdefault}{\mddefault}{\updefault}$N_0:$ $p$ }}}
\put(6751,-586){\makebox(0,0)[lb]{\smash{\SetFigFont{8}{9.6}{\rmdefault}{\mddefault}{\updefault}$C_{a_1}$}}}
\put(6601,-1336){\makebox(0,0)[lb]{\smash{\SetFigFont{9}{10.8}{\rmdefault}{\mddefault}{\updefault}$N_4:$ $p$ }}}
\put(6751,-136){\makebox(0,0)[lb]{\smash{\SetFigFont{8}{9.6}{\rmdefault}{\mddefault}{\updefault}$C_{p_1}$}}}
\put(5626,-511){\makebox(0,0)[lb]{\smash{\SetFigFont{8}{9.6}{\rmdefault}{\mddefault}{\updefault}$a\in TB_t^1$}}}
\put(6601,-811){\makebox(0,0)[lb]{\smash{\SetFigFont{9}{10.8}{\rmdefault}{\mddefault}{\updefault}$N_3:$ $\neg b,p$ }}}
\put(6376,-361){\makebox(0,0)[lb]{\smash{\SetFigFont{9}{10.8}{\rmdefault}{\mddefault}{\updefault}$N_1:$ $a,p$ }}}
\put(6076,-736){\makebox(0,0)[lb]{\smash{\SetFigFont{9}{10.8}{\rmdefault}{\mddefault}{\updefault}$\Box_f$ }}}
\put(5926,-886){\makebox(0,0)[lb]{\smash{\SetFigFont{9}{10.8}{\rmdefault}{\mddefault}{\updefault}$N_2:$ $p$ }}}
\put(4351,-1036){\makebox(0,0)[lb]{\smash{\SetFigFont{8}{9.6}{\rmdefault}{\mddefault}{\updefault}$b\in TB_f^1$}}}
\put(6751,-1036){\makebox(0,0)[lb]{\smash{\SetFigFont{8}{9.6}{\rmdefault}{\mddefault}{\updefault}$b\in TB_f^1$}}}
\end{picture}

\caption{The generalized SLT-trees $GT_{G_0}^1$, $GT_{G_0}^2$
and $GT_{G_0}^3$. $\qquad\qquad$}\label{fig5-3}
\end{figure}

\begin{example}
\label{eg5-3}
{\em
Let $P_3$ be $P_2$ of Example \ref{eg5-1} plus the program clause
$C_{p_1}: p\leftarrow a,p$. Let $G_0=\leftarrow p$.
SLT-resolution (with Optimization \ref{opt1}) first builds the generalized
SLT-tree $GT_{G_0}^1$ as shown in Figure \ref{fig5-3} (a). Note that
$N_1-N_7$ are loop-independent nodes, and $N_0$ and $N_9$
are loop-dependent nodes. So $TB_t^1=\{c,a\}$ and 
$TB_f^1=\{d,b\}$. Using these tabled answers SLT-resolution
then builds the second generalized SLT-tree $GT_{G_0}^2$ as 
shown in Figure \ref{fig5-3} (b). Since no new tabled positive answers 
are generated in $GT_{G_0}^2$, $p$ is judged to be false.
Hence $TB_t^2=TB_t^1=\{c,a\}$ and $TB_f^2=\{d,b,p\}$.
Since $p$ is a new tabled negative answer, SLT-resolution
starts a new recursion $SLT(P_3\cup TB_t^2, G_0,R,TB_t^2,TB_f^2)$, 
which will build the third
generalized SLT-tree $GT_{G_0}^3$ that is the same as $GT_{G_0}^2$.
Since $GT_{G_0}^3$ contains no new tabled answers, the process stops.
}
\end{example}

Examining $GT_{G_0}^2$ in Figure \ref{fig5-3} we observe that since by
Theorem \ref{th5-1} all answers of $a$ have already been stored in $TB_t^1$, 
the sub-derivation for $a$ via the clause $C_{a_1}$ (circumscribed by
the dotted box) is redundant and hence can be removed. Similarly,
since the unique answer of $p$ has already been stored in $TB_f^2$, the 
circumscribed sub-derivations for $p$ via the clause $C_{p_1}$ in $GT_{G_0}^3$
are redundant and thus can be removed. We now discuss how to 
realize such type of optimization. 

First, we associate with each
selected positive literal $A$ (or its variant) a {\em completion}
flag $comp(A)$, defined by

\[comp(A)=\left\{\begin{array}{ll}
          Yes  & \mbox{if the answers of $A$ are completed;}\\
          No   & \mbox{otherwise.}
                 \end{array}
          \right. \] 

\noindent We say the answers of $A$ are {\em completed} if all its
answers have been stored in some $TB_t^i$ or $TB_f^i$. 
The determination of whether a selected positive literal
$A$ has got its complete answers is based on
Theorem \ref{th5-1}. That is, for a
selected positive literal $A$ at node $N_k$, $comp(A)=Yes$ if 
$N_k$ is loop-independent (assume Optimization \ref{opt1}
has already been applied). In addition, for each tabled negative answer 
$A$ in $TB_f^i$, $comp(A)$ should be $Yes$. 

Then, before applying program clauses to a selected
positive literal $A$ as in point 3 of Definition \ref{slt-tree},
we do the following:

\begin{optimization}
\label{opt2}
{\em
Check the flag $comp(A)$. If it is $Yes$ then apply to $A$ no program clauses
but tabled answers. 
}
\end{optimization}

\begin{example}
{\em (Cont. of Example \ref{eg5-3})
Based on $GT_{G_0}^1$ in Figure \ref{fig5-3}, $comp(a)$, $comp(b)$,
$comp(c)$ and $comp(d)$ will be set to $Yes$ since
$N_1$, $N_3$, $N_5$ and $N_7$ are loop-independent.
Therefore, the circumscribed sub-derivation in $GT_{G_0}^2$ will 
not be generated by the optimized SLT-resolution. 
Likewise, although $N_0:p$ in $GT_{G_0}^2$ is loop-dependent,
once $p$ is added to $TB_f^2$, $comp(p)$ will be set to $Yes$.
As a result, the circumscribed sub-derivations in $GT_{G_0}^3$
will never occur, so that $GT_{G_0}^3$ will consist only of a single
failure leaf at its root.
}
\end{example}

\subsection{Eliminating Duplicated Sub-Branches 
Based on a Fixed Depth-First Control Strategy}
\label{sec-dup}
Consider two selected positive literals $A_1$ at 
node $N_1$ and $A_2$ at node $N_2$ in
$GT_{G_0}^i$ such that $A_1$ is a variant of $A_2$.
Let $\{C_1,...,C_m\}$ be the set of program clauses in $P$ 
whose heads can unify with $A_1$. Then both $A_1$ and $A_2$ will
use all the $C_j$s except for looping clauses. This introduces
obvious redundant sub-branches, starting at $N_1$ and $N_2$ respectively. 
In this subsection we optimize SLT-resolution by eliminating this type of
redundant computations. We begin by making
the following two simple and yet practical assumptions.
\begin{enumerate}
\item
We assume that program clauses and tabled answers are
stored separately, and that new intermediate answers in SLT-trees 
are added into their tables once they are generated (i.e. new tabled 
positive answers are collected during the construction of each $GT_{G_0}^i$).
All tabled answers can be used once they are added to tables.
For instance, in Figure \ref{fig5-2} the intermediate answer $c$ is added to
the table $TB_t$ right after node $N_7$ is generated. Such an answer
can then be used thereafter. Obviously, this assumption does
not affect the correctness of SLT-resolution.
\item 
We assume nodes in each $GT_{G_0}^i$ are generated 
one after another in an order specified by a depth-first control strategy.
A {\em control strategy} consists of a search rule, a computation rule,
and policies for selecting program clauses and tabled answers.
A {\em search rule} is a rule for selecting a node among all nodes
in a generalized SLT-tree. A {\em depth-first} search rule is a 
search rule that starting from the root node always selects the most 
recently generated node. Depth-first rules are the most widely used search rules in
artificial intelligence and programming languages because they can
be very efficiently implemented using a simple stack-based memory structure.
For this reason, in this paper we choose depth-first control strategies,
i.e. control strategies with a depth-first search rule.
\end{enumerate}

The intuitive idea behind the optimization is that after a clause
$C_j$ has been completely used by $A_1$ at $N_1$, 
it needs not be used by $A_2$ at $N_2$. 
We describe how to achieve this.

Let $CS$ be a depth-first control strategy and assume $A_1$ 
at $N_1$ is currently selected by $CS$. 
Instead of generating all child nodes of $N_1$
by simultaneously applying to $A_1$ all 
program clauses and tabled answers (as in point 3
of Definition \ref{slt-tree}), each time only one clause or tabled answer, say
$C_j$, is selected by $CS$ to apply to $A_1$. This yields one child node,
say $N_s$. Then $N_s$ will be immediately expanded in the same way (recursively) since
it is the most recently generated node. 
After the expansion of $N_s$ has been finished, its parent node
$N_1$ is selected again by $CS$ (since it is 
the most recently generated node among all unfinished nodes) and expanded
by applying to $A_1$ another clause or tabled answer (selected by $CS$). 
If no new clause or tabled answer is left for $A_1$, which means that 
all sub-branches starting at $N_1$ in $GT_{G_0}^i$
have been exhausted, the expansion of $N_1$ is finished.
The control is then back to the parent node of $N_1$.
This process is usually called {\em backtracking}.
Continue this way until we finish the expansion of the root node
of $GT_{G_0}^i$. Since $GT_{G_0}^i$ is finite (for programs with the bounded-term-size
property), $CS$ is {\em complete} for SLT-resolution in the sense
that all nodes of $GT_{G_0}^i$ will be generated using this control strategy.
This shows a significant advantage over SLDNF-resolution, which
is incomplete with a depth-first control strategy because of possible
infinite loops in SLDNF-trees \cite{Ld87}. Moreover, the above description
clearly demonstrates that SLT-resolution is linear for query evaluation.

In the above description, when backtracking to $N_1$ from $N_s$, 
all sub-branches starting at $N_1$ via $C_j$ in $GT_{G_0}^i$ must
have been exhausted. In this case, we say $C_j$ {\em has been completely
used} by $A_1$. For each program clause $C_j$
whose head can unify with $A_1$, we associate with $A_1$ (or its variant) 
a flag $comp\_used(A_1,C_j)$, defined by  

\[comp\_used(A_1,C_j)=\left\{\begin{array}{ll}
          Yes  & \mbox{if $C_j$ has been completely used by $A_1$ (or its variant);}\\
          No   & \mbox{otherwise.}
                 \end{array}
          \right. \] 

From the above description we see that given a fixed depth-first
control strategy, program clauses will be selected and applied
in a fixed order. Therefore, by the time $C_j$ is selected for $A_2$ at $N_2$,
we check the flag $comp\_used(A_2,C_j)$. 
If $comp\_used(A_2,C_j)=Yes$ then $C_j$ needs not be applied to $A_2$
since similar sub-derivations have been completed before with all
intermediate answers along these sub-derivations already stored in tables 
for $A_2$ to use (under the above first assumption).

Observe that in addition to deriving new answers, the application of 
$C_j$ to $A_1$ may change the property of loop dependency of $N_1$,
which is important to Optimizations \ref{opt1} and \ref{opt2}.
That is, if some sub-branch starting at $N_1$ via $C_j$ contains
loop nodes then $N_1$ will be loop-dependent. If $N_1$ is loop-dependent,
neither Optimization \ref{opt1} nor Optimization \ref{opt2} is
applicable, so the answers of $A_1$ can be completed only through
the recursions of SLT-resolution. Since $A_2$ is a variant
of $A_1$, $N_2$ should have the same property as $N_1$. To achieve this,
we associate with $A_1$ (or its variant) a
flag $loop\_depend(A_1)$, defined by

\[loop\_depend(A_1)=\left\{\begin{array}{ll}
          Yes  & \mbox{if $A_1$ (or its variant)
                 has been selected at some} \\
              &  \mbox{loop-dependent node;}\\
          No   & \mbox{otherwise.}
                 \end{array}
          \right. \]
Then at node $N_2$ we check the flag.
If $loop\_depend(A_2)=Yes$ then mark $N_2$ as a loop node, 
so that $N_2$ becomes loop-dependent.
 
To sum up, SLT-trees can be generated using a fixed depth-first  
control strategy $CS$, where the following mechanism is used for selecting
program clauses (not tabled answers):

\begin{optimization}
\label{opt3}
{\em 
Let $A$ be the currently selected
positive literal at node $N_k$. If $loop\_depend(A)$ $=Yes$, 
mark $N_k$ as a loop node. A clause $C_j$ is selected for $A$
based on $CS$ such that $C_j$ is not a looping clause of $A$ and
$comp\_used(A,C_j)=No$.
}
\end{optimization}

\begin{theorem}
\label{th5-2}
Optimization \ref{opt3} is correct.
\end{theorem}

\noindent {\bf Proof:} The exclusion of looping clauses has been
justified in SLT-resolution before. Here we justify the exclusion
of program clauses that have been completely used.
Let $A_1$ at node $N_1$ and $A_2$ at node $N_2$
be two variant subgoals in $GT_{G_0}^i$ and let $C_j$ have been completely 
used by $A_1$ by the time $C_j$ is selected for $A_2$. 
Since we use a fixed depth-first control strategy, 
all sub-derivations for $A_1$ via $C_j$ must have been
generated, independently of applying $C_j$ to $A_2$.
This means that applying $C_j$ to $A_2$ will generate similar 
sub-derivations. Thus skipping $C_j$ at $N_2$ will not lose
any answers to $A_2$ provided that $A_2$ has access to the answers
of $A_1$ and that $N_2$ has the same property of loop dependency
as $N_1$. Clearly, that $N_2$ has the same property of loop dependency
as $N_1$ is guaranteed by using the flag $loop\_depend(A_1)$, and
the access of $A_2$ to the answers of $A_1$ is achieved by
the first assumption above. 

Observe that the application of
the first assumption may lead to more sub-derivations for $A_2$ via 
$C_j$ than those for $A_1$ via $C_j$. These extra sub-branches are
generated by using some newly added tabled answers, $S_1$, during
the construction of $GT_{G_0}^i$, which
were not yet available during the generation of sub-derivations of 
$A_1$ via $C_j$. If these extra sub-branches would yield new tabled answers, $S_2$,
the sub-derivations of $A_1$ via $C_j$ must have a loop. In this case, 
however, the newly added tabled answers $S_1$ will be
applied to the generation of sub-derivations of $A_1$ via $C_j$ in the
next recursion of SLT-resolution, which produces similar sub-branches
with new tabled answers $S_2$. Since $A_2$ is loop-dependent as $A_1$, it will be
generated in this recursion and use the answers $S_2$ from $A_1$.
$\Box$

\vspace{4mm}

The following two results show that redundant applications of
program clauses to variant subgoals are reduced by 
Optimization \ref{opt3}.

\begin{theorem}
\label{th5-3}
Let $A_1$ at node $N_1$ be an ancestor variant subgoal of $A_2$ at node $N_2$.
The program clauses used by the two subgoals are disjoint.
\end{theorem}

\noindent {\bf Proof:} Let $CS$ be a fixed depth-first control strategy
and $\{C_1,...,C_m\}$ be the set
of program clauses whose heads can unify with $A_1$. Assume these clauses 
are selected by $CS$ sequentially from left to right.
Since $A_1$ at $N_1$ is an ancestor variant of $A_2$ at $N_2$,
let $C_i$ be the clause via which the sub-branch starting at 
$N_1$ leads to $N_2$. Obviously, $C_i$ will not be used by $A_2$
since it is a looping clause of $A_2$. 

By Optimization \ref{opt3},
for each $1\leq j<i$ by the time $t$ when $C_i$ was selected for $A_1$ at $N_1$,
$C_j$ is either a looping clause of $A_1$ or $comp\_used(A_1,C_j)=Yes$.
Since $N_2$ was generated after $t$,  
$C_j$ is either a looping clause of $A_2$ or $comp\_used(A_2,C_j)=Yes$.
So $C_j$ will not be used by $A_2$ at $N_2$. 

Since $CS$ adopts a depth-first search rule,
by the time $t_1$ when $A_1$ tries to select the next clause $C_k$ $(k>i)$
$C_i$ must have been completely used by $A_1$ (via backtracking).
This implies that all $C_j$s $(i< j\leq m)$ must have been
completely used before $t_1$ by $A_2$. Hence
for no $i< j\leq m$ $C_j$ will be available to $A_1$. $\Box$

\begin{theorem}
\label{th5-4}
Let $A_1=p(.)$ and $C_{p_j}$ be a program clause whose head can unify
with $A_1$. Assume the number of tabled answers of $A_1$ is bounded by $N$.
Then $C_{p_j}$ is applied in $GT_{G_0}^i$ by $O(N)$ variant subgoals 
of $A_1$.
\end{theorem}

\noindent {\bf Proof:}
Let $\{A_1,...,A_m\}$ be the set of variant subgoals that are selected
in $GT_{G_0}^i$. The worst case is like this: The application of $C_{p_j}$
to $A_1$ yields the first tabled answer of $A_1$, but $C_{p_j}$ has not yet
been completely used after this. Next $A_2$ is selected, which uses the first
tabled answer and then applies $C_{p_j}$ to produce the second tabled
answer. Again $C_{p_j}$ has not yet been completely used after this.
Continue this way until $A_{N+1}$ is selected, which uses all the $N$
tabled answers and then applies $C_{p_j}$. This time it will fail to
produce any new tabled answer after exhausting all the remaining 
branches of $A_{N+1}$ via $C_{p_j}$. So $C_{p_j}$
has been completely used by $A_{N+1}$ and the flag $comp\_used(A_{N+1},C_{p_j})$
is set to $Yes$. Therefore $C_{p_j}$ will never be applied to any selected
variant subgoals of $A_1$ thereafter. $\Box$

\begin{example}
\label{dsz}
{\em
Consider the following program and let $G_0=\leftarrow p(X,5)$ be the 
top goal.\footnote{This program is suggested by B. Demon, K. Sagonas and N. F. Zhou.}
\begin{tabbing} 
\hspace{.2in} $P_3$: \= 
$p(X,N)\leftarrow loop(N),p(Y,N),odd(Y),X$ is $Y+1,X<N.$ \`$C_{p_1}$ \\
\> $p(X,N)\leftarrow p(Y,N),even(Y),X$ is $Y+1,X<N.$ \`$C_{p_2}$ \\
\> $p(1,N).$ \`$C_{p_3}$\\
\> $loop(N).$ \`$C_{l_1}$ 
\end{tabbing}
Here, $odd(Y)$ is true if $Y$ is an odd number, and 
$even(Y)$ is true if $Y$ is an even number.`` $X$ is $Y+1$'' is 
a meta-predicate which computes $Y+1$ and then assigns the 
result to $X$. 

We assume using the Prolog control strategy:
depth-first for node/goal selection + 
left-most for subgoal selection
+ top-down for clause selection. Obviously, it is a 
depth-first control strategy. We also assume using the 
first-in-first-out policy for selecting answers in tables.
If both program clauses and tabled answers are available,
tabled answers are used first. Let $CS$ represent the whole
control strategy. Then SLT-resolution (enhanced with 
Optimization \ref{opt3}) evaluates $G_0$ step by step and
generates a sequence of nodes $N_0,$ $N_1,$ $N_2,$ and so on, as shown in
Figures \ref{fig5-4} and \ref{fig5-5}.

\begin{figure}[htb]

\setlength{\unitlength}{3947sp}%
\begingroup\makeatletter\ifx\SetFigFont\undefined%
\gdef\SetFigFont#1#2#3#4#5{%
  \reset@font\fontsize{#1}{#2pt}%
  \fontfamily{#3}\fontseries{#4}\fontshape{#5}%
  \selectfont}%
\fi\endgroup%
\begin{picture}(5862,2871)(2476,-2536)
\thinlines
\put(6676,-1711){\vector( 0,-1){225}}
\put(6676,-1186){\vector( 0,-1){225}}
\put(7651,164){\vector( 3,-1){675}}
\put(6676,-736){\vector( 0,-1){225}}
\put(6676,-286){\vector( 0,-1){225}}
\put(6676,164){\vector(-3,-1){675}}
\put(6676,-2161){\vector( 0,-1){225}}
\put(7726,-736){\vector( 4,-1){300}}
\put(5176,-736){\vector(-4,-1){600}}
\put(3751,-1411){\vector( 0,-1){225}}
\put(6751,-1861){\makebox(0,0)[lb]{\smash{\SetFigFont{7}{8.4}{\rmdefault}{\mddefault}{\updefault}$X=2$}}}
\put(6751,-2236){\makebox(0,0)[lb]{\smash{\SetFigFont{7}{8.4}{\rmdefault}{\mddefault}{\updefault}Add $p(2,5)$ to $TB_t^1$}}}
\put(6751,-886){\makebox(0,0)[lb]{\smash{\SetFigFont{7}{8.4}{\rmdefault}{\mddefault}{\updefault}$p(1,5)$}}}
\put(6076,-1636){\makebox(0,0)[lb]{\smash{\SetFigFont{9}{10.8}{\rmdefault}{\mddefault}{\updefault}$N_6:$  $X$ is $1+1,X<5$}}}
\put(6451,-2086){\makebox(0,0)[lb]{\smash{\SetFigFont{9}{10.8}{\rmdefault}{\mddefault}{\updefault}$N_7:$  $2<5$}}}
\put(4276,-811){\makebox(0,0)[lb]{\smash{\SetFigFont{7}{8.4}{\rmdefault}{\mddefault}{\updefault}$C_{p_2}$}}}
\put(6076, 89){\makebox(0,0)[lb]{\smash{\SetFigFont{7}{8.4}{\rmdefault}{\mddefault}{\updefault}$C_{p_1}$}}}
\put(8101, 89){\makebox(0,0)[lb]{\smash{\SetFigFont{7}{8.4}{\rmdefault}{\mddefault}{\updefault}$p(1,5)$}}}
\put(6751,-436){\makebox(0,0)[lb]{\smash{\SetFigFont{7}{8.4}{\rmdefault}{\mddefault}{\updefault}$C_{l_1}$}}}
\put(6376,-2536){\makebox(0,0)[lb]{\smash{\SetFigFont{9}{10.8}{\rmdefault}{\mddefault}{\updefault}$N_8:$   $\Box_t$}}}
\put(2476,-1111){\makebox(0,0)[lb]{\smash{\SetFigFont{9}{10.8}{\rmdefault}{\mddefault}{\updefault}$N_3:$  $p(Y_1,5),even(Y_1),Y$ is $Y_1+1,Y<5,$}}}
\put(5626,-1111){\makebox(0,0)[lb]{\smash{\SetFigFont{9}{10.8}{\rmdefault}{\mddefault}{\updefault}$N_5:$  $odd(1),X$ is $1+1,X<5$}}}
\put(7801,-1111){\makebox(0,0)[lb]{\smash{\SetFigFont{9}{10.8}{\rmdefault}{\mddefault}{\updefault}$N_9:$  $odd(2),X$ is $2+1,X<5$}}}
\put(3976,-211){\makebox(0,0)[lb]{\smash{\SetFigFont{9}{10.8}{\rmdefault}{\mddefault}{\updefault}$N_1:$  $loop(5),p(Y,5),odd(Y),X$ is $Y+1,X<5$}}}
\put(5176,-661){\makebox(0,0)[lb]{\smash{\SetFigFont{9}{10.8}{\rmdefault}{\mddefault}{\updefault}$N_2:$  $p(Y,5),odd(Y),X$ is $Y+1,X<5$}}}
\put(8101,-811){\makebox(0,0)[lb]{\smash{\SetFigFont{7}{8.4}{\rmdefault}{\mddefault}{\updefault}$p(2,5)$}}}
\put(8101,-211){\makebox(0,0)[lb]{\smash{\SetFigFont{9}{10.8}{\rmdefault}{\mddefault}{\updefault}$N_{10}:$   $\Box_t$}}}
\put(8101,-961){\makebox(0,0)[lb]{\smash{\SetFigFont{9}{10.8}{\rmdefault}{\mddefault}{\updefault}$\Box_f$}}}
\put(3076,-1336){\makebox(0,0)[lb]{\smash{\SetFigFont{9}{10.8}{\rmdefault}{\mddefault}{\updefault}$odd(Y),X$ is $Y+1,X<5$}}}
\put(2626,-1936){\makebox(0,0)[lb]{\smash{\SetFigFont{9}{10.8}{\rmdefault}{\mddefault}{\updefault}$N_4:$  $even(1),Y$ is $1+1,Y<5,$}}}
\put(3001,-2161){\makebox(0,0)[lb]{\smash{\SetFigFont{9}{10.8}{\rmdefault}{\mddefault}{\updefault}$odd(Y),X$ is $Y+1,X<5$}}}
\put(3451,-1561){\makebox(0,0)[lb]{\smash{\SetFigFont{7}{8.4}{\rmdefault}{\mddefault}{\updefault}$C_{p_3}$}}}
\put(3826,-1561){\makebox(0,0)[lb]{\smash{\SetFigFont{7}{8.4}{\rmdefault}{\mddefault}{\updefault}Add $p(1,5)$ to $TB_t^1$}}}
\put(3676,-1786){\makebox(0,0)[lb]{\smash{\SetFigFont{9}{10.8}{\rmdefault}{\mddefault}{\updefault}$\Box_f$}}}
\put(6751,239){\makebox(0,0)[lb]{\smash{\SetFigFont{9}{10.8}{\rmdefault}{\mddefault}{\updefault}$N_0:$ $p(X,5)$ }}}
\end{picture}

\caption{$GT_{G_0}^1$. $\qquad\qquad\qquad\qquad\qquad$}\label{fig5-4}
\end{figure}

\begin{figure}

\setlength{\unitlength}{3947sp}%
\begingroup\makeatletter\ifx\SetFigFont\undefined%
\gdef\SetFigFont#1#2#3#4#5{%
  \reset@font\fontsize{#1}{#2pt}%
  \fontfamily{#3}\fontseries{#4}\fontshape{#5}%
  \selectfont}%
\fi\endgroup%
\begin{picture}(6225,7596)(3001,-6886)
\thinlines
\put(6301,389){\vector( 0,-1){225}}
\put(4426,389){\vector( 0,-1){225}}
\put(3601,389){\vector( 0,-1){225}}
\put(8626,389){\vector( 0,-1){225}}
\put(3601,389){\line( 1, 0){5025}}
\put(6301,-61){\vector( 0,-1){225}}
\put(6301,-661){\vector( 0,-1){225}}
\put(7951,-1861){\vector( 0,-1){225}}
\put(7951,-1336){\vector( 0,-1){225}}
\put(7951,-2311){\vector( 0,-1){225}}
\put(3601,-1861){\vector( 0,-1){225}}
\put(3601,-1336){\vector( 0,-1){225}}
\put(3601,-2311){\vector( 0,-1){225}}
\put(3676,-661){\line( 1, 0){5475}}
\put(6301,-661){\line( 0, 1){150}}
\put(3676,-661){\vector( 0,-1){225}}
\put(7951,-661){\vector( 0,-1){225}}
\put(4876,-661){\vector( 0,-1){225}}
\put(9151,-661){\vector( 0,-1){225}}
\put(6301,-1561){\line( 0,-1){1350}}
\put(4651,-3961){\vector( 0,-1){225}}
\put(4651,-4636){\vector( 0,-1){225}}
\put(4651,-5086){\vector( 0,-1){225}}
\put(4651,-6061){\vector( 0,-1){225}}
\put(4651,-5536){\vector( 0,-1){225}}
\put(4651,-6511){\vector( 0,-1){225}}
\put(7726,-3961){\vector( 0,-1){225}}
\put(7726,-4636){\vector( 0,-1){225}}
\put(3376,-2911){\line( 1, 0){5700}}
\put(3376,-2911){\vector( 0,-1){225}}
\put(6301,-2911){\vector( 0,-1){225}}
\put(9076,-2911){\vector( 0,-1){225}}
\put(4726,-2911){\vector( 0,-1){375}}
\put(7801,-2911){\vector( 0,-1){375}}
\put(6301,539){\line( 0,-1){150}}
\put(3151,239){\makebox(0,0)[lb]{\smash{\SetFigFont{7}{8.4}{\rmdefault}{\mddefault}{\updefault}$p(1,5)$}}}
\put(4051,239){\makebox(0,0)[lb]{\smash{\SetFigFont{7}{8.4}{\rmdefault}{\mddefault}{\updefault}$p(2,5)$}}}
\put(6001,239){\makebox(0,0)[lb]{\smash{\SetFigFont{7}{8.4}{\rmdefault}{\mddefault}{\updefault}$C_{p_1}$}}}
\put(8251, 14){\makebox(0,0)[lb]{\smash{\SetFigFont{9}{10.8}{\rmdefault}{\mddefault}{\updefault}$N_{29}:$   $\Box_t$}}}
\put(8701,239){\makebox(0,0)[lb]{\smash{\SetFigFont{7}{8.4}{\rmdefault}{\mddefault}{\updefault}$p(3,5)$}}}
\put(5701,-1036){\makebox(0,0)[lb]{\smash{\SetFigFont{9}{10.8}{\rmdefault}{\mddefault}{\updefault}$N_{10}:$  $p(Y_1,5),even(Y_1),$}}}
\put(7651,-1036){\makebox(0,0)[lb]{\smash{\SetFigFont{9}{10.8}{\rmdefault}{\mddefault}{\updefault}$N_{24}:$  $odd(3),$}}}
\put(8026,-2011){\makebox(0,0)[lb]{\smash{\SetFigFont{7}{8.4}{\rmdefault}{\mddefault}{\updefault}$X=4$}}}
\put(7351,-1786){\makebox(0,0)[lb]{\smash{\SetFigFont{9}{10.8}{\rmdefault}{\mddefault}{\updefault}$N_{25}:$  $X$ is $3+1,X<5$}}}
\put(7726,-2236){\makebox(0,0)[lb]{\smash{\SetFigFont{9}{10.8}{\rmdefault}{\mddefault}{\updefault}$N_{26}:$  $4<5$}}}
\put(7651,-2686){\makebox(0,0)[lb]{\smash{\SetFigFont{9}{10.8}{\rmdefault}{\mddefault}{\updefault}$N_{27}:$   $\Box_t$}}}
\put(3301,-1036){\makebox(0,0)[lb]{\smash{\SetFigFont{9}{10.8}{\rmdefault}{\mddefault}{\updefault}$N_5:$  $odd(1),$}}}
\put(3676,-2011){\makebox(0,0)[lb]{\smash{\SetFigFont{7}{8.4}{\rmdefault}{\mddefault}{\updefault}$X=2$}}}
\put(3001,-1786){\makebox(0,0)[lb]{\smash{\SetFigFont{9}{10.8}{\rmdefault}{\mddefault}{\updefault}$N_6:$  $X$ is $1+1,X<5$}}}
\put(3376,-2236){\makebox(0,0)[lb]{\smash{\SetFigFont{9}{10.8}{\rmdefault}{\mddefault}{\updefault}$N_7:$  $2<5$}}}
\put(3301,-2686){\makebox(0,0)[lb]{\smash{\SetFigFont{9}{10.8}{\rmdefault}{\mddefault}{\updefault}$N_8:$   $\Box_t$}}}
\put(3076,-1261){\makebox(0,0)[lb]{\smash{\SetFigFont{9}{10.8}{\rmdefault}{\mddefault}{\updefault}$X$ is $1+1,X<5$}}}
\put(3226,-811){\makebox(0,0)[lb]{\smash{\SetFigFont{7}{8.4}{\rmdefault}{\mddefault}{\updefault}$p(1,5)$}}}
\put(5926,-811){\makebox(0,0)[lb]{\smash{\SetFigFont{7}{8.4}{\rmdefault}{\mddefault}{\updefault}$C_{p_2}$}}}
\put(7576,-811){\makebox(0,0)[lb]{\smash{\SetFigFont{7}{8.4}{\rmdefault}{\mddefault}{\updefault}$p(3,5)$}}}
\put(5851,-1261){\makebox(0,0)[lb]{\smash{\SetFigFont{9}{10.8}{\rmdefault}{\mddefault}{\updefault}$Y$ is $Y_1+1,Y<5,$}}}
\put(5551,-1486){\makebox(0,0)[lb]{\smash{\SetFigFont{9}{10.8}{\rmdefault}{\mddefault}{\updefault}$odd(Y),X$ is $Y+1,X<5$}}}
\put(7501,-1261){\makebox(0,0)[lb]{\smash{\SetFigFont{9}{10.8}{\rmdefault}{\mddefault}{\updefault}$X$ is $3+1,X<5$}}}
\put(4426,-811){\makebox(0,0)[lb]{\smash{\SetFigFont{7}{8.4}{\rmdefault}{\mddefault}{\updefault}$p(2,5)$}}}
\put(4426,-1111){\makebox(0,0)[lb]{\smash{\SetFigFont{9}{10.8}{\rmdefault}{\mddefault}{\updefault}$N_9:$  $odd(2),...$}}}
\put(4801,-961){\makebox(0,0)[lb]{\smash{\SetFigFont{9}{10.8}{\rmdefault}{\mddefault}{\updefault}$\Box_f$}}}
\put(8776,-1111){\makebox(0,0)[lb]{\smash{\SetFigFont{9}{10.8}{\rmdefault}{\mddefault}{\updefault}$N_{28}:$  $odd(4),...$}}}
\put(9076,-961){\makebox(0,0)[lb]{\smash{\SetFigFont{9}{10.8}{\rmdefault}{\mddefault}{\updefault}$\Box_f$}}}
\put(5701,-3436){\makebox(0,0)[lb]{\smash{\SetFigFont{9}{10.8}{\rmdefault}{\mddefault}{\updefault}$N_{19}:$  $even(3),...$}}}
\put(3001,-3436){\makebox(0,0)[lb]{\smash{\SetFigFont{9}{10.8}{\rmdefault}{\mddefault}{\updefault}$N_{11}:$  $even(1),...$}}}
\put(4276,-3436){\makebox(0,0)[lb]{\smash{\SetFigFont{9}{10.8}{\rmdefault}{\mddefault}{\updefault}$N_{12}:$  $even(2),$}}}
\put(4051,-3661){\makebox(0,0)[lb]{\smash{\SetFigFont{9}{10.8}{\rmdefault}{\mddefault}{\updefault}$Y$ is $2+1,Y<5,$}}}
\put(4726,-6211){\makebox(0,0)[lb]{\smash{\SetFigFont{7}{8.4}{\rmdefault}{\mddefault}{\updefault}$X=4$}}}
\put(4726,-6586){\makebox(0,0)[lb]{\smash{\SetFigFont{7}{8.4}{\rmdefault}{\mddefault}{\updefault}Add $p(4,5)$ to $TB_t^2$}}}
\put(4426,-6436){\makebox(0,0)[lb]{\smash{\SetFigFont{9}{10.8}{\rmdefault}{\mddefault}{\updefault}$N_{17}:$  $4<5$}}}
\put(4351,-6886){\makebox(0,0)[lb]{\smash{\SetFigFont{9}{10.8}{\rmdefault}{\mddefault}{\updefault}$N_{18}:$   $\Box_t$}}}
\put(3901,-3886){\makebox(0,0)[lb]{\smash{\SetFigFont{9}{10.8}{\rmdefault}{\mddefault}{\updefault}$odd(Y),X$ is $Y+1,X<5$}}}
\put(3976,-4336){\makebox(0,0)[lb]{\smash{\SetFigFont{9}{10.8}{\rmdefault}{\mddefault}{\updefault}$N_{13}:$  $Y$ is $2+1,Y<5,$}}}
\put(3976,-4561){\makebox(0,0)[lb]{\smash{\SetFigFont{9}{10.8}{\rmdefault}{\mddefault}{\updefault}$odd(Y),X$ is $Y+1,X<5$}}}
\put(3676,-5011){\makebox(0,0)[lb]{\smash{\SetFigFont{9}{10.8}{\rmdefault}{\mddefault}{\updefault}$N_{14}:$ $3<5,odd(3),X$ is $3+1,X<5$}}}
\put(3676,-5461){\makebox(0,0)[lb]{\smash{\SetFigFont{9}{10.8}{\rmdefault}{\mddefault}{\updefault}$N_{15}:$ $odd(3),X$ is $3+1,X<5$}}}
\put(3901,-5986){\makebox(0,0)[lb]{\smash{\SetFigFont{9}{10.8}{\rmdefault}{\mddefault}{\updefault}$N_{16}:$  $X$ is $3+1,X<5$}}}
\put(4726,-4786){\makebox(0,0)[lb]{\smash{\SetFigFont{7}{8.4}{\rmdefault}{\mddefault}{\updefault}$Y=3$}}}
\put(4726,-5236){\makebox(0,0)[lb]{\smash{\SetFigFont{7}{8.4}{\rmdefault}{\mddefault}{\updefault}Add $p(3,5)$ to $TB_t^2$}}}
\put(7351,-3436){\makebox(0,0)[lb]{\smash{\SetFigFont{9}{10.8}{\rmdefault}{\mddefault}{\updefault}$N_{20}:$  $even(4),$}}}
\put(7126,-3661){\makebox(0,0)[lb]{\smash{\SetFigFont{9}{10.8}{\rmdefault}{\mddefault}{\updefault}$Y$ is $4+1,Y<5,$}}}
\put(6976,-3886){\makebox(0,0)[lb]{\smash{\SetFigFont{9}{10.8}{\rmdefault}{\mddefault}{\updefault}$odd(Y),X$ is $Y+1,X<5$}}}
\put(7051,-4336){\makebox(0,0)[lb]{\smash{\SetFigFont{9}{10.8}{\rmdefault}{\mddefault}{\updefault}$N_{21}:$  $Y$ is $4+1,Y<5,$}}}
\put(7051,-4561){\makebox(0,0)[lb]{\smash{\SetFigFont{9}{10.8}{\rmdefault}{\mddefault}{\updefault}$odd(Y),X$ is $Y+1,X<5$}}}
\put(7276,-5086){\makebox(0,0)[lb]{\smash{\SetFigFont{9}{10.8}{\rmdefault}{\mddefault}{\updefault}$N_{22}:$ $5<5,...$}}}
\put(7651,-4936){\makebox(0,0)[lb]{\smash{\SetFigFont{9}{10.8}{\rmdefault}{\mddefault}{\updefault}$\Box_f$}}}
\put(3301,-3286){\makebox(0,0)[lb]{\smash{\SetFigFont{9}{10.8}{\rmdefault}{\mddefault}{\updefault}$\Box_f$}}}
\put(6226,-3286){\makebox(0,0)[lb]{\smash{\SetFigFont{9}{10.8}{\rmdefault}{\mddefault}{\updefault}$\Box_f$}}}
\put(9076,-3286){\makebox(0,0)[lb]{\smash{\SetFigFont{9}{10.8}{\rmdefault}{\mddefault}{\updefault}$\Box_f$}}}
\put(8776,-3436){\makebox(0,0)[lb]{\smash{\SetFigFont{9}{10.8}{\rmdefault}{\mddefault}{\updefault}$N_{23}:$  $even(1),...$}}}
\put(3451,-3061){\makebox(0,0)[lb]{\smash{\SetFigFont{7}{8.4}{\rmdefault}{\mddefault}{\updefault}$p(1,5)$}}}
\put(4801,-3061){\makebox(0,0)[lb]{\smash{\SetFigFont{7}{8.4}{\rmdefault}{\mddefault}{\updefault}$p(2,5)$}}}
\put(6376,-3061){\makebox(0,0)[lb]{\smash{\SetFigFont{7}{8.4}{\rmdefault}{\mddefault}{\updefault}$p(3,5)$}}}
\put(7876,-3061){\makebox(0,0)[lb]{\smash{\SetFigFont{7}{8.4}{\rmdefault}{\mddefault}{\updefault}$p(4,5)$}}}
\put(9151,-3061){\makebox(0,0)[lb]{\smash{\SetFigFont{7}{8.4}{\rmdefault}{\mddefault}{\updefault}$C_{p_3}$}}}
\put(5926,614){\makebox(0,0)[lb]{\smash{\SetFigFont{9}{10.8}{\rmdefault}{\mddefault}{\updefault}$N_0:$ $p(X,5)$ }}}
\put(3151, 14){\makebox(0,0)[lb]{\smash{\SetFigFont{9}{10.8}{\rmdefault}{\mddefault}{\updefault}$N_1:$   $\Box_t$}}}
\put(3976, 14){\makebox(0,0)[lb]{\smash{\SetFigFont{9}{10.8}{\rmdefault}{\mddefault}{\updefault}$N_2:$   $\Box_t$}}}
\put(5026, 14){\makebox(0,0)[lb]{\smash{\SetFigFont{9}{10.8}{\rmdefault}{\mddefault}{\updefault}$N_3:$  $loop(5),p(Y,5),odd(Y),X$ is $Y+1,X<5$}}}
\put(5251,-436){\makebox(0,0)[lb]{\smash{\SetFigFont{9}{10.8}{\rmdefault}{\mddefault}{\updefault}$N_4:$  $p(Y,5),odd(Y),X$ is $Y+1,X<5$}}}
\put(6376,-211){\makebox(0,0)[lb]{\smash{\SetFigFont{7}{8.4}{\rmdefault}{\mddefault}{\updefault}$C_{l_1}$}}}
\put(9226,-811){\makebox(0,0)[lb]{\smash{\SetFigFont{7}{8.4}{\rmdefault}{\mddefault}{\updefault}$p(4,5)$}}}
\end{picture}

\caption{$GT_{G_0}^2$. $\qquad\qquad\qquad\qquad\qquad$}\label{fig5-5}
\end{figure}

Since $P_3$ is a positive program, $SLT(P_3,G_0,CS,\emptyset,\emptyset)
=SLTP(P_3,G_0,CS,\emptyset,\emptyset)$. The first generalized SLT-tree
$GT_{G_0}^1$ is shown in Figure \ref{fig5-4}. We
explain a few main points. At $N_3$ the (non-looping) program
clause $C_{p_3}$ is applied to $p(Y_1,5)$, which yields the
first tabled answer $p(1,5)$. $p(1,5)$ is immediately added to 
the table $TB_t^1$. After the failure of $N_4$, we 
backtrack to $N_3$ and then $N_2$. By this time $C_{p_3}$
has been completely used by $p(Y_1,5)$ at $N_3$, so we set 
$comp\_used(p(Y_1,5),C_{p_3})=Yes$. Due to this $C_{p_3}$ is skipped at
$N_2$. Applying the first tabled answer $p(1,5)$ to $p(Y,5)$ at $N_2$ 
generates $N_5$. At $N_8$ the second tabled answer $p(2,5)$
is produced, which yields the first answer to $G_0$. 
$p(2,5)$ is then applied to $p(Y,5)$ at $N_2$, leading to $N_9$.
When we backtrack to $N_0$ from $N_9$, $C_{p_2}$ 
has been completely used by $p(Y,5)$ at $N_2$. So both 
$C_{p_2}$ and $C_{p_3}$ are ignored at $N_0$. The
tabled answer $p(1,5)$ is then applied to $p(X,5)$ at $N_0$, yielding the second answer
$p(1,5)$ to $G_0$ at $N_{10}$. Note that the tabled answer 
$p(2,5)$ was obtained from a correct answer substitution
for $p(X,5)$ at $N_0$, so it was used by $p(X,5)$ while
it was generated. As a result, $GT_{G_0}^1$ is completed
with the table $TB_t^1=\{p(1,5),p(2,5)\}$.

We then do the first recursion of SLT-resolution by calling
$SLTP(P_3\cup TB_t^1,G_0,CS,TB_t^1,\emptyset)$, which builds the
second generalized SLT-tree $GT_{G_0}^2$ as shown in Figure \ref{fig5-5}.
From $GT_{G_0}^2$ we get two new tabled answers $p(3,5)$ and $p(4,5)$.
That is, $TB_t^2=\{p(1,5),p(2,5),p(3,5),p(4,5)\}$.

The second recursion of SLT-resolution is done by calling
$SLTP(P_3\cup TB_t^2,G_0,CS,TB_t^2,\emptyset)$, which
produces no new tabled answers. Therefore SLT-resolution stops here.

}
\end{example}

\begin{remark}
{\em
Consider node $N_{10}$ in $GT_{G_0}^2$ (Figure \ref{fig5-5}).
For each tabled answer $p(E,N)$ with $E$ an even number,
apply it to $p(Y_1,N)$ will always produce two new tabled answers
$p(E+1,N)$ and $p(E+2,N)$. Since these new answers will
be fed back immediately to $N_{10}$ for $p(Y_1,N)$ to use, all the 
remaining answers of $G_0$ will be produced at $N_{10}$.
This means that for any $N$ evaluating $p(X,N)$ requires doing
at most two recursions of SLT-resolution.
}
\end{remark}

It is easy to combine Optimizations \ref{opt1}, \ref{opt2} and \ref{opt3} 
with Definition \ref{slt-tree}, which leads to an algorithm
for generating {\em optimized} SLT-trees based on a fixed
depth-first control strategy, as described in appendix \ref{apx1}.
This algorithm is useful for the implementation of SLT-resolution.

\subsection{Computational Complexity of SLT-Resolution}

Theorem \ref{termin-normal} shows that
SLT-resolution terminates in finite time
for any programs with the bounded-term-size property.
In the above subsections we present three effective optimizations 
for reducing redundant computations. In this subsection we prove
the computational complexity of (the optimized) SLT-resolution.

SLT-resolution evaluates queries by building some generalized
SLT-trees. So the size of these generalized SLT-trees, i.e.
the number of edges (except for the dotted edges) in the trees,
represents the major part of its computational complexity. Since each edge
in an SLT-tree is generated by applying either a program clause
or a tabled answer, the size of a generalized SLT-tree is the number of 
applications of program clauses and tabled answers during the resolution.

The following notation is borrowed from \cite{chen96}.

\begin{definition}
{\em
Let $P$ be a program. Then $|P|$ denotes the number of clauses in $P$,
and $\Pi_P$ denotes the maximum number of literals in the body of
a clause in $P$. Let $s$ be an arbitrary positive integer. Then $N(s)$
denotes the number of atoms of predicates in $P$ that are not variants 
of each other and whose arguments do not exceed $s$ in size.
}
\end{definition}

\begin{theorem}
\label{th5-5}
Let $P$ be a program with the bounded-term-size property,
$G_0=\leftarrow A$ be a top goal (with $A$ an atom), and $CS$
be a fixed depth-first
control strategy. Then the size of each generalized 
SLT-tree $GT_{G_0}^i$ is $O(|P|N(s)^{\Pi_P+2})$ for some $s>0$.
\end{theorem}

\noindent {\bf Proof:}
Let $n$ be the maximum size of arguments in $A$. Since $P$ has
the bounded-term-size property, neither subgoal nor tabled answer has 
arguments whose size exceeds $f(n)$ for some function $f$. 
Let $s=f(n)$. Then the number of distinct subgoals (up to
variable renaming) in $GT_{G_0}^i$ is bounded by $N(s)$.

Let $B=p(.)$ be a subgoal. By Theorem \ref{th5-4}, each clause
$C_{p_j}$ will be applied to all variant subgoals of $B$
in $GT_{G_0}^i$ at most $N(s)+1$ times. So the number of applications
of all program clauses to all selected subgoals in $GT_{G_0}^i$ is bounded by 
\begin{equation}
N(s)*|P|*(N(s)+1)
\qquad\qquad\qquad\qquad\qquad\qquad \label{num-cl}
\end{equation}
Moreover, when a program clause is applied, it 
introduces at most $\Pi_P$ subgoals. Since the number of tabled
answers to each subgoal is bounded by $N(s)$, the $\Pi_P$ subgoals
access at most $N(s)^{\Pi_P}$ times to tabled answers. Hence the number of
applications of tabled answers to all subgoals in $GT_{G_0}^i$ 
is bounded by 
\begin{equation}
N(s)*|P|*(N(s)+1)*N(s)^{\Pi_P}
\qquad\qquad\qquad\qquad\qquad \label{num-ans}
\end{equation}
Therefore the size of $GT_{G_0}^i$ is bounded by $(\ref{num-cl}) + (\ref{num-ans})$,
i.e. $O(|P|N(s)^{\Pi_P+2})$. $\Box$

\vspace{4mm}

The second part of the computational complexity of SLT-resolution comes from
loop checking, which occurs during the determination of looping
clauses (see point 3 of Definition \ref{slt-tree}). Let $A_k=p(.)$ be 
a selected subgoal at node $N_k$ in $GT_{G_0}^i$ and 
$AL_{A_k}=\{(N_{k-1},A_{k-1}),...,(N_0,A_0)\}$ be its ancestor list.
For convenience we express the ancestor-descendant relationship in
$AL_{A_k}$ as a path like
\begin{equation}
N_0:A_0\Rightarrow_{C_{A_0}}...N_j:A_j\Rightarrow_{C_{A_j}}
...N_{k-1}:A_{k-1}\Rightarrow_{C_{A_{k-1}}} N_k: A_k
\qquad\qquad\qquad \label{ans-des}
\end{equation}
where $C_{A_j}$ is a program clause used by $A_j$. By Definitions
\ref{alist} and \ref{loop-def}, $N_0$ is the root of $GT_{G_0}^i$
and $A_j$ is an ancestor subgoal of $A_{j+l}$ $(0\leq j<k, l>0)$.
If $A_j$ is a variant of $A_k$, a loop occurs between $N_j$ and $N_k$
so that the looping clause $C_{A_j}$ will be skipped by $A_k$.

It is easily seen that $k$ subgoal comparisons may be made to check
if $A_k$ has ancestor variants. So if we do such loop checking for
every $A_j$ in the path, then we may need $O(K^2)$ comparisons.

By Optimization \ref{opt3} program clauses are selected in a fixed
order which is specified by a fixed control strategy. Let all clauses
with head predicate $p$ be selected in the order: $C_{p_1},C_{p_2},
...,C_{p_m}$. Then $A_k$ and all its ancestor variant subgoals
should follow this order. Assume $A_j$ is the closest ancestor
variant subgoal of $A_k$ in the path (\ref{ans-des}). Let
$C_{A_j}=C_{p_l}$. Then by Optimization \ref{opt3} each 
$C_{p_h}$ $(h<l)$ either is a looping clause of $A_j$ or 
has been completely used by a variant of $A_j$. This applies to $A_k$
as well. So $A_k$ should skip all $C_{p_h}$s $(h\leq l)$. 
This shows the following important fact.

\begin{fact}
\label{f1}
{\em
To determine looping clauses or clauses that have been
completely used for $A_k$, it suffices to find the closest
ancestor variant subgoal of $A_k$.
}
\end{fact}

\begin{theorem}
\label{th5-6}
Let $P$ be a program with the bounded-term-size property,
$G_0=\leftarrow A$ be a top goal (with $A$ an atom), and $CS$
be a fixed depth-first 
control strategy. Then the number of subgoal comparisons
performed in searching for the closest ancestor variant subgoals 
of all selected subgoals in  each generalized 
SLT-tree $GT_{G_0}^i$ is $O(|P|N(s)^3)$.
\end{theorem}

\noindent {\bf Proof:}
Note that loop checking only relies on ancestor lists of subgoals, which
only depend on program clauses with non-empty bodies (see Definition
\ref{alist}). By formula (\ref{num-cl}) in the proof of Theorem \ref{th5-5},
the total number of applications of program clauses to all selected
subgoals in $GT_{G_0}^i$ is bounded by $N(s)*|P|*(N(s)+1)$. 
Since each subgoal in the ancestor-descendant path (\ref{ans-des})
has at most $|P|$ ancestor variant subgoals (i.e. the first variant uses the first
program clause, the second uses the second, ..., and the $|P|$-th uses
the last program clause), the length of the path
is bounded by $N(s)*|P|$. Assume in the worst case that all $N(s)*|P|*(N(s)+1)$ 
applications of clauses generate $N(s)+1$ ancestor-descendant paths
like (\ref{ans-des}) of length $N(s)*|P|$. Since
each subgoal in a path needs at most $N(s)$ comparisons to find its
closest ancestor variant subgoal, the number of comparisons for
all subgoals in each path is bounded by $N(s)*|P|*N(s)$. Therefore,
the total number of subgoal comparisons in $N(s)+1$ paths
is bounded by
\begin{equation}
N(s)*|P|*N(s)*(N(s)+1)
\qquad\qquad\qquad\qquad\qquad\qquad \label{num-comp}
\end{equation}  
i.e. $O(|P|N(s)^3)$. $\Box$

\vspace{4mm}

Combining Theorems \ref{th5-5} and \ref{th5-6} and Fact \ref{f1} 
leads to the following.

\begin{theorem}
\label{th5-7}
The time complexity of SLT-resolution is $O(|P|N(s)^{\Pi_P+3}log N(s))$.
\end{theorem}

\noindent {\bf Proof:}
The time complexity of SLT-resolution consists of the part of
accessing program clauses, which is formula (\ref{num-cl}) times the complexity
of accessing one clause, the part of accessing tabled answers,
which is formula (\ref{num-ans}) times the complexity
of accessing one tabled answer,
and the part of subgoal comparisons in loop checking,
which is formula (\ref{num-comp}) times the complexity
of comparing two subgoals. The access
to one program clause and the comparison of two subgoals
can be assumed to be in constant time. A global table of subgoals 
and their answers can be maintained, so that the time for retrieving and 
inserting a tabled answer can be assumed to be $O(log N(s))$.
So the time complexity of constructing one generalized SLT-tree $GT_{G_0}^i$ is
\begin{equation}
O((\ref{num-cl}) + (\ref{num-ans})*log N(s) + (\ref{num-comp}))
=O(|P|N(s)^{\Pi_P+2}log N(s))
\qquad\qquad\
\end{equation}
Since the number of $GT_{G_0}^i$s, i.e. the number of recursions of 
SLT-resolution, is bounded by $N(s)$ (since each $GT_{G_0}^i$ produces
at least one new tabled answer), the time complexity of SLT-resolution is
$O(|P|N(s)^{\Pi_P+3}log N(s))$. $\Box$

\vspace{4mm}

It is shown in \cite{VRS91} that the {\em data complexity} of
the well-founded semantics, as defined by Vardi \cite{vardi82},
is polynomial time for function-free programs. 
This is obviously true with SLT-resolution because in this case, $s=1$ and 
$N(1)$ is a polynomial in the size of the {\em extensional
database} (EDB) \cite{chen96}.

\section{Related Work}

So far only two operational 
procedures for top-down evaluation of the well-founded
semantics of general logic programs have been extensively
studied: Global SLS-resolution and SLG-resolution. 
Global SLS-resolution is not effective
since it is not terminating even for function-free
programs \cite{Prz89, Ross92}. Therefore, in this section  we make
a detailed comparison of SLT-resolution with SLG-resolution.

There are three major differences between these two approaches. First,
SLG-resolution is based on program
transformations, instead of on standard tree-based formulations
like SLDNF- or Global SLS-resolution. Starting from the predicates
of the top goal, it transforms (instantiates) a set of clauses, called a
{\em system}, into another system based on six basic transformation 
rules. A special class of literals, called {\em delaying literals},
is used to represent and handle
temporarily undefined negative literals.
Negative loops are identified by maintaining a {\em dependency graph}
of subgoals \cite{CSW95, chen96}. In contrast, SLT-resolution is based on SLT-trees
in which the flow of the query evaluation is naturally depicted by the ordered
expansions of tree nodes.
It appears that this style of formulations is easier for users to understand and 
keep track of the computation. In addition, SLT-resolution handles temporarily 
undefined negative literals simply by replacing them with $u^*$, and 
treats positive and negative loops in the same way based on ancestor 
lists of subgoals. 

The second difference is that like all existing tabling methods,
SLG-resolution adopts the solution-lookup mode. Since all variant 
subgoals acquire answers from the same source $-$ the solution node,
SLG-resolution essentially generates a search graph instead of a
search tree, where every lookup node has a hidden edge towards
the solution node, which demands the solution node to produce 
new answers. Consequently it has to jump back and forth between
lookup and solution nodes. This is the reason why
SLG-resolution is not linear for query evaluation. In contrast, SLT-resolution makes
linear tabling derivations by generating SLT-trees. SLT-trees
can be viewed as SLDNF-trees with no infinite loops and with
significantly less redundant sub-branches.

Since SLG-resolution deviates from SLDNF-resolution,
some standard Prolog techniques for the implementation of SLDNF-resolution, 
such as the depth-first control strategy 
and the efficient stack-based memory management,\footnote{Bol and 
Degerstedt \cite{BD93} defined a special depth-first strategy that may be suitable
for SLG-resolution. However, their definition of ``depth-first''
is quite different from the standard one used in Prolog \cite{Ld87, NM95}.} 
cannot be used for its
implementation. This shows a third essential difference. SLT-resolution
bridges the gap between the well-founded semantics
and standard Prolog implementation techniques, and
can be implemented by an extension to any existing Prolog abstract machines 
such as WAM or ATOAM.  

The major shortcoming of SLT-resolution is that it is a little more time
costly than SLG-resolution. The time complexity of SLG-resolution
is $O(|P|N(s)^{\Pi_P+1}log N(s))$ \cite{chen96}, whereas ours is
$O(|P|N(s)^{\Pi_P+3}log N(s))$ (see Theorem \ref{th5-7}). The extra
price of our approach, i.e. $O(N(s))$ recursions (see Definition \ref{slt})
and $O(N(s))$ applications of each program clause to each distinct 
(up to variable renaming) subgoal (see Theorem \ref{th5-4}), 
is paid for the preservation of the linearity
for query evaluation. It should be pointed out, however, that
in practical situations, the number of recursions and that of clause 
applications are far less than $O(N(s))$. We note that in many typical cases,
such as Examples \ref{eg3-3}, \ref{eg5-2} and \ref{dsz}, 
both numbers are less than $3$. Moreover, the efficiency of SLT-resolution
can be further improved by completing its recursions locally; see
\cite{shen2000} for such special techniques.

Finally, for space consumption we note that SLG-resolution takes much more space
than SLT-resolution. The solution-lookup mode used in SLG-resolution 
requires that solution nodes stay forever
whenever they are generated even if they will never be invoked later.
In contrast, SLT-resolution will easily reclaim the space through backtracking
using the efficient stack-based memory structure.

\section{Conclusion}
We have presented a new operational procedure, SLT-resolution,
for the well-founded semantics of general logic programs. Unlike
Global SLS-resolution, it is free of infinite loops and with significantly 
less redundant sub-derivations; it terminates for all programs
with the bounded-term-size property. Unlike SLG-resolution, it preserves
the linearity of SLDNF-resolution, which bridges the gap between
the well-founded semantics and standard Prolog implementation techniques.

Prolog has many well-known nice features, but the problem
of infinite loops and redundant computations considerably 
undermines its beauties. The general goal of our research is then
to extend Prolog with tabling 
to compute the well-founded semantics while resolving infinite
loops and redundant computations. SLT-resolution serves as a nice model for 
such an extension. (Note that XSB \cite{SSW94, SSWFR98} 
is the only existing system that top-down computes the well-founded semantics
of general logic programs, but it is not an extension of Prolog since
SLG-resolution and SLDNF-resolution are 
totally heterogeneous.) 

For positive programs, we have developed special
methods for the implementation of SLT-resolution based
on the control strategy used by Prolog \cite{shen2000}.
The handling of cuts of Prolog is also discussed there.
A preliminary report on methods for the implementation of
SLT-resolution for general logic programs appears in \cite{shen99}.

\section*{Acknowledgements}
The first author is supported in part by Chinese National
Natural Science Foundation and Trans-Century Training Program
Foundation for the Talents by the Chinese Ministry of Education.

\appendix
\section{Optimized SLT-Trees}
\label{apx1}

Assume that program clauses and tabled answers are
stored separately, and that new tabled positive answers in SLT-trees 
are added into the table $TB_t$ once they are generated (see Section \ref{sec-dup}).
Combining Optimizations \ref{opt1}, \ref{opt2} and \ref{opt3} in Section \ref{sec5} 
with Definition \ref{slt-tree}, we obtain an algorithm
for generating optimized SLT-trees based on a fixed
depth-first control strategy. 

\begin{definition}[SLT-trees, an optimized version]
\label{opslt-tree}
{\em
Let $P=P^c\cup TB_t$ be a program with $P^c$ a set of program clauses
and $TB_t$ a set of tabled positive answers. Let $G_0$ be a top goal 
and $CS$ be a depth-first control strategy.
Let $TB_f$ be a set of ground atoms such that for each $A\in TB_f$ $\neg A\in WF(P)$.
The {\em optimized SLT-tree} $T_{G_0}$ for $(P \cup \{G_0\},TB_f)$ via $CS$ is a tree
rooted at node $N_0:G_0$, which is generated as follows.

\begin{enumerate}
\item
Select the root node for expansion.

\item
\label{l1} (Node Expansion)
Let $N_i:G_i$ be the node selected for expansion, with $G_i=\leftarrow L_1,...,L_n$. 

\begin{enumerate}
\item
If $n=0$ then mark $N_i$ by $\Box_t$ (a {\em success} leaf) and
goto \ref{l2} with $N=N_i$.

\item
If $L_1=u^*$ then mark $N_i$ by $\Box_{u^*}$ (a {\em temporarily undefined} 
leaf) and goto \ref{l2} with $N=N_i$.

\item
\label{l13}
Let $L_j$ be a positive literal selected by $CS$. Select a tabled answer
or program clause, $C$, from $P$ based on $CS$ 
while applying Optimizations \ref{opt2}
and \ref{opt3}. If $C$ is empty, then if $N_i$ has already had child nodes then
goto \ref{l2} with $N=N_i$ else mark $N_i$ by $\Box_f$ (a {\em failure} leaf)
and goto \ref{l2} with $N=N_i$. Otherwise, $N_i$ has a new child node labeled by
the resolvent of $G_i$ and $C$ over the literal $L_j$. Select the new child node
for expansion and goto \ref{l1}.

\item
Let $L_j=\neg A$ be a negative ground literal selected by $CS$. If  
$A$ is in $TB_f$ then $N_i$ has only one child that is labeled
by the goal $\leftarrow L_1,...,L_{j-1},L_{j+1},...,L_n$,
select this child node for expansion, and goto \ref{l1}. 
Otherwise, build an optimized SLT-tree
$T_{\leftarrow A}$ for $(P \cup \{\leftarrow A\},TB_f)$ via $CS$, where 
the subgoal $A$ at the root inherits the ancestor
list $AL_{L_j}$ of $L_j$. We consider the following cases:

\begin{enumerate}
\item
If $T_{\leftarrow A}$ has a success leaf then mark
$N_i$ by $\Box_f$ and goto \ref{l2} with $N=N_i$;

\item
\label{l142}
If the root of $T_{\leftarrow A}$ is loop-independent
and all branches of $T_{\leftarrow A}$ end with a failure leaf
then $N_i$ has only one child that is labeled
by the goal $\leftarrow L_1,...,L_{j-1},L_{j+1},...,L_n$,
select this child node for expansion, and goto \ref{l1};

\item
Otherwise, $N_i$ has only one child that is labeled
by the goal $\leftarrow L_1,...,L_{j-1},L_{j+1},$ $..., L_n,u^*$ if $L_n\neq u^*$
or $\leftarrow L_1,...,L_{j-1},L_{j+1},...,L_n$ if $L_n= u^*$. 
Select this child node for expansion and goto \ref{l1}.
\end{enumerate}
\end{enumerate}
\item
\label{l2} (Backtracking)
If $N$ is loop-independent and the selected literal $A$ at $N$
is positive then set $comp(A)=Yes$. If $N$ is the root node then return.
Otherwise, let $N_f:G_f$ be the parent node of $N$, with the selected
literal $L_f$. If $L_f$ is negative then goto \ref{l2} with $N=N_f$.
Else, if $N$ was generated from $N_f$ by resolving $G_f$ with a program
clause $C$ on $L_f$ then set $comp\_used(L_f,C)=Yes$.
Select $N_f$ for expansion and goto \ref{l1}.
\end{enumerate} 
}
\end{definition}

Optimization \ref{opt1} is used at item \ref{l142}.
Optimizations \ref{opt2} and \ref{opt3} are applied at item \ref{l13}
for the selection of program clauses. The flags $comp(\_)$ and 
$comp\_used(\_,\_)$ are updated during backtracking (point \ref{l2}).
The flag $loop\_depend(\_)$ is assumed to be updated automatically based
on loop dependency of nodes.

\end{document}